\documentclass{article}

\usepackage{microtype}
\usepackage{graphicx}
\usepackage{booktabs} 
\usepackage{soul} 

\usepackage{hyperref}

\usepackage[accepted]{icml2021}

\usepackage{multirow}

\usepackage{wrapfig}
\usepackage{subfig}

\usepackage{enumitem}

\usepackage{twoopt}
\usepackage{tikz}
\usetikzlibrary{fit}
\newcommand\tikzmark[1]{%
  \tikz[remember picture,overlay]\node[inner xsep=0pt] (#1) {};}
\newcommandtwoopt\Textbox[5][2.5cm][2cm]{%
\begin{tikzpicture}[remember picture,overlay]
  \coordinate (aux) at ([xshift=#1]#4);
  \node[inner ysep=3pt,yshift=0.6ex,draw=green,thick,
    fit=(#3) (aux),baseline] 
    (box) {};
  \node[text width=#2,anchor=north east,
    font=\sffamily\footnotesize,align=right] 
    at (box.north east) {#5};
\end{tikzpicture}%
}
\DeclareCaptionSubType*{algorithm}

\DeclareCaptionLabelFormat{alglabel}{Alg.~#2}

\usepackage[colorinlistoftodos]{todonotes}

\usepackage{xspace}

\newcommand{\road}{\textsc{Road}\xspace}
\newcommand{\roadca}{\textsc{Road-CA}\xspace}
\newcommand{\roadtx}{\textsc{Road-TX}\xspace}
\newcommand{\kgen}{\textsc{DbEn}\xspace}
\newcommand{\kgzh}{\textsc{DbZh}\xspace}
\newcommand{\kgenzh}{\textsc{Dbpd}\xspace}

\newcommand{\enron}{\textsc{Enro}\xspace}
\newcommand{\amazon}{\textsc{CoPr}\xspace}
\newcommand{\circuit}{\textsc{Circ}\xspace}
\newcommand{\ppi}{\textsc{HPpi}\xspace}

\newcommand{\ptc}{\textsc{Ptc}\xspace}
\newcommand{\aids}{\textsc{Aids}\xspace}
\newcommand{\linux}{\textsc{Linux}\xspace}
\newcommand{\imdb}{\textsc{Imdb}\xspace}
\newcommand{\mutag}{\textsc{Mutag}\xspace}
\newcommand{\web}{\textsc{Web}\xspace}
\newcommand{\mcspconn}{\textsc{mcsplain-connected}\xspace}
\newcommand{\nci}{\textsc{Nci109}\xspace}
\newcommand{\reddit}{\textsc{Reddit}\xspace}

\newcommand{\mcstreemodel}{\textsc{NeuralMcs}\xspace}
\newcommand{\randq}{\textsc{GLSearch-Rand}\xspace}

\newcommand{\mcsrlmodel}{\textsc{GLSearch}\xspace}
\newcommand{\mcsrlmodelscl}{\textsc{GLSearch-Scal}\xspace}

\newcommand{\ga}{\mathcal{G}_1\xspace}
\newcommand{\gb}{\mathcal{G}_2\xspace}
\newcommand{\gas}{\mathcal{G}_{1s}\xspace}
\newcommand{\gbs}{\mathcal{G}_{2s}\xspace}

\newcommand{\nnodes}{\textsc{Best Solution Size}\xspace}
\newcommand{\mcsp}{\textsc{McSp}\xspace}
\newcommand{\mcsprl}{\textsc{McSp+RL}\xspace}

\newcommand{\pca}{\textsc{I-pca}\xspace}

\newcommand{\gwqap}{\textsc{gw-qap}\xspace}

\usepackage{pifont}

\usepackage{amsmath}
\usepackage{amssymb}
\usepackage{amsthm}
\usepackage{bm}
\usepackage{algorithm}
\usepackage{algorithmic}

\usepackage{bbm}

\DeclareMathOperator*{\argmax}{arg\,max}

\icmltitlerunning{GLSearch: Maximum Common Subgraph Detection via Learning to Search
}

\begin{document}

\twocolumn[
\icmltitle{GLSearch: Maximum Common Subgraph Detection via Learning to Search}



\icmlsetsymbol{equal}{*}

\begin{icmlauthorlist}
\icmlauthor{Yunsheng Bai}{equal,ucla}
\icmlauthor{Derek Xu}{equal,ucla}
\icmlauthor{Yizhou Sun}{ucla}
\icmlauthor{Wei Wang}{ucla}
\end{icmlauthorlist}

\icmlaffiliation{ucla}{Department of Computer Science, University of California, Los Angeles, California, USA}

\icmlcorrespondingauthor{Yunsheng Bai}{yba@ucla.edu}
\icmlcorrespondingauthor{Derek Xu}{derekqxu@ucla.edu}

\icmlkeywords{Machine Learning, ICML}

\vskip 0.3in
]



\printAffiliationsAndNotice{\icmlEqualContribution} 


\begin{abstract}


Detecting the Maximum Common Subgraph (MCS) between two input graphs is fundamental for applications in drug synthesis, malware detection, cloud computing, etc. However, MCS computation is NP-hard, and state-of-the-art MCS solvers rely on heuristic search algorithms which in practice cannot find good solution for large graph pairs given a limited computation budget. We propose \mcsrlmodel, a Graph Neural Network (GNN) based {\it learning to search} model. Our model is built upon the branch and bound algorithm, which selects one pair of nodes from the two input graphs to expand at a time. Instead of using heuristics, we propose a novel GNN-based Deep Q-Network (DQN) to select the node pair, making the search process faster and more adaptive. To further enhance the training of DQN, we leverage the search process to provide supervision in a pre-training stage and guide our agent during an imitation learning stage. Experiments on synthetic and real-world graph pairs demonstrate that our model learns a search strategy that is able to detect significantly larger common subgraphs than existing MCS solvers given the same computation budget. \mcsrlmodel can be potentially extended to solve many other combinatorial problems with constraints on graphs. 

\end{abstract}

\section{Introduction}
\label{sec-intro}

\begin{figure}
\centering
\includegraphics[width=0.4\textwidth]{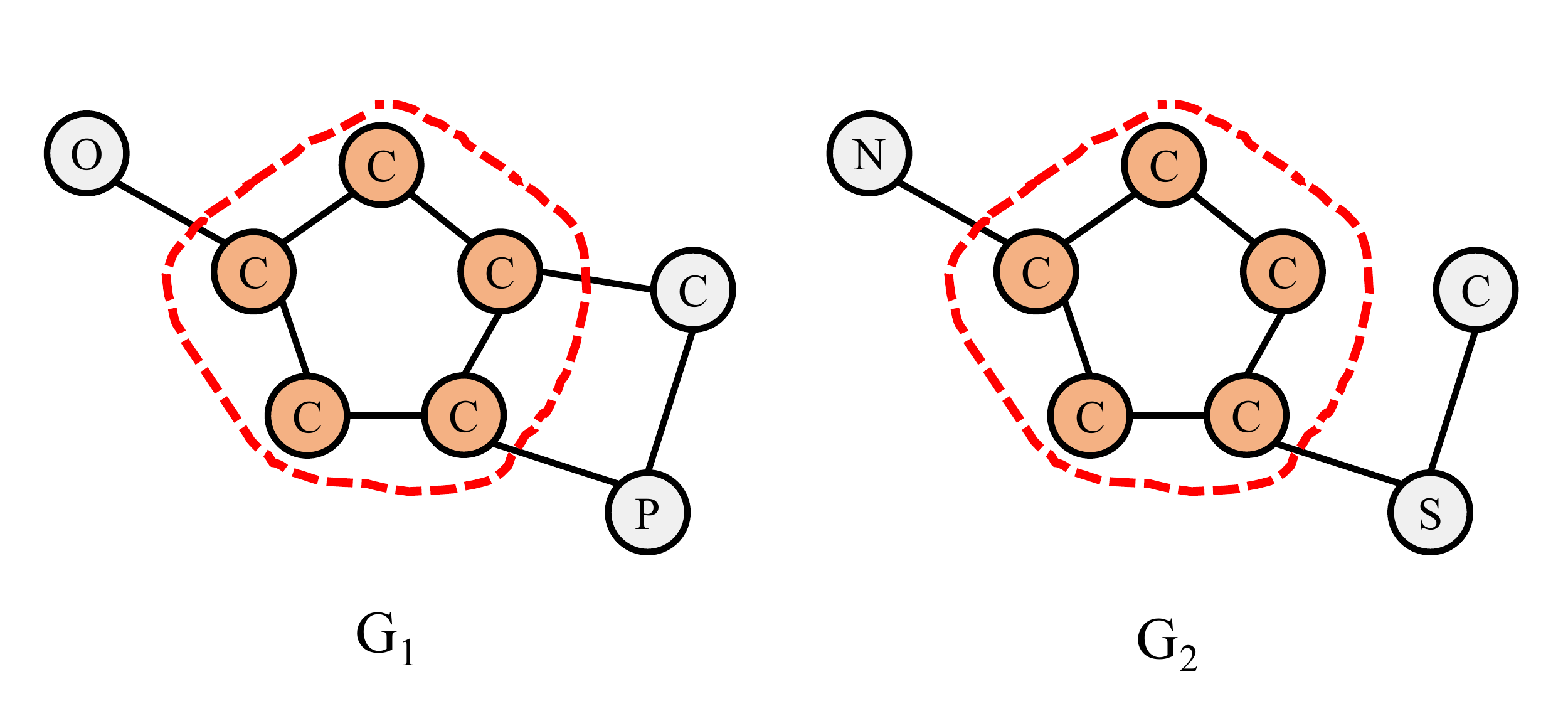}
\vspace*{-2mm}
\caption{For graph pair $(\mathcal{G}_1,\mathcal{G}_2)$ with node labels, the induced connected Maximum Common Subgraph (MCS) is the five-member ring structure highlighted in circle. 
}
\label{fig:mcs}
\vspace*{-6mm}
\end{figure}

Graphs gain increasing attention recently due to their expressive nature in representing real-world data and recent successes in addressing challenging graph tasks via learning, represented by graph neural networks. Among various graph tasks, 
detecting the largest subgraph that is commonly present in both input graphs, known as Maximum Common Subgraph (MCS)~\citep{bunke1998graph} (as shown in Figure~\ref{fig:mcs}), is 
an important yet particularly hard task.
MCS naturally encodes the degree of similarity between two graphs, is domain-agnostic, and thus has broad utilities
in many domains such as software analysis~\citep{park2013deriving}, graph database systems~\citep{yan2005substructure} and cloud computing platforms~\citep{cao2011privacy}. 
For example, in drug synthesis, finding similar substructures in compounds with similar properties can reduce manual labor~\citep{ehrlich2011maximum}.

MCS detection is NP-hard in its nature and is thus very challenging. The state-of-the-art exact MCS detection algorithms, which use a powerful branch and bound search framework, still run in exponential time in the worst case~\citep{liu2019learning}. These algorithms aim to provably extract the MCS by exhausting the search space as efficiently as possible. However, in large real-world graphs, exhausting the search space is not computationally tractable. What is worse, they rely on several heuristics on how to explore the search space.
For example, \mcsp~\citep{mccreesh2017partitioning} uses node degree as its heuristic by choosing high-degree nodes to visit first, but in many cases the true MCS contains low-degree nodes.

Recently, there are some related efforts from the learning community; 
however, these methods fall short in tackling the constraint posed by the MCS definition that the two extracted subgraphs must be isomorphic to each other.
For example, \citet{wang2019learning} aims to detect a soft matching matrix between nodes in two input graphs, which, however, cannot be easily transformed into the discrete matched subgraph. \citet{bai2020neural} is the first attempt to use learning based approach to directly output MCS. However, it heavily relies on labeled MCS instances, which requires pre-computation of MCS results by running exact solvers.



In this paper, we present \mcsrlmodel (\emph{\underline{G}}raph \emph{\underline{L}}earning to \emph{\underline{S}}earch), a general framework for MCS detection combining the advantages of search and deep reinforcement learning. \mcsrlmodel learns to search by adopting a Deep Q-Network (DQN)~\citep{mnih2015human} to replace the node selection heuristics required in state-of-the-art MCS solvers, leading to faster arrival of the optimal solution for an input graph pair, which is particularly useful
when applied to large real-world graphs and/or with a limited search budget. Our method reformulates DQN in a novel way to better capture the effect of different node selections, exploiting the representational power of Graph Neural Networks (GNN).
Given the large action space incurred by large graph pairs, to enhance the training of DQN, we leverage the search algorithm to not only provide supervised signals in a pre-training stage but also offer guidance during an imitation learning stage. 


Experiments on large real graph datasets (that are significantly larger than the datasets adopted by state-of-the-art MCS solvers) demonstrate that \mcsrlmodel outperforms baseline solvers and machine learning models for graph matching, in terms of effectiveness, by a large margin. Our contributions can be summarized as follows:
\begin{itemize}
\item We address the important yet challenging yet task of Maximum Common Subgraph detection for general-domain input graph pairs and propose \mcsrlmodel as the solution.
\item The key novelty is the GNN-based DQN which learns to search. With a DQN reformulation trick, it is trained under the reinforcement learning framework to make the best decision at each search step in order to quickly find the best MCS solution during search. The search in turns helps DQN training in a pre-training stage and an imitation learning stage.
\item We conduct extensive experiments on medium, large, and million-node real-world graphs to demonstrate the effectiveness of the proposed approach compared against a series of strong baselines in MCS detection and graph matching.
\end{itemize}

\section{Preliminaries and Related Work}
\label{sec-prelim}

\subsection{The MCS Detection Problem}
\label{subsec-probdef}

We denote a graph as $\mathcal{G}=(\mathcal{V},\mathcal{E})$ where $\mathcal{V}$ and $\mathcal{E}$ denote the vertex and edge set. An induced subgraph is defined as $\mathcal{G}_s=(\mathcal{V}_s,\mathcal{E}_s)$ where $\mathcal{E}_s$ preserves all the edges between nodes in $\mathcal{V}_s$, i.e. $\forall i,j \in \mathcal{V}_s$, $(i,j) \in \mathcal{E}_s$ if and only if $(i,j) \in \mathcal{E}$. 
In this paper, we aim to detect
the Maximum Common induced Subgraph (MCS) between an input graph pair, denoted as $\mathrm{MCS}(\ga,\gb)$, which is the largest induced subgraph contained in both $\ga$ and $\gb$. In addition, we require $\mathrm{MCS}(\ga,\gb)$ to be a connected subgraph. We allow the nodes of input graphs to be labeled, in which case the labels of nodes in the MCS must match as in Figure~\ref{fig:mcs}.
Graph isomorphism and subgraph isomorphism can be regarded as two special tasks of MCS: $|\mathrm{MCS}(\ga,\gb)| = |\mathcal{V}_1| = |\mathcal{V}_2|$ if $\ga$ are isomorphic to $\gb$, $|\mathrm{MCS}(\ga,\gb)| = \min{( |\mathcal{V}_1|,|\mathcal{V}_2|)}$ when $\ga$ (or $\gb$) is subgraph isomorphic to $\gb$ (or $\ga$).

\subsection{Related Work} 
\label{sec-related}

\paragraph{Traditional Efforts}
MCS detection is NP-hard, with existing methods based on constraint programming~\citep{vismara2008finding,mccreesh2016clique}, branch and bound~\citep{mccreesh2017partitioning,liu2019learning}, integer programming~\citep{bahiense2012maximum}, conversion to maximum clique detection~\citep{levi1973note,mccreesh2016clique}, etc., among which \mcsprl~\citep{liu2019learning} (details presented in Section~\ref{subsec-search}) is the state-of-the-art method, which guarantees to find common subgraphs satisfying the isomorphism constraint, but usually cannot extract large common subgraphs when input graphs become large. 


\paragraph{Efforts on Learning to Solve Graph Similarity and Matching} There is a growing trend of using machine learning approaches to graph matching and similarity score computation~\cite{zanfir2018deep,wang2019learning,Yu2020Learning,xu2019gromov,xu2019scalable,bai2018graph,bai2020convolutional,li2019graph,ling2020hierarchical}.
These methods cannot handle the isomorphism constraint in MCS well, since they were mainly designed for tasks without hard constraints, e.g. finding the similarity score or node-node matching between two graphs supervised by true similarity or matching.
Thus, to better satisfy the constraints of MCS, these models need to be embedded into a search framework that uses the scores provided by the models to guide the search for MCS, which will be described next. 
For example, \gwqap performs Gromov-Wasserstein discrepancy~\citep{peyre2016gromov} based optimization and outputs a matching matrix for all node pairs indicating the likelihood of matching~\citep{zanfir2018deep}. \pca performs image matching by outputting a doubly-stochastic matching matrix computed from intermediary Convolution Neural Network features from an input image pair~\citep{wang2019learning}.


\paragraph{Efforts on Learning to Solve NP-hard Graph Problems} Existing works such as \citet{dai2017learning} and \citet{fan2020finding}
focus on designing learning based approaches for solving NP-hard tasks on graphs, e.g. Minimum Vertex Cover, Network Dismantling, etc., but our problem, Maximum Common Subgraph detection, operates on a pair of input graphs instead of a single graph. Besides, MCS detection requires hard constraint satisfaction, i.e. isomorphism of extracted subgraphs, which is handled by a search algorithm described next.

\subsection{Search Algorithms for MCS}
\label{subsec-search}

\begin{figure*}[h]
\centering
\includegraphics[width=0.78\textwidth]{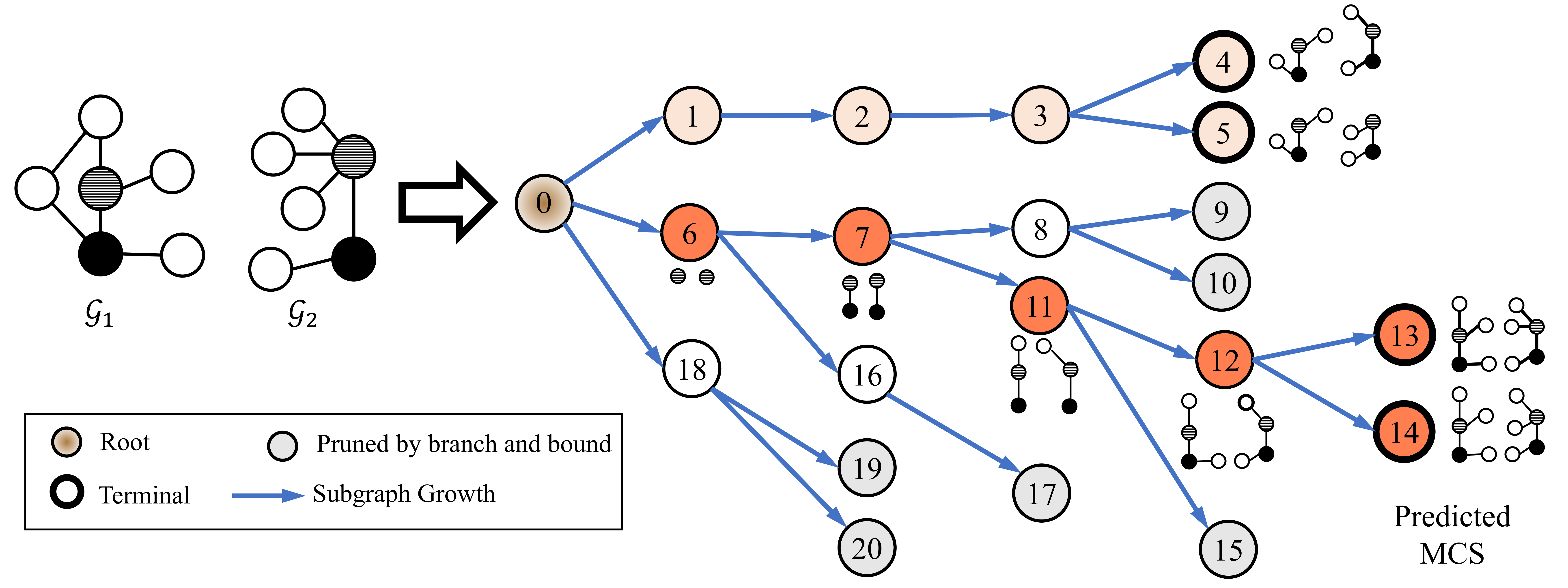}
\vspace{-3mm}
\caption{An illustration of the search process for MCS detection. For $(\ga,\gb)$, the branch and bound search algorithm (Section~\ref{subsec-search} and Algorithm~\ref{algo-mcs}) yields a tree structure where each node represents one state ($s_t$) with node id reflecting the order in which states are visited, and each edge represents an action ($a_t$) of selecting one more node pair. The search is essentially depth-first with pruning by the upper bound check. Our model learns the node pair selection strategy, i.e. which state to visit first. The policy, i.e. node pair to select, affects the order the search is tree is explored, e.g. if state 6 can be visited before state 1, a large solution can be found in less iterations (since $maxSol$ is larger in earlier search steps and more pruning may happen in subsequent search steps). When the search completes or a pre-defined search iteration budget is used up, the best solution (output subgraphs) will be returned, corresponding to state 13 (and 14). 
}
\label{fig:model}
\vspace*{-2mm}
\end{figure*}

In this section, we present the state-of-the-art branch and bound search framework for detecting MCS as shown in Algorithm~\ref{algo-mcs} and Figure~\ref{fig:model}, which allows the exploration of search space and guarantees the satisfaction of the isomorphism constraint posed by MCS. Thus, it serves as the backbone of our proposed approach. We then discuss several drawbacks in the existing search-based MCS detection algorithms.



\begin{algorithm}[h]
	\caption{Branch and Bound for MCS. 
We highlight in green boxes the two places that will be replaced by \mcsrlmodel.
	}
	\begin{algorithmic}[1]
		\STATE \textbf{Input:} Input graph pair $\ga$, $\gb$.
		
		\STATE \textbf{Output:} $maxSol$.

		\STATE Initialize $s_0 \leftarrow$ empty state.
		
		\STATE Initialize $\mathrm{stack} \leftarrow \texttt{new}$ Stack($s_0$). 
		
		\STATE Initialize $maxSol \leftarrow$ empty solution. 
		
		
    
        \WHILE {$\mathrm{stack} \neq \O$}
        
        
            \STATE 
            \tikzmark{start3}
            $s_{t} \leftarrow \mathrm{stack}.pop()$;\tikzmark{end3}
            
            \STATE $curSol \leftarrow s_{t}.getCurSol()$; 
            
            \IF {$|curSol| > |maxSol|$}
                \STATE $maxSol \leftarrow curSol$;
            \ENDIF
            
            
            \STATE $\mathit{UB}_t$ $\leftarrow$ $|curSol|$ + overestimate$(s_{t})$;
            
            \IF {$\mathit{UB}_t$ $\leq |maxSol|$ \textbf{or} $|s_t$.actions$|=0$}
                \STATE \textbf{continue};
            \ENDIF
            
            \STATE 
            $\mathcal{A}_{t} \leftarrow s_{t}$.actions;
            
            \STATE
            \tikzmark{start4}  $a_{t} \leftarrow policy(s_{t}, \mathcal{A}_{t})$;   \tikzmark{end4}
            
            \STATE 
            $s_{t}$.actions $\leftarrow s_{t}$.actions $\setminus \{a_{t}\}$;
            
            \STATE $\mathrm{stack}.push(s_{t})$;
            
            \STATE 
            $s_{t+1} \leftarrow$ environment.update($s_{t},\mathcal{A}_t$);
            
            \STATE $\mathrm{stack}.push(s_{t+1})$;

        \ENDWHILE
        
	\end{algorithmic}
	\label{algo-mcs}
	\Textbox{start3}{end3}{}
	\Textbox{start4}{end4}{}
	\vspace*{-4mm}

\end{algorithm}

\textbf{\mcsp and Its Limitations} \enspace The basic version, \mcsp, is presented in \citet{mccreesh2017partitioning} and the more advanced version, \mcsprl, is proposed in \citet{liu2019learning}. The whole search algorithm, outlined in Algorithm~\ref{algo-mcs}\footnote{The original algorithm is recursive. To highlight our novelty, we rewrite into an equivalent iterative version.}, is a branch-and-bound algorithm that, starting from an empty subgraph, grows the matching subgraph one node pair (between the two graphs) at a time and  maintains the best solution found so far. 
In each search iteration, denote the current search state as $s_t$ consisting of $\ga$, $\gb$, the current matched 
subgraphs $\gas=(\mathcal{V}_{1s},\mathcal{E}_{1s})$ and $\gbs=(\mathcal{V}_{2s},\mathcal{E}_{2s})$ as well as their node-node mappings. The algorithm tries to select one node pair, $(i,j)$ added to the currently selected subgraphs,
where node $i$ is from $\ga$ and node $j$ is from $\gb$, 
as its action, denoted as $a_t$. It then decides to either continue the search if the solution is promising, or otherwise backtrack to the parent search state, i.e. the current search state is pruned (line 14).
Various heuristics on node pair selection policy, denoted as ``$policy$'' in line 17, are proposed in \mcsp and \mcsprl. For example, in \mcsp, nodes of large degrees are selected before small-degree nodes.



\begin{figure}[h]
\centering
\includegraphics[width=0.4\textwidth]{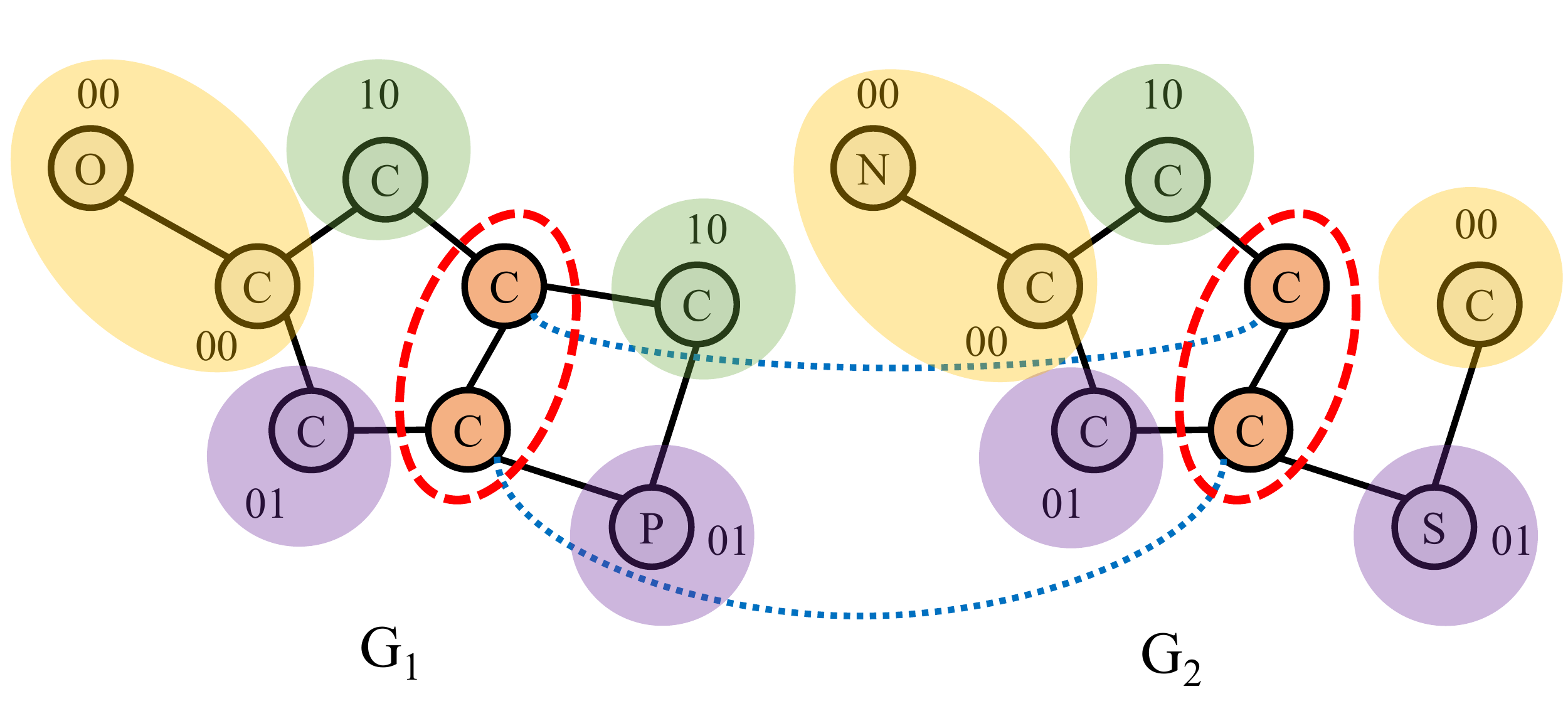}
\vspace*{-2mm}
\caption{An example to illustrate the concept of bidomains.
According to whether each node is connected to the two selected nodes (circled in red) or not, the nodes not in the current solution are split into three bidomains (Section~\ref{subsec-search}), denoted as ``00'' ($D_0$), ``01'' ($D_1$), and ``10'' ($D_2$), where ``0'' indicates not connected to a node in the selected two nodes, and ``1'' indicates connected. For example, each node in the ``10'' bidomain is connected to the top ``C'' node in the subgraph and disconnected to the bottom ``C'' node. By definition, the bidomain denoted with all zeros, e.g. ``00'' in this case, is called the disconnected bidomain. Notice the bidomains are derived from the node-node mappings between the two ``C'' nodes. 
}
\label{fig:bidomain}
\vspace*{-4mm}
\end{figure}

At each search state, in order to determine whether the solution is promising or not, an upper bound of the size of the MCS, ``$\mathit{UB}_t$'' in line 12 is computed. A concept of ``bidomain'' is introduced to facilitate its estimation.
Bidomains partition the nodes in the remaining subgraphs, i.e. outside $\gas$ and $\gbs$, into equivalent classes. Among all bidomains of a given state, $\mathcal{D}$, the $k$-th bidomain $D_{k}$ consists of two sets of nodes, $D_{k} = \langle \mathcal{V}'_{k1}, \mathcal{V}'_{k2} \rangle$ where $ \mathcal{V}'_{k1}$ and $ \mathcal{V}'_{k2}$ have the same connectivity pattern with respect to the already matched nodes $\mathcal{V}_{1s}$ and $\mathcal{V}_{2s}$. Figure~\ref{fig:bidomain} shows an example with three bidomains. 
Due to the subgraph isomorphism constraint posed by MCS, only nodes in $\mathcal{V}'_{k1}$ can match to $\mathcal{V}'_{k2}$ and vice versa. Since we require the MCS to be connected subgraphs, we differentiate bidomains $\mathcal{D}^{(c)}$ that are connected (adjacent) to $\gas$ and $\gbs$ (e.g. bidomain ``01'' and ``10'' in Figure~\ref{fig:bidomain}) from the single bidomain $D_0$ disconnected (unconnected) from $\gas$ and $\gbs$ (e.g. bidomain ``00'' in Figure~\ref{fig:bidomain}).
The candidate node pairs to select from, i.e. the action space ``$\mathcal{A}_{t}$'', consists of all node pairs in all connected bidomains, $\mathcal{D}^{(c)}$. This also guarantees the extracted subgraphs at each state are isomorphic to each other. 

To estimate the upper bound, it is noteworthy that each bidomain can contribute at most $\min(|\mathcal{V}'_{k1}|, |\mathcal{V}'_{k2}|)$ nodes to the future best solution. The upper bound can therefore be estimated as $\sum_{D_k \in \mathcal{D}} \min(|\mathcal{V}'_{k1}|, |\mathcal{V}'_{k2}|)$, which is the ``overestimate($s_t$)'' function in line 12. 
This upper bound computation is consistently used for all the methods in the paper. The major difference is in the policy for node pair selection, i.e. line 17.



As mentioned previously, \mcsp adopts a heuristic that selects node pairs with the largest degree as its policy. The most severe limitation of \mcsp is that the node-degree-based heuristic 
is not adaptive to the complex real-world graph structures.

\textbf{\mcsprl and Its Limitations} \enspace \mcsprl improves \mcsp by replacing the node pair selection policy with a value function for each node or node pair. Their goal is to minimize the search tree size so that a search tree leaf can be reached as early as possible. Specifically, \mcsprl aims to reduce the $\mathit{UB}_t$ to make it tighter, so that more pruning (line 14) can happen in subsequent search steps, 
resulting in a smaller search tree (search space).
To achieve that goal, they design the reward function for each state-action pair as the reduction (or reduction rate) of search space by selecting that node pair. The value function maintains a score for each node (or node pair), which is initialized to 0 and updated during search. In each step during search, the policy is to select the node pair with the largest score.

We identify two limitations of \mcsprl: 
(1) Since the reward definition is defined using a heuristic and there is no training stage, for each new graph pair, the scores must be re-initialized and the policy has to be re-updated. (2) The fact that the scores for each node (or node pair) are 0 at the beginning of search has another problem: \mcsprl breaks ties using node degrees, essentially degenerating to the same policy as \mcsp initially for each graph pair, which is also verified by our experiments 
as shown in Figure~\ref{fig-overhead} where \mcsp and \mcsprl perform the same.

Besides, a common issue of \mcsp and \mcsprl is that, during the search, they can enter a bad locally optimal search state and get ``stuck'' without finding a better (larger) solution, $maxSol$, for many iterations, as shown in the flat line segments of Figure~\ref{fig-overhead}.

\section{Proposed Method 
}
\label{sec-model}

In this section, we present our RL based MCS detection method, \mcsrlmodel. The rest of Section~\ref{sec-model} is organized as follows. Section~\ref{subsec-dqn-for-search} presents a high-level overview of how to leverage Deep Q-Network (DQN)~\cite{mnih2015human} for search, including the basic definitions of state, action, reward, etc., and how DQN can address the various issues of search methods for MCS described previously. Section~\ref{subsec-policy} describes the details of how to leverage DQN for search, focusing on how to effectively design representation learning for DQN for the task of MCS detection. Section~\ref{subsec-search-for-dqn} explains how to effectively train the DQN with the help of search, i.e. how search can in turn help DQN training.

\subsection{Leveraging DQN for Search: Overview}
\label{subsec-dqn-for-search}


\mcsrlmodel enables graph representation learning techniques to tackle the hard isomorphism constraint posed by MCS and uses deep Q-learning to select node pairs smartly in each search state.
\mcsrlmodel represents states and actions in continuous embeddings, and maps $(s_t,a_t)$ to a score $Q(s_t,a_t)$ via a DQN which consists of a Graph Neural Network encoder and learnable components to project the representations into the final score.
\mcsrlmodel is trained on a set of diverse small and medium-sized graphs, and once trained, can be applied to any new graph pair. 

Unlike \mcsp and \mcsprl, which aim to reduce the search tree size, the aim of our agent is to directly maximize the common subgraph size, allowing large common subgraph to be found even on very large graph pairs.

State $s_t$ consists of the (1) current selected subgraphs, (2) the node-node mappings between the nodes in the selected subgraphs, and (3) the input graphs. We include the node-node mappings as part of the state definition since node-node mappings can be used to derive the bidomain partitioning as illustrated in Figure~\ref{fig:bidomain}, which constrains the node pairs that can be selected in future, and thus affects the future common subgraph size. Action $a_t$ is defined as a node pair to select.
For \mcsrlmodel,
given our goal, the immediate reward for transitioning from one state to any next state is defined as $r_t=+1$ since one new node pair is selected, 
so that $Q(s_t,a_t)$ captures the largest common subgraph size starting at $s_t$ by performing $a_t$.

\mcsrlmodel is trained to find large common subgraphs quickly, but due to the large action space of large graph pairs, our model may still be susceptible to the local optimum without increasing $maxSol$ as described in Section~\ref{subsec-search}. Thus, when this occurs, 
we utilize additional information stored in the search tree to backtrack to a state that will most likely improve $maxSol$.
We find that in practice, states with a large action space, $\mathcal{A}_t$, tend to include more high-quality unexplored actions. Hence, if the best solution found so far does not increase\footnote{If the search does not enter line 10 of Algorithm~\ref{algo-mcs}.} for a pre-defined number of iterations, then in the next iteration, instead of popping from the stack\footnote{Line 7 of Algorithm~\ref{algo-mcs}.}, we find the state with the largest action space, 
and visit it. 
We refer to this improved search methodology as \textbf{promise-based search}.
More details can be found in the supplementary material.

\subsection{
Search Policy Learning via GNN-based DQN}
\label{subsec-policy}


Since the action space can be large for MCS, we leverage the representation learning capacity of continuous representations for DQN design.
At state $s_{t}$, for each action $a_{t}$, our DQN predicts a $Q(s_t,a_t)$ representing the remaining future reward after selecting action $a_t = (i, j)$ where $i\in \mathcal{V}_1$ and $j \in \mathcal{V}_2$, 
which intuitively corresponds to the largest number of nodes that will be eventually selected starting from the action edge $(s_t,a_t)$ as shown in tree in Figure~\ref{fig:model}.

Based on the above insights, one can design a simple DQN leveraging the representation learning power of Graph Neural Networks (GNN) such as \citet{kipf2016semi} and \citet{velickovic2017graph} by passing $\ga$ and $\gb$ to a GNN to obtain one embedding per node, $\{\bm{h}_i|\forall i \in \mathcal{V}_1\}$ and $\{\bm{h}_j|\forall j \in \mathcal{V}_2\}$. Denote $\textsc{concat}$ as concatenation, $\textsc{readout}$ as a readout operation that aggregates node-level embeddings into subgraph embeddings $\bm{h}_{s1}$ and $\bm{h}_{s2}$, and whole-graph embeddings $\bm{h}_{\mathcal{G}_1}$ and $\bm{h}_{\mathcal{G}_2}$. A state can then be represented as $\bm{h}_{s_t}=\textsc{concat}(\bm{h}_{\mathcal{G}_1},\bm{h}_{\mathcal{G}_2},\bm{h}_{s1},\bm{h}_{s2})$. An action can be represented as $\bm{h}_{a_t}=\textsc{concat}(\bm{h}_{i},\bm{h}_{j})$. The Q function would then be designed as:
\begin{eqnarray}
\begin{aligned}
\label{eq:dqn}
    Q(s_t,a_t) &=\textsc{mlp}\big( \textsc{concat}(\bm{h}_{s_t},\bm{h}_{a_t}) \big) \\
    &=
    \textsc{mlp}
    \big( \textsc{concat}
    (
    \bm{h}_{\mathcal{G}_1},\bm{h}_{\mathcal{G}_2}, 
    \bm{h}_{s1},\bm{h}_{s2},
    \bm{h}_{i},\bm{h}_{j}
    )
    \big).
\end{aligned}
\end{eqnarray}
However, there are several flaws to this simple design of Q function: 
\begin{enumerate}[label=(\Alph*)]
    \item \label{chall-nodepairweak} $\bm{h}_i$ and $\bm{h}_j$, generated by typical GNNs, encode only \emph{local} neighborhood information, but $Q(s_t,a_t)$ should capture the \emph{long-term} effect of adding $(i,j)$.
    What is worse, different node pairs have different embeddings, but their immediate rewards are always $+1$, a constant, in MCS, making differentiating the quality of different actions even more difficult.
    \item \label{chall-gorder} Swapping the order of $\ga$ and $\gb$ should not cause $Q(s_t,a_t)$ to change, but concatenating embeddings from the two graphs causes the DQN to be sensitive to their ordering.
    \item \label{chall-wholegraphcoarse} Lastly, how to effectively leverage the node-node mappings between $\gas$ and $\gbs$, an important part of the state definition as explained in Section~\ref{subsec-dqn-for-search}, for predicting $Q(s_t,a_t)$ remains a challenge.
\end{enumerate}

To address these issues, we propose the following improvements over the simple DQN design.

\textbf{Factoring out Action} \enspace
In order to maximally reflect the effect of adding node pair $(i,j)$ to $\gas$ and $\gbs$, we reformulate the optimal $Q$ score, $Q^*(s_t,a_t)$, as $r_t+\gamma V^*(s_{t+1})=1+\gamma V^*(s_{t+1})$ (using the fact that $r_t=+1$) in MCS, where $V$ is the value function, and $\gamma$ is the discount factor. Then, in order to compute the effect of $a_t$, we can compute the value associated with $s_{t+1}$ which does not depend on $a_{t}$ and avoids the use of local $\bm{h}_i$ and $\bm{h}_j$. In this case, we can rely on our state embedding to capture global information and amplify differences between different actions by looking at the states they will arrive. 


\textbf{Interaction between Input Graphs} \enspace
To resolve the graph symmetry issue, 
we first construct the interaction between the embeddings from two graphs, i.e. $\textsc{interact}(\bm{h}_{x1},\bm{h}_{x2})$, where $\bm{h}_{x1}$ and $\bm{h}_{x2}$ represent any embedding from $\ga$ and $\gb$ respectively, and $\textsc{interact}(\cdot)$ is any commutative function to combine the two embeddings (e.g. summation). This interacted embedding is later concatenated with other useful representations and fed into a final MLP to compute the $Q$ score. 


\textbf{Bidomain Representations} \enspace
Bidomains are derived from node-node mappings and partition the rest of $\ga$ and $\gb$, which is a more useful signal for predicting the future reward. In fact, as described in Section~\ref{subsec-search}, bidomains have been adopted in search-based MCS solvers to estimate the upper bound. Here, we require the harder prediction of $Q(s_t,a_t)$ for which we propose to also use the representation of bidomains to amplify the differences in different states. Denote $\bm{h}_{D_k}$ as the representation for bidomain $D_k$ $=\langle \mathcal{V}'_{k1}, \mathcal{V}'_{k2} \rangle$. 
Similar to computing the graph-level and subgraph-level embeddings, we compute $\bm{h}_{D_k}$ as
\begin{eqnarray}
\begin{aligned}
\label{eq:connected_bidomain}
    \bm{h}_{D_k} = \textsc{interact}\big(
    &\textsc{readout} (\{\bm{h}_{i} | i \in \mathcal{V}'_{k1} \}), \\ 
    &\textsc{readout} (\{\bm{h}_{j} | j \in \mathcal{V}'_{k2} \})
    \big).
\end{aligned}
\end{eqnarray}
Given all the bidomain embeddings, we compute a single representation for all the connected bidomains, $\mathcal{D}^{(c)}$, $\bm{h}_{\mathcal{D}c}=\textsc{readout} (\{\bm{h}_{D_k} | k \in \mathcal{D}^{(c)} \})$. Our final DQN has the form:
\begin{eqnarray}
\begin{aligned}
\label{eq:dvn_final}
    Q(s_t,a_t) = 1+ & \gamma \textsc{mlp}
    \Big( \textsc{concat}
    \big(
    \textsc{interact}(\bm{h}_{\mathcal{G}_1},\bm{h}_{\mathcal{G}_2}), \\
    &\textsc{interact}(\bm{h}_{s1},\bm{h}_{s2}), \bm{h}_{\mathcal{D}c},
    \bm{h}_{D_0}
    \big)
    \Big).
\end{aligned}
\end{eqnarray}

\subsection{Leveraging Search for DQN Training}
\label{subsec-search-for-dqn}

For large graph pairs, the action space can be quite large.
Thus, to enhance the training of our DQN, before the standard training of DQN~\citep{mnih2013playing}, we pre-train DQN and guide its exploration with expert trajectories supplied by the search algorithm. 

For the pre-training stage, we run the search to completion on small graph pairs (thus, the exact MCS solution is found), and use a supervised mse loss function to replace the DQN loss function. The overall loss function is $(y_t-Q(s_t,a_t))^2$ where $y_t$ the target for iteration $t$ and $Q(s_t,a_t)$ is the predicted score. In this case, $y_t$ denotes the remaining size of the largest common subgraph starting from $s_t$ to its leaf node in the current branch of the search tree. 


For larger graph pairs though, finding the true target becomes too slow. In that case, after pre-training, we enter the imitation learning stage where we follow the expert trajectories provided by \mcsp instead of its own predicted $Q(s_t,a_t)$, to incorporate more trustworthy policy decisions into the training signal.
More details can be in found in the supplementary material. 

\section{Experiments}
\label{sec-exp}

We evaluate \mcsrlmodel against two state-of-the-art exact MCS detection algorithms and a series of graph matching methods from various domains. 
We conduct experiments on 7 hundred-node medium-sized synthetic and real-world graph datasets, 8 thousand-node large real-world graph datasets, and 2 million-node very large real-world datasets, whose details can be found in the supplementary material. 
Among the different baseline models, we find no consistent trend. This indicates the difficulty of our task, as existing methods cannot find a consistent policy that guarantees good performance on datasets from different domains. Our model can substantially outperform the baselines, highlighting the significance of our contributions to learning for search. 

\subsection{Baseline Methods}
\label{subsec-baseline}

There are two groups of methods: Exact MCS algorithms including \mcsp
and \mcsprl,
learning based graph matching models including \gwqap,
\pca,
and \mcstreemodel.

All the methods either originally use or are adapted to use the branch and bound search framework in Section~\ref{subsec-search} with differences in node pair selection policy and training strategies. 
During testing, we apply the trained model on all testing graph pairs. 
We give a budget of 500 and 7500 search iterations for medium-size and large graph pairs. For each of the two million-node graph pairs, since the true MCS is much larger, we run each method for 50 minutes and plot the subgraph size growth across time.
Due to the search algorithm, all the methods can find exact MCS solutions given long enough budget, albeit an unrealistic assumption in practice for large graph pairs.


To validate the usefulness of the learned DQN, we compare \mcsrlmodel, our full model, with a randomly initialized model, \randq, which replaces the output of our DQN with a completely random scalar. 

\subsection{Hyperparameter Settings} 
For \pca, \mcstreemodel and \mcsrlmodel, we utilize 3 layers of Graph Attention Networks (GAT)~\citep{velickovic2017graph} each with 64 dimensions for the embeddings. The initial node embedding is encoded using the local degree profile~\citep{cai2018simple}. We use $\mathrm{ELU}(x) = \alpha (\mathrm{exp}(x) – 1)$ for $x \le 0$ and $x$ for $x > 0$ as our activation function where $\alpha=1$. 
We run all experiments with Intel i7-6800K CPU and one Nvidia Titan GPU. For DQN, we use MLP layers to project concatenated embeddings to a scalar. We use $\textsc{sum}$ followed by an MLP for $\textsc{readout}$ and $\textsc{1dconv+maxpool}$ followed by an MLP for $\textsc{interact}$. 
Further details can be found in supplementary material.
For training, we set the learning rate to 0.001, the number of training iterations to 10000, and use the Adam optimizer~\citep{kingma2014adam}. The models were implemented with the PyTorch and PyTorch Geometric libraries~\citep{Fey/Lenssen/2019}.

\begin{table*}
\footnotesize
  \begin{center}
    \caption{Results on medium graphs. Each synthetic dataset consists of 50 randomly generated pairs labeled as ``$\langle$generation algorithm$\rangle$-$\langle$number of nodes in each graph$\rangle$''. ``BA'', ``ER'', and ``WS'' refer to the Barab{\'a}si-Albert (BA)~\citep{barabasi1999emergence}, the Erd\H{o}s-R{\'e}nyi (ER)~\citep{gilbert1959random}, and the Watts–Strogatz (WS)~\citep{watts1998collective} algorithms, respectively. \nci consists of 100 chemical compound graph pairs whose average graph size is 28.73. We show the ratio of the (average) size of the subgraphs found by each method with respect to the best result on that dataset.
    }
    \begin{tabular}{l|lllllll}
    \label{table-acc-syn}
    \textbf{Method} &
    \textbf{BA-50} &
    \textbf{BA-100} &
    \textbf{ER-50} &
    \textbf{ER-100} &
    \textbf{WS-50} &
    \textbf{WS-100} & 
    \textbf{\nci} \\
      \hline
    \mcsp         & 0.913	& 0.892	& 0.842	& 0.896	& 0.905	& 0.856 & 0.948 \\
    \mcsprl       & 0.923	& 0.857	& 0.844	& 0.877	& 0.913	& 0.875 & 0.948 \\
    \hline
    \gwqap        & 0.945	& 0.887	& 0.855	& 0.925	& 0.916	& 0.898 & 0.966 \\
    \pca          & 0.899	& 0.863	& 0.848	& 0.923	& 0.879	& 0.852 & 0.951 \\
    \mcstreemodel & 0.908	& 0.889	& 0.846	& 0.906	& 0.889	& 0.865 & 0.954 \\
    \hline
    \randq        & 0.995   & 0.987 & 0.920 & 0.978 & 0.967 & 0.931 & 0.989 \\
    \mcsrlmodel   & \textbf{1.000}	& \textbf{1.000}	& \textbf{1.000}	& \textbf{1.000}	& \textbf{1.000}	& \textbf{1.000} & \textbf{1.000} \\
    \hline
    \hline
    \nnodes & 19.12 & 34.38 & 26.56 & 37.64 & 29.48 & 55.56 & 10.48 \\
    \end{tabular}
  \end{center}
\vspace*{-4mm}
\end{table*}

\vspace{0.5cm}

\begin{table*}
\small
  \begin{center}
    \caption{Results on real-world large graph pairs. Each dataset consists of one large real graph pair ($\ga$, $\gb$ may not be isomorphic, but $\gas$, $\gbs$ are isomorphic guaranteed by search). Below each dataset name, we show its size $\min(|\mathcal{V}_1|,|\mathcal{V}_2|)$ to indicate these pairs are significantly larger than the ones in Table~\ref{table-acc-syn}. Consistent with Table~\ref{table-acc-syn}, we show the ratio of the subgraph sizes. 
    }


    \begin{tabular}{l|llllllll}
    \label{table-acc}
    \multirow{2}{*}{\textbf{Method}} &
    \textbf{\road} &
    \textbf{\kgen} &
    \textbf{\kgzh} &
    \textbf{\kgenzh} &
    \textbf{\enron} &
    \textbf{\amazon} &
    \textbf{\circuit} &
    \textbf{\ppi} \\
    & 652 & 1945 & 1907 & 1907 & 3369 & 3518 & 4275 & 2152 \\
      \hline
      
    \mcsp         & 0.374 & 0.815 & 0.797 & 0.722 & 0.694 & 0.684 & 0.498 & 0.864\\
    \mcsprl       & 0.771 & 0.699 & 0.589 & 0.434 & 0.742 & 0.674 & 0.583 & 0.787 \\
    \hline
    \gwqap        & 0.305 & 0.929 & 0.855 & 0.808 & 0.711 & 0.860 & 0.354 & 0.834\\
    \pca          & 0.267 & 0.551 & 0.589 & 0.607 & 0.650 & 0.707 & 0.203 & 0.762 \\
    \mcstreemodel & 0.977 & 0.785 & 0.616 & 0.620 & 0.737 & 0.742 & 0.561 & 0.785 \\
    \hline
    \randq        & 0.641 & 0.762 & 0.658 & 0.639 & 0.814 & 0.755 & 0.603 & 0.814 \\
    \mcsrlmodel   & \textbf{1.000} & \textbf{1.000} & \textbf{1.000} & \textbf{1.000} & \textbf{1.000} & \textbf{1.000} & \textbf{1.000} & \textbf{1.000} \\
    \hline
    \hline
    \nnodes & 131 & 508 & 482 & 521 & 543 & 791 & 3515 & 404 \\
    \end{tabular}
  \end{center}
\vspace*{-4mm}
\end{table*}

\subsection{Results}
\label{subsec-result}



The key property of \mcsrlmodel is its ability to find the best solution in the fewest number of search iterations. As shown in Table~\ref{table-acc-syn}, our model outperforms baselines in terms of size of extracted subgraphs on all medium-sized synthetic graph datasets and the chemical compound dataset \nci. 

Regarding large real-world graphs, as shown in Table~\ref{table-acc}, our model outperforms baselines in terms of the size of the extracted subgraphs on all datasets. The exact solvers rely on heuristics for node selection, and consistently find much smaller subgraphs compared to our results.

\begin{figure*}[h]
     \centering
     \subfloat[Result on \roadca with 978513 nodes.]{\includegraphics[width=0.47\textwidth]{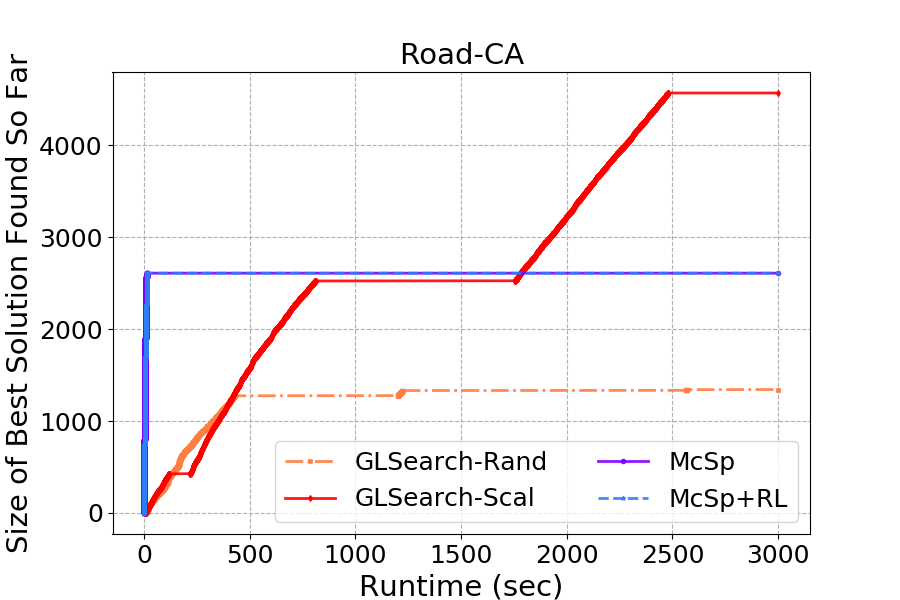}}
     \subfloat[Result on \roadtx with 1080909 nodes.]{\includegraphics[width=0.47\textwidth]{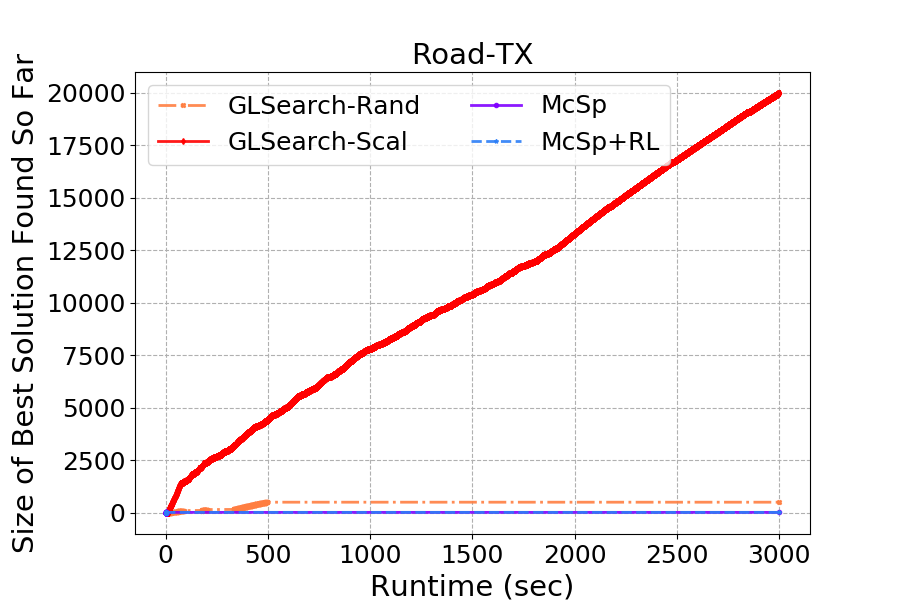}}
     \caption{Comparison of the best solution sizes of different methods on two million-node graph pairs, \roadca and \roadtx. \gwqap, \pca, and \mcstreemodel encounter memory error on these graph pairs due to their computation of a quadratic node-node matching matrix}.
     \label{fig-overhead}
\vspace*{-4mm}
\end{figure*}

Compared with learning based graph matching models, \mcsrlmodel is the only model which learns a reward that is dependent on both state and action, i.e. $Q(s_t,a_t)$. \gwqap, \pca, and \mcstreemodel essentially pre-compute the matching scores for all the node pairs in the input graphs, and therefore at each search step, the scores cannot adapt to the particular state, i.e. the matching scores only depend on $\ga,\gb$. Notice our state representation includes $\ga,\gb$ as well, hence \mcsrlmodel has more representational power than baselines. Trained under a reinforcement learning framework guided by search, \mcsrlmodel also performs the best among learning based baselines.

\begin{table*}
\small
  \begin{center}
    \caption{Ablation study on large real-world datasets. We demonstrate our Q function design choices indeed solve the various shortcomings presented in Section~\ref{subsec-policy}, through better representation learning.}
    
    \begin{tabular}{l|llllllll}
    \label{table-abl}
    \textbf{Method}&
    \textbf{\road} &
    \textbf{\kgen} &
    \textbf{\kgzh} &
    \textbf{\kgenzh} &
    \textbf{\enron} &
    \textbf{\amazon} &
    \textbf{\circuit} &
    \textbf{\ppi} \\
    \hline
\mcsrlmodel (no $\bm{h}_{\mathcal{G}}$) & 0.977 & 0.878 & 0.925 & 0.845 & 0.860 & 0.987 & 0.980 & 0.960\\ 
\mcsrlmodel (no $\bm{h}_{s}$) & \textbf{1.000} & 0.874 & 0.894 & 0.869 & 0.928 & \textbf{1.000} & 0.801 & 0.913\\ 
\mcsrlmodel (no $\bm{h}_{\mathcal{D}c}$) & 0.803 & 0.780 & 0.687 & 0.818 & 0.740 & 0.804 & 0.505 & 0.849\\ 
\mcsrlmodel (no $\bm{h}_{\mathcal{D}0}$) & 0.576 & 0.856 & 0.782 & 0.768 & 0.823 & 0.932 & 0.323 & 0.938\\ 
    \hline
\mcsrlmodel ($\textsc{sum}$ interact) & 0.902 & 0.913 & 0.963 & 0.885 & 0.899 & 0.957 & \textbf{1.000} & 0.948\\ 
    \hline
\mcsrlmodel (unfactored) &  0.447 & 0.807 & 0.712 & 0.582 & 0.816 & 0.816 & 0.512 & 0.861\\ 
\mcsrlmodel (unfactored-i) & 0.500 & 0.789 & 0.741 & 0.772 & 0.748 & 0.825 & 0.902 & 0.864\\ 
    \hline
\mcsrlmodel & 0.992 & \textbf{1.000} & \textbf{1.000} & \textbf{1.000} & \textbf{1.000} & 0.990 & 0.881 & \textbf{1.000}\\ 
    \hline
    \hline
\nnodes & 132 & 508 & 482 & 521 & 543 & 799 & 3989 & 404

    \end{tabular}
  \end{center}
\vspace*{-4mm}
\end{table*}

\subsection{Million-Node Graph Pairs}
\mcsrlmodel can scale to very large graph pairs, the limit of which is only bounded by the scalability of the GNN embedding step. 
To demonstrate this, we run \mcsrlmodel on million-node real-world graph datasets, \roadca and \roadtx.
To fit the model onto our GPU resources, we construct a lighter version of \mcsrlmodel, called \mcsrlmodelscl, which reduces the GAT encoder dimensions from 64 to 16.

As shown in Figure~\ref{fig-overhead}, \mcsrlmodel significantly outperforms baseline solvers on the two million-node real-world datasets. 
On \roadtx, \mcsp and \mcsprl perform poorly (getting ``stuck'' in local optimum as pointed out in Section~\ref{subsec-search}) while \mcsrlmodel continues to find larger and larger common subgraph after 50 minutes.

\subsection{Ablation Study}



To evaluate the effectiveness of different components proposed in our DQN model, we run ablation studies on the 8 large real world datasets.

We first measure the importance of each embedding vector fed to our DQN module, as described by Equation~\ref{eq:dvn_final}. We remove each embedding vector (specifically: $\bm{h}_{\mathcal{G}} =\textsc{interact}(\bm{h}_{\mathcal{G}_1},\bm{h}_{\mathcal{G}_2})$, $\bm{h}_{s} =\textsc{interact}(\bm{h}_{s1},\bm{h}_{s2})$, $\bm{h}_{\mathcal{D}c}$, and $\bm{h}_{D_{0}}$) individually from the DQN model and retrain the model under the same training settings. Table~\ref{table-abl} is consistent with our conclusion that every embedding vector used by \mcsrlmodel is critical in capturing the search state's representation. Furthermore, we find leveraging bidomain representations
is very beneficial to our model.

We next measure the importance of interaction to address the symmetry issue of the MCS calculation, where input graph pairs must be order insensitive. We first test the necessity of using more complex interaction functions, by replacing our $\textsc{1dconv+maxpool}$ interaction with simple $\textsc{sum}$ for interaction (still followed by an MLP). As shown in Table~\ref{table-abl}, we see that simpler interaction functions may not be powerful enough to encode the interaction between 2 graphs. Particularly, this suggests that interaction is quite important to model performance. 

Finally, we measure the importance of factoring out actions from our DQN model. We test this with 2 models. The first utilizes Equation~\ref{eq:dqn} to encode the Q-value, which we refer to as \mcsrlmodel (unfactored). Since Equation~\ref{eq:dqn} also suffers from the issue of graph symmetry, we adapt this model to use the same interaction function as \mcsrlmodel to construct 3 order-invariant embeddings $\bm{h}_{\mathcal{G}}=\textsc{interact}(\bm{h}_{\mathcal{G}_1},\bm{h}_{\mathcal{G}_2})$, $h_{s}=\textsc{interact}(\bm{h}_{s1},\bm{h}_{s2})$, $\bm{h}_{a}=\textsc{interact}(\bm{h}_{i},\bm{h}_{j})$ to concatenate and pass to the final MLP layer in Equation~\ref{eq:dqn}. We refer to this model as \mcsrlmodel (unfactored-i). Our results show that without factoring out the action, our performance is comparable to or worse than \mcsp, indicating the significant performance boost introduced by maximally reflecting the effect of adding node pairs. 
\section{Conclusion}
\label{sec-conc}

We believe the interplay of search and learning is a promising research direction, and take a step towards bridging the gap by tackling the NP-hard challenging task, Maximum Common Subgraph detection. We have proposed a reinforcement learning method which unifies search and deep Q-learning into a single framework. By using the search to train our carefully designed DQN, the DQN provides better node selection policy for search to find large common subgraph solutions faster, which is experimentally verified on synthetic and real-world large graph pairs. In the future, we will explore the adaptation of our framework which combines learning with search to other constrained combinatorial problems, e.g. Maximum Clique Detection.

\clearpage
\bibliography{bibliography}

\begin{thebibliography}{57}
\providecommand{\natexlab}[1]{#1}
\providecommand{\url}[1]{\texttt{#1}}
\expandafter\ifx\csname urlstyle\endcsname\relax
  \providecommand{\doi}[1]{doi: #1}\else
  \providecommand{\doi}{doi: \begingroup \urlstyle{rm}\Url}\fi

\bibitem[Agrawal et~al.(2018)Agrawal, Zitnik, Leskovec,
  et~al.]{agrawal2018large}
Agrawal, M., Zitnik, M., Leskovec, J., et~al.
\newblock Large-scale analysis of disease pathways in the human interactome.
\newblock In \emph{PSB}, pp.\  111--122. World Scientific, 2018.

\bibitem[Bahiense et~al.(2012)Bahiense, Mani{\'c}, Piva, and
  De~Souza]{bahiense2012maximum}
Bahiense, L., Mani{\'c}, G., Piva, B., and De~Souza, C.~C.
\newblock The maximum common edge subgraph problem: A polyhedral investigation.
\newblock \emph{Discrete Applied Mathematics}, 160\penalty0 (18):\penalty0
  2523--2541, 2012.

\bibitem[Bai et~al.(2019)Bai, Ding, Bian, Chen, Sun, and Wang]{bai2018graph}
Bai, Y., Ding, H., Bian, S., Chen, T., Sun, Y., and Wang, W.
\newblock Simgnn: A neural network approach to fast graph similarity
  computation.
\newblock \emph{WSDM}, 2019.

\bibitem[Bai et~al.(2020{\natexlab{a}})Bai, Ding, Gu, , Sun, and
  Wang]{bai2020convolutional}
Bai, Y., Ding, H., Gu, K., , Sun, Y., and Wang, W.
\newblock Learning-based efficient graph similarity computation via multi-scale
  convolutional set matching.
\newblock \emph{AAAI}, 2020{\natexlab{a}}.

\bibitem[Bai et~al.(2020{\natexlab{b}})Bai, Xu, Gu, Wu, Marinovic, Ro, Sun, and
  Wang]{bai2020neural}
Bai, Y., Xu, D., Gu, K., Wu, X., Marinovic, A., Ro, C., Sun, Y., and Wang, W.
\newblock Neural maximum common subgraph detection with guided subgraph
  extraction, 2020{\natexlab{b}}.
\newblock URL \url{https://openreview.net/forum?id=BJgcwh4FwS}.

\bibitem[Balcilar et~al.(2021)Balcilar, Renton, H{\'e}roux, Ga{\"u}z{\`e}re,
  Adam, and Honeine]{balcilar2021analyzing}
Balcilar, M., Renton, G., H{\'e}roux, P., Ga{\"u}z{\`e}re, B., Adam, S., and
  Honeine, P.
\newblock Analyzing the expressive power of graph neural networks in a spectral
  perspective.
\newblock In \emph{ICLR}, 2021.
\newblock URL \url{https://openreview.net/forum?id=-qh0M9XWxnv}.

\bibitem[Barab{\'a}si \& Albert(1999)Barab{\'a}si and
  Albert]{barabasi1999emergence}
Barab{\'a}si, A.-L. and Albert, R.
\newblock Emergence of scaling in random networks.
\newblock \emph{science}, 286\penalty0 (5439):\penalty0 509--512, 1999.

\bibitem[Bastian et~al.(2009)Bastian, Heymann, and Jacomy]{bastian2009gephi}
Bastian, M., Heymann, S., and Jacomy, M.
\newblock Gephi: an open source software for exploring and manipulating
  networks.
\newblock In \emph{Proceedings of the International AAAI Conference on Web and
  Social Media}, volume~3, 2009.

\bibitem[Bello et~al.(2017)Bello, Pham, Le, Norouzi, and
  Bengio]{bello2016neural}
Bello, I., Pham, H., Le, Q.~V., Norouzi, M., and Bengio, S.
\newblock Neural combinatorial optimization with reinforcement learning.
\newblock \emph{ICLR}, 2017.

\bibitem[Bengio et~al.(2009)Bengio, Louradour, Collobert, and
  Weston]{bengio2009curriculum}
Bengio, Y., Louradour, J., Collobert, R., and Weston, J.
\newblock Curriculum learning.
\newblock In \emph{ICML}, pp.\  41--48, 2009.

\bibitem[Bunke \& Shearer(1998)Bunke and Shearer]{bunke1998graph}
Bunke, H. and Shearer, K.
\newblock A graph distance metric based on the maximal common subgraph.
\newblock \emph{Pattern recognition letters}, 19\penalty0 (3-4):\penalty0
  255--259, 1998.

\bibitem[Cai \& Wang(2018)Cai and Wang]{cai2018simple}
Cai, C. and Wang, Y.
\newblock A simple yet effective baseline for non-attributed graph
  classification.
\newblock \emph{arXiv preprint arXiv:1811.03508}, 2018.

\bibitem[Cao et~al.(2011)Cao, Yang, Wang, Ren, and Lou]{cao2011privacy}
Cao, N., Yang, Z., Wang, C., Ren, K., and Lou, W.
\newblock Privacy-preserving query over encrypted graph-structured data in
  cloud computing.
\newblock In \emph{2011 31st International Conference on Distributed Computing
  Systems}, pp.\  393--402. IEEE, 2011.

\bibitem[Dai et~al.(2017)Dai, Khalil, Zhang, Dilkina, and
  Song]{dai2017learning}
Dai, H., Khalil, E.~B., Zhang, Y., Dilkina, B., and Song, L.
\newblock Learning combinatorial optimization algorithms over graphs.
\newblock \emph{NeurIPS}, 2017.

\bibitem[Dai et~al.(2019)Dai, Li, Wang, Singh, Huang, and
  Kohli]{dai2019learning}
Dai, H., Li, Y., Wang, C., Singh, R., Huang, P.-S., and Kohli, P.
\newblock Learning transferable graph exploration.
\newblock \emph{NeurIPS}, 2019.

\bibitem[Debnath et~al.(1991)Debnath, Lopez~de Compadre, Debnath, Shusterman,
  and Hansch]{debnath1991structure}
Debnath, A.~K., Lopez~de Compadre, R.~L., Debnath, G., Shusterman, A.~J., and
  Hansch, C.
\newblock Structure-activity relationship of mutagenic aromatic and
  heteroaromatic nitro compounds. correlation with molecular orbital energies
  and hydrophobicity.
\newblock \emph{Journal of medicinal chemistry}, 34\penalty0 (2):\penalty0
  786--797, 1991.

\bibitem[Duesbury et~al.(2018)Duesbury, Holliday, and
  Willett]{duesbury2018comparison}
Duesbury, E., Holliday, J., and Willett, P.
\newblock Comparison of maximum common subgraph isomorphism algorithms for the
  alignment of 2d chemical structures.
\newblock \emph{ChemMedChem}, 13\penalty0 (6):\penalty0 588--598, 2018.

\bibitem[Ehrlich \& Rarey(2011)Ehrlich and Rarey]{ehrlich2011maximum}
Ehrlich, H.-C. and Rarey, M.
\newblock Maximum common subgraph isomorphism algorithms and their applications
  in molecular science: a review.
\newblock \emph{Wiley Interdisciplinary Reviews: Computational Molecular
  Science}, 1\penalty0 (1):\penalty0 68--79, 2011.

\bibitem[Fan et~al.(2020)Fan, Zeng, Sun, and Liu]{fan2020finding}
Fan, C., Zeng, L., Sun, Y., and Liu, Y.-Y.
\newblock Finding key players in complex networks through deep reinforcement
  learning.
\newblock \emph{Nature Machine Intelligence}, 2\penalty0 (6):\penalty0
  317--324, 2020.

\bibitem[Fey \& Lenssen(2019)Fey and Lenssen]{Fey/Lenssen/2019}
Fey, M. and Lenssen, J.~E.
\newblock Fast graph representation learning with {PyTorch Geometric}.
\newblock In \emph{ICLR Workshop on Representation Learning on Graphs and
  Manifolds}, 2019.

\bibitem[Gilbert(1959)]{gilbert1959random}
Gilbert, E.~N.
\newblock Random graphs.
\newblock \emph{The Annals of Mathematical Statistics}, 30\penalty0
  (4):\penalty0 1141--1144, 1959.

\bibitem[Kingma \& Ba(2015)Kingma and Ba]{kingma2014adam}
Kingma, D.~P. and Ba, J.
\newblock Adam: A method for stochastic optimization.
\newblock \emph{ICLR}, 2015.

\bibitem[Kipf \& Welling(2016)Kipf and Welling]{kipf2016semi}
Kipf, T.~N. and Welling, M.
\newblock Semi-supervised classification with graph convolutional networks.
\newblock \emph{ICLR}, 2016.

\bibitem[Klimt \& Yang(2004)Klimt and Yang]{klimt2004introducing}
Klimt, B. and Yang, Y.
\newblock Introducing the enron corpus.
\newblock In \emph{CEAS}, 2004.

\bibitem[Lee \& Schachter(1980)Lee and Schachter]{lee1980two}
Lee, D.-T. and Schachter, B.~J.
\newblock Two algorithms for constructing a delaunay triangulation.
\newblock \emph{International Journal of Computer \& Information Sciences},
  9\penalty0 (3):\penalty0 219--242, 1980.

\bibitem[Leskovec et~al.(2009)Leskovec, Lang, Dasgupta, and
  Mahoney]{leskovec2009community}
Leskovec, J., Lang, K.~J., Dasgupta, A., and Mahoney, M.~W.
\newblock Community structure in large networks: Natural cluster sizes and the
  absence of large well-defined clusters.
\newblock \emph{Internet Mathematics}, 6\penalty0 (1):\penalty0 29--123, 2009.

\bibitem[Levi(1973)]{levi1973note}
Levi, G.
\newblock A note on the derivation of maximal common subgraphs of two directed
  or undirected graphs.
\newblock \emph{Calcolo}, 9\penalty0 (4):\penalty0 341, 1973.

\bibitem[Li et~al.(2019)Li, Gu, Dullien, Vinyals, and Kohli]{li2019graph}
Li, Y., Gu, C., Dullien, T., Vinyals, O., and Kohli, P.
\newblock Graph matching networks for learning the similarity of graph
  structured objects.
\newblock \emph{ICML}, 2019.

\bibitem[Ling et~al.(2020)Ling, Wu, Wang, Ma, Xu, Wu, and
  Ji]{ling2020hierarchical}
Ling, X., Wu, L., Wang, S., Ma, T., Xu, F., Wu, C., and Ji, S.
\newblock Hierarchical graph matching networks for deep graph similarity
  learning, 2020.
\newblock URL \url{https://openreview.net/forum?id=rkeqn1rtDH}.

\bibitem[Liu et~al.(2019)Liu, Li, Jiang, and He]{liu2019learning}
Liu, Y.-l., Li, C.-m., Jiang, H., and He, K.
\newblock A learning based branch and bound for maximum common subgraph
  problems.
\newblock \emph{IJCAI}, 2019.

\bibitem[McCreesh et~al.(2016)McCreesh, Ndiaye, Prosser, and
  Solnon]{mccreesh2016clique}
McCreesh, C., Ndiaye, S.~N., Prosser, P., and Solnon, C.
\newblock Clique and constraint models for maximum common (connected) subgraph
  problems.
\newblock In \emph{International Conference on Principles and Practice of
  Constraint Programming}, pp.\  350--368. Springer, 2016.

\bibitem[McCreesh et~al.(2017)McCreesh, Prosser, and
  Trimble]{mccreesh2017partitioning}
McCreesh, C., Prosser, P., and Trimble, J.
\newblock A partitioning algorithm for maximum common subgraph problems.
\newblock 2017.

\bibitem[Mnih et~al.(2013)Mnih, Kavukcuoglu, Silver, Graves, Antonoglou,
  Wierstra, and Riedmiller]{mnih2013playing}
Mnih, V., Kavukcuoglu, K., Silver, D., Graves, A., Antonoglou, I., Wierstra,
  D., and Riedmiller, M.
\newblock Playing atari with deep reinforcement learning.
\newblock \emph{NeurIPS Deep Learning Workshop 2013}, 2013.

\bibitem[Mnih et~al.(2015)Mnih, Kavukcuoglu, Silver, Rusu, Veness, Bellemare,
  Graves, Riedmiller, Fidjeland, Ostrovski, et~al.]{mnih2015human}
Mnih, V., Kavukcuoglu, K., Silver, D., Rusu, A.~A., Veness, J., Bellemare,
  M.~G., Graves, A., Riedmiller, M., Fidjeland, A.~K., Ostrovski, G., et~al.
\newblock Human-level control through deep reinforcement learning.
\newblock \emph{Nature}, 518\penalty0 (7540):\penalty0 529--533, 2015.

\bibitem[Park et~al.(2013)Park, Reeves, and Stamp]{park2013deriving}
Park, Y., Reeves, D.~S., and Stamp, M.
\newblock Deriving common malware behavior through graph clustering.
\newblock \emph{Computers \& Security}, 39:\penalty0 419--430, 2013.

\bibitem[Peyr{\'e} et~al.(2016)Peyr{\'e}, Cuturi, and Solomon]{peyre2016gromov}
Peyr{\'e}, G., Cuturi, M., and Solomon, J.
\newblock Gromov-wasserstein averaging of kernel and distance matrices.
\newblock In \emph{ICML}, pp.\  2664--2672, 2016.

\bibitem[Riesen \& Bunke(2008)Riesen and Bunke]{riesen2008iam}
Riesen, K. and Bunke, H.
\newblock Iam graph database repository for graph based pattern recognition and
  machine learning.
\newblock In \emph{Joint IAPR International Workshops on Statistical Techniques
  in Pattern Recognition (SPR) and Structural and Syntactic Pattern Recognition
  (SSPR)}, pp.\  287--297. Springer, 2008.

\bibitem[Schietgat et~al.(2013)Schietgat, Ramon, and
  Bruynooghe]{schietgat2013polynomial}
Schietgat, L., Ramon, J., and Bruynooghe, M.
\newblock A polynomial-time maximum common subgraph algorithm for outerplanar
  graphs and its application to chemoinformatics.
\newblock \emph{Annals of Mathematics and Artificial Intelligence}, 69\penalty0
  (4):\penalty0 343--376, 2013.

\bibitem[Shchur et~al.(2018)Shchur, Mumme, Bojchevski, and
  G{\"u}nnemann]{shchur2018pitfalls}
Shchur, O., Mumme, M., Bojchevski, A., and G{\"u}nnemann, S.
\newblock Pitfalls of graph neural network evaluation.
\newblock \emph{Relational Representation Learning Workshop (R2L 2018), NeurIPS
  2018}, 2018.

\bibitem[Shrivastava \& Li(2014)Shrivastava and Li]{shrivastava2014new}
Shrivastava, A. and Li, P.
\newblock A new space for comparing graphs.
\newblock In \emph{Proceedings of the 2014 IEEE/ACM International Conference on
  Advances in Social Networks Analysis and Mining}, pp.\  62--71. IEEE Press,
  2014.

\bibitem[Solozabal et~al.(2020)Solozabal, Ceberio, and
  Tak{\'a}{\v{c}}]{solozabal2020constrained}
Solozabal, R., Ceberio, J., and Tak{\'a}{\v{c}}, M.
\newblock Constrained combinatorial optimization with reinforcement learning.
\newblock \emph{arXiv preprint arXiv:2006.11984}, 2020.

\bibitem[Sun et~al.(2017)Sun, Hu, and Li]{sun2017cross}
Sun, Z., Hu, W., and Li, C.
\newblock Cross-lingual entity alignment via joint attribute-preserving
  embedding.
\newblock In \emph{International Semantic Web Conference}, pp.\  628--644.
  Springer, 2017.

\bibitem[Velickovic et~al.(2018)Velickovic, Cucurull, Casanova, Romero, Lio,
  and Bengio]{velickovic2017graph}
Velickovic, P., Cucurull, G., Casanova, A., Romero, A., Lio, P., and Bengio, Y.
\newblock Graph attention networks.
\newblock \emph{ICLR}, 2018.

\bibitem[Vismara \& Valery(2008)Vismara and Valery]{vismara2008finding}
Vismara, P. and Valery, B.
\newblock Finding maximum common connected subgraphs using clique detection or
  constraint satisfaction algorithms.
\newblock In \emph{International Conference on Modelling, Computation and
  Optimization in Information Systems and Management Sciences}, pp.\  358--368.
  Springer, 2008.

\bibitem[Wale et~al.(2008)Wale, Watson, and Karypis]{wale2008comparison}
Wale, N., Watson, I.~A., and Karypis, G.
\newblock Comparison of descriptor spaces for chemical compound retrieval and
  classification.
\newblock \emph{Knowledge and Information Systems}, 14\penalty0 (3):\penalty0
  347--375, 2008.

\bibitem[Wang et~al.(2019)Wang, Yan, and Yang]{wang2019learning}
Wang, R., Yan, J., and Yang, X.
\newblock Learning combinatorial embedding networks for deep graph matching.
\newblock \emph{ICCV}, 2019.

\bibitem[Wang et~al.(2012)Wang, Ding, Tung, Ying, and Jin]{wang2012efficient}
Wang, X., Ding, X., Tung, A.~K., Ying, S., and Jin, H.
\newblock An efficient graph indexing method.
\newblock In \emph{ICDE}, pp.\  210--221. IEEE, 2012.

\bibitem[Watts \& Strogatz(1998)Watts and Strogatz]{watts1998collective}
Watts, D.~J. and Strogatz, S.~H.
\newblock Collective dynamics of ‘small-world’networks.
\newblock \emph{nature}, 393\penalty0 (6684):\penalty0 440, 1998.

\bibitem[Xu et~al.(2019{\natexlab{a}})Xu, Luo, and Carin]{xu2019scalable}
Xu, H., Luo, D., and Carin, L.
\newblock Scalable gromov-wasserstein learning for graph partitioning and
  matching.
\newblock In \emph{NeurIPS}, pp.\  3046--3056, 2019{\natexlab{a}}.

\bibitem[Xu et~al.(2019{\natexlab{b}})Xu, Luo, Zha, and Carin]{xu2019gromov}
Xu, H., Luo, D., Zha, H., and Carin, L.
\newblock Gromov-wasserstein learning for graph matching and node embedding.
\newblock \emph{ICML}, 2019{\natexlab{b}}.

\bibitem[Xu et~al.(2019{\natexlab{c}})Xu, Hu, Leskovec, and
  Jegelka]{xu2018powerful}
Xu, K., Hu, W., Leskovec, J., and Jegelka, S.
\newblock How powerful are graph neural networks?
\newblock \emph{ICLR}, 2019{\natexlab{c}}.

\bibitem[Yan et~al.(2005)Yan, Yu, and Han]{yan2005substructure}
Yan, X., Yu, P.~S., and Han, J.
\newblock Substructure similarity search in graph databases.
\newblock In \emph{SIGMOD}, pp.\  766--777. ACM, 2005.

\bibitem[Yanardag \& Vishwanathan(2015)Yanardag and
  Vishwanathan]{yanardag2015deep}
Yanardag, P. and Vishwanathan, S.
\newblock Deep graph kernels.
\newblock In \emph{SIGKDD}, pp.\  1365--1374. ACM, 2015.

\bibitem[You et~al.(2019)You, Ying, and Leskovec]{you2019position}
You, J., Ying, R., and Leskovec, J.
\newblock Position-aware graph neural networks.
\newblock In \emph{ICML}, pp.\  7134--7143. PMLR, 2019.

\bibitem[Yu et~al.(2020)Yu, Wang, Yan, and Li]{Yu2020Learning}
Yu, T., Wang, R., Yan, J., and Li, B.
\newblock Learning deep graph matching with channel-independent embedding and
  hungarian attention.
\newblock In \emph{ICLR}, 2020.
\newblock URL \url{https://openreview.net/forum?id=rJgBd2NYPH}.

\bibitem[Zanfir \& Sminchisescu(2018)Zanfir and Sminchisescu]{zanfir2018deep}
Zanfir, A. and Sminchisescu, C.
\newblock Deep learning of graph matching.
\newblock In \emph{CVPR}, pp.\  2684--2693, 2018.

\bibitem[Zeng et~al.(2009)Zeng, Tung, Wang, Feng, and Zhou]{zeng2009comparing}
Zeng, Z., Tung, A.~K., Wang, J., Feng, J., and Zhou, L.
\newblock Comparing stars: On approximating graph edit distance.
\newblock \emph{PVLDB}, 2\penalty0 (1):\penalty0 25--36, 2009.

\end{thebibliography}
\bibliographystyle{icml2021}

\setcounter{section}{0} 
\clearpage
\newpage
\section*{{\Large Supplementary Material}}

\def\thesection{\Alph{section}}

\section{Insights and Contributions of \mcsrlmodel}
\label{sec-insight}

\textbf{To Search community on MCS detection} \enspace
The major challenge that prevents existing search algorithms from extracting large common subgraphs for large input graph pairs is that the focus of these algorithms is on reducing the \textbf{overall} search space rather than making smarter node pair selections in \textbf{each} search step, as shown in Section~\ref{subsec-vis-big}. By improving the order it searches candidate solutions, \mcsrlmodel can quickly find \textbf{\textit{better}} MCS candidates, without much (or any) backtracking and pruning, than state-of-the-art search algorithms. 

\textbf{To General Learning community} \enspace
\mcsrlmodel tackles the hard constraint that the subgraphs must be isomorphic to each other by only choosing actions from connected bidomains as illustrated in the main text. However, there is an additional advantage of introducing the bidomain concept illustrated in \mcsp: Bidomains partition the rest of the input graphs into different regions where future actions can be selected Thus, properly encoding of the bidomains gives more information about hard constraints to the DQN, which improves performance as experimentally verified by the ablation study in the main text. More generally, this shows that learning components can be further enriched by incorporating knowledge on tackling hard constraints of an NP-hard task, e.g. bidomain in our case into their model design.

\textbf{To Graph Deep Learning community} \enspace
Although various works have pointed out and analyzed the limitation of GNN's expressive power~\citep{xu2018powerful,balcilar2021analyzing}, for particular tasks such as MCS detection, GNNs can still be used if augmented properly. We aim to predict a $Q$ score for a state-action pair, using a DQN with two components, one component computing the local node embeddings using several layers of GNN, the other component combining embeddings at a different granularity, i.e. embeddings at the subgraph, whole-graph, and bidomain levels, to produce the final score. Overall our DQN design adopts a similar general principle as Position-Aware GNN~\citep{you2019position}, which allows a regular GNN to absorb information from non-local nodes (called ``anchor'' nodes which are randomly selected nodes globally). In essence, the DQN in \mcsrlmodel also enhances the existing GNN by leveraging non-local information.

\textbf{To Reinforcement Learning community} \enspace
We are aware of efforts in the RL community to tackle NP-hard problems, but they either focus on non-graph tasks, such as KnapSack~\citep{bello2016neural} and Job Shop Scheduling~\citep{solozabal2020constrained,dai2019learning}, or address single-graph NP-hard tasks without hard constraints, such as Minimum Vertex Cover~\citep{dai2017learning}, Graph Exploration~\citep{dai2019learning}, and Network Dismantling~\citep{,fan2020finding}. Fundamentally different from these works, MCS detection requires a graph pair as input, and we show how to properly encode such an input into states and actions. The subgraph isomorphism constraint of the task also sets us apart from the aforementioned graph tasks which we tackle via fully taking advantage of a key property of the task, i.e. bidomain, while in contrast, \citet{solozabal2020constrained} relies on penalty signals generated from constraint dissatisfaction in order to guide the agent to achieve feasible solutions for non-graph tasks.

\section{Dataset Description}
\label{sec-dataset}


This section describes the datasets used for evaluating our model and baselines. Section~\ref{subsec-cur-learning} describes the dataset we use for training \mcsrlmodel as well as baseline learning based graph matching models.


We use the following real-world datasets for evaluation:
\begin{itemize}
    \item \nci: It is a collection of small-sized chemical compounds \citep{wale2008comparison} whose nodes are labeled indicating atom type. We form 100 graph pairs from the dataset whose average graph size (number of nodes) is 28.73.
    \item \road: The graph is a road network of California whose nodes indicate intersections and endpoints and edges represent the roads connecting the intersections and endpoints~\citep{leskovec2009community}. The graph contains 1965206 nodes, from which we randomly sample a connected subgraph of around 0.05\% nodes twice to generate two subgraphs for the graph pair $\ga=(\mathcal{V}_1,\mathcal{E}_1)$ and $\gb=(\mathcal{V}_2,\mathcal{E}_2)$. 
    \item \kgen, \kgzh, and \kgenzh: It is a dataset originally used in a work on cross-lingual entity alignment~\citep{sun2017cross}. The dataset contains pairs of DBpedia knowledge graphs in different languages. For \kgen, we use the English knowledge graph and sample 10\% nodes twice to generate two graphs for our task. For \kgzh, we sample around 10\% nodes from the knowledge graph in Chinese. For \kgenzh, we sample once from the English graph to get $\ga$ and sample once from the Chinese graph to get $\gb$. Note that although the nodes have features, we do not use them because our task is more about graph structural matching rather than node semantic meanings, and leave the incorporation of continuous node initial representations as future work.
    \item \enron: The graph is an email communication network whose nodes represent email addresses and (undirected) edges represent at least one email sent between the addresses~\citep{klimt2004introducing}. From the total 36692 nodes, we sample around 10\% nodes to generate the graph pair.
    \item \amazon: An Amazon computer product network whose nodes represent goods and edges represent two goods frequently purchased together~\citep{shchur2018pitfalls}. The graph contains 703655 nodes from which we sample around 0.5\% to get the pair we use.
    \item \circuit: This is a graph pair 
    where each graph is a circuit diagram whose nodes represent devices/wires and edges represent the connecting relations between devices and wires. In other words, each node is either a device or a wire, and the entire graph is bipartite. The two graphs given are known to be \textbf{isomorphic}\footnote{Section~\ref{subsec-known-mcs} shows results on synthetic datasets where the MCS size lower bound known.} and we do not perform any sampling. Nodes have labels about the type of the device/wire. In real world, the successful matching of circuit layout diagrams is an essential process in circuit design verification.
    \item \ppi: It is a human protein-protein interaction network whose nodes represent proteins and edges represent physical interaction between proteins in a human cell~\citep{agrawal2018large}. From the 21557 nodes, we sample around ~10\% nodes to generate the pair used in experiments.
    \item \roadca: We use the same road network of California as the \road dataset, but this time, from the 1965206 nodes, we randomly sample a connected subgraph of around 50.0\% nodes twice to generate two subgraphs for the graph pair. 
    \item \roadtx: Similar to \roadca, but the road network is in Texas~\citep{leskovec2009community}. The graph contains 1379917 nodes, from which we randomly sample a connected subgraph of around 80\% nodes twice to generate two subgraphs for the graph pair. 

\end{itemize}
The details of all the graph pairs can be found in Table~\ref{table-test-data}.



\begin{table*}[h]
  \begin{center}
    \caption{Details of real-world graph pairs used in evaluating the performance of baseline methods and \mcsrlmodel.}
    \begin{tabular}{l|lllllllll}
    \label{table-test-data}
      \textbf{Name} & \textbf{Description} & $|\mathcal{V}_1|$ & $|\mathcal{V}_2|$ & $|\mathcal{E}_1|$ & $|\mathcal{E}_2|$ & $\frac{|\mathcal{E}_1|}{|\mathcal{V}_1|}$ & $\frac{|\mathcal{E}_2|}{|\mathcal{V}_2|}$ \\
      \hline
      \textbf{\road} & Road Network & 1114 & 652 & 1454 & 822 & 1.305 & 1.261 \\
      \textbf{\kgen} & Knowledge Graph & 1945 & 1945 & 6242 & 5851 & 3.209 & 3.008 \\
      \textbf{\kgzh} & Knowledge Graph & 1907 & 1907 & 4856 & 4948 & 2.546 & 2.595 \\
      \textbf{\kgenzh} & Knowledge Graph & 1945 & 1907 & 6242 & 4856 & 3.209 & 2.546 \\
      \textbf{\enron} & Email Communication Network & 3369 & 3369 & 46399 & 50637 & 13.772 & 15.030 \\
      \textbf{\amazon} & Product Co-purchasing Network & 3518 & 3518 & 56028 & 40633 & 15.926 & 11.550 \\
      \textbf{\circuit} & Circuit Layout Diagram & 4275 & 4275 & 6128 & 6128 & 1.433 & 1.433 \\
      \textbf{\ppi} & Protein-Protein Interaction Network & 2152 & 2152 & 54910 & 54132 & 25.516 & 25.154 \\
      \textbf{\roadca} & Road Network & 978513 & 978513 & 1404115 & 1366917 & 1.435 & 1.397 \\
      \textbf{\roadtx} & Road Network & 1080909 & 1080909 & 1503531 & 1507440 & 1.391 & 1.395 \\
    \end{tabular}
  \end{center}
\end{table*}

For synthetic datasets, we generate graph pairs using the Barab{\'a}si-Albert (BA)~\citep{barabasi1999emergence} algorithm (edge density set to 5), the Erd\H{o}s-R{\'e}nyi (ER)~\citep{gilbert1959random} algorithm (edge density set to 0.08), and the Watts–Strogatz (WS)~\citep{watts1998collective} algorithm (rewiring probability set to 0.2 and ring density set to 4), respectively.

\section{Details on DQN and Training \mcsrlmodel}

\subsection{Training Data Preparation: Curriculum Learning}
\label{subsec-cur-learning}

Curriculum learning~\citep{bengio2009curriculum} is a strategy for training machine learning models whose core idea is to train a model first using ``easy'' examples before moving on to using ``hard'' ones. Our goal is to train a general model for MCS detection task which works well on general testing graph pairs from different domains. Therefore, we employ the idea of curriculum learning in training our \mcsrlmodel. More specifically, we prepare the training graph pairs in the following way:
\begin{itemize}
\item Curriculum 1: The first curriculum consists of the easiest graph pairs that are small: (1) We sample 30 graph pairs from \aids~\citep{zeng2009comparing}, a chemical compound dataset usually for graph similarity computation~\citep{bai2018graph} where each graph has less than or equal to 10 nodes; (2) We sample 30 graph pairs from \linux~\citep{wang2012efficient}, another dataset commonly used for graph matching consisting of small program dependency graphs generated from Linux kernel; (3) So far we have 60 real-world graph pairs. We then generate 60 graph pairs using popular graph generation algorithms. Specifcally, we generate 20 graph pairs using the BA algorithm, 20 graph pairs using the ER algorithm, and 20 graph pairs using the Watts–Strogatz WS algorithm, respectively. Details of the graphs can be found in Table~\ref{table-curriculum}. In summary, the first curriculum contains 120 graph pairs in total.
\item Curriculum 2: After the first curriculum, each next curriculum contains graphs that are larger and harder to match than the previous curriculum. For the second curriculum, we sample 30 graph pairs from \ptc~\citep{shrivastava2014new}, a collection of chemical compounds, 30 graph paris from \imdb~\citep{yanardag2015deep}, a collection of ego-networks of movie actors/actresses, and generate 20 graph pairs again using the BA, ER, and WS algorithms but with larger graph sizes.
\item Curriculum 3: For the third curriculum, we sample 30 graph pairs from \mutag~\citep{debnath1991structure}, a collection of chemical compounds, 30 graph paris from \reddit~\citep{yanardag2015deep}, a collection of ego-networks corresponding to online discussion threads, and generate 20 even larger graph pairs using the BA, ER, and WS algorithms.
\item Curriculum 4: For the last curriculum, we sample 30 graph pairs from \web~\citep{riesen2008iam}, a collection of text document graphs, 30 graph paris from \mcspconn~\citep{mccreesh2017partitioning}, a collection of synthetic graph pairs adopted by \mcsp, and generate 20 graph pairs again using BA, ER, and WS algorithms but with larger graph sizes.
\end{itemize}
For each curriculum, we train the model for 2500 iterations before moving on to the next, resulting in 10000 training iterations in total.

\begin{table}[h]
  \begin{center}
    \caption{Training graph details. For synthetic graphs, ``ed'', ``p'', and ``rd'' represent edge density, rewiring probability, and ring density, respectively.}
    \begin{tabular}{l|ll}
    \label{table-curriculum}
      \textbf{Curriculum} & \textbf{Data Source} & \textbf{\# Pairs} \\
      \hline
      \multirow{5}{*}{\textbf{Curriculum 1}} & 
      \aids & 30 \\
      & \linux & 30 \\
      & BA:n=16,ed=5 & 20 \\
      & ER:n=14,ed=0.14 & 20 \\
      & WS:n=18,p=0.2,rd=2 & 20 \\
      \hline
     \multirow{5}{*}{\textbf{Curriculum 2}} & 
      \ptc & 30 \\
      & \imdb & 30 \\
      & BA:n=32,ed=4 & 20 \\
      & ER:n=30,ed=0.12 & 20 \\
      & WS:n=34,p=0.2,rd=2 & 20 \\
      \hline
     \multirow{5}{*}{\textbf{Curriculum 3}} & 
      \mutag & 30 \\
      & \reddit & 30 \\
      & BA:n=48,ed=4 & 20 \\
      & ER:n=46,ed=0.1 & 20 \\
      & WS:n=50,p=0.2,rd=4 & 20 \\
      \hline
     \multirow{5}{*}{\textbf{Curriculum 4}} & 
      \web & 30 \\
      & \mcspconn & 30 \\
      & BA:n=62,ed=3 & 20 \\
      & ER:n=64,ed=0.08 & 20 \\
      & WS:n=66,p=0.2,rd=4 & 20 \\
    \end{tabular}
  \end{center}
\end{table}

\subsection{Training Techniques and Details}
\label{subsec-train-details}

\subsubsection{Stage 1: Pre-training}
For the first 1250 iterations, we pre-train our DQN with the supervised true target $y_t$ obtained as follows: 
\begin{itemize}
\item For each graph pair, we run the complete search, i.e. we do not perform any pruning for unpromising states. The entire search space is explored, and the future reward for every action can be found by finding the longest path starting from the action to a terminal state. Since graphs are small in the initial stage, such complete search can be affordable. Using Figure 2 in the main text as an example, for the action that causes state 0 to transition to state 6, the longest path is 0, 6, 7, 11, 12, 13 (or 0, 6, 7, 11, 12, 14). 
\item Given the longest path found for each action, we then compute $y_t=1+\gamma+\gamma^2+...+\gamma^{(L-1)}$, where $\gamma$ is the discount factor set to 1.0, $L$ is the length of the longest path. In the example above, $y_t=5$, intuitively meaning that at state 0, for the action that leads to state 6, in future the best solution will have 5 more nodes. In contrast, the action 0 to 1 has $y_t=4$, meaning the action 0 to 6 is more preferred. 
\item Given the true target computed for each action, we run the mini-batch gradient descents over the mse loss $(y_t-Q(s_t,a_t))^2$, where the batch size (number of sampled actions) is set to 32. 
\end{itemize}

\subsubsection{Stage 2: Imitation Learning and Stage 3}

For stage 2 (2500 iterations) and stage 3 (6250 iterations), we train the DQN using the framework proposed in ~\citet{mnih2013playing}. The difference is that in stage 2, instead of allowing the model to use its own predicted $Q(s_t,a_t)$ at each state, we let the model make a decision using the heuristics by the \mcsp algorithm, which serves as an expert providing trajectories in stage 2. We aim to outperform \mcsp eventually after training using our own predicted $Q(s_t,a_t)$ in stage 3.

Here we describe the procedure of the training process. In each training iteration, we sample a graph pair from the current curriculum for which we run the DQN multiple times until a terminal state is reached to collect all the transitions, i.e. 4-tuples in the form of $(s_t,a_t,r_t,s_{t+1})$ where $r_t$ is 1
and $y_t = 1 + \gamma \max_{a'} Q(s_{t+1},a')$, and store them into a global experience replay buffer, a queue that maintains the most recent $L$ 4-tuples. In our calculations, $L=1024$. Afterwards, at the end of the iteration, the agent gets updated by performing the mini-batch gradient descents over the mse loss $(y_t-Q(s_t,a_t))^2$, where the batch size (number of sampled transitions from the replay buffer) is set to 32.

To stabilize our training, we adopt a target network which is a copy of the DQN network and use it for computing $\max_{a'} \gamma Q(s_{t+1},a')$. This target network is synchronized with the DQN periodically, in every 100 iterations.

Since at the beginning of stage 3, the Q approximation may still be unsatisfactory, and random behavior may be better, we adopt the epsilon-greedy method by switching between random policy and Q policy using a probability hyperparameter $\epsilon$. Thus, the decision is made as $\argmax{\bm{q}}$ where $\bm{q}$ is the our predicted $Q(s_t,a_t)$ for all possible actions $(i,j)$ with $1-\epsilon$ probability; With $\epsilon$ probability, the decision is random. This probability is tuned to decay slowly as the agent learns to play the game, eventually stabilizing at a fixed probability. We set the starting epsilon to 0.1 decaying to 0.01.

\subsection{DQN Parameter Details}
\label{subsec-dqn-parameter-details}



In experiments, we use $\textsc{sum}$ followed by an MLP for $\textsc{readout}$ and $\textsc{1dconv+maxpool}$ followed by an MLP for $\textsc{interact}$.  Specifically, the MLP has 2 layers down-projecting the node embeddings from 64 to 32 dimensions. Notice that different types of embeddings require different MLPs, e.g. the MLP used for aggregating and generating graph-level embeddings is different from the MLP used for aggregating and generating subgraph-level embeddings.

For $\textsc{1dconv+maxpool}$, we apply a 1-dimensional convolutional neural network to each one of two embeddings being interacted, followed by performing max pooling across each dimension in the two embeddings before feeding into an MLP to generate the final interacted embedding. Specifically, the $\textsc{1dconv}$ contains a filter of size 3 and stride being 1. The MLP afterwards is again a 2-layer MLP projecting the dimension to 32. As shown in the main text, such learnable interaction operator brings performance gain compared to simple summation based interaction. 

The final MLP takes four components, $\bm{h}_{\mathcal{G}} =\textsc{interact}(\bm{h}_{\mathcal{G}_1},\bm{h}_{\mathcal{G}_2})$, $\bm{h}_{s} =\textsc{interact}(\bm{h}_{s1},\bm{h}_{s2})$, $\bm{h}_{\mathcal{D}c}$, and $\bm{h}_{\mathcal{D}0}$, each with dimension 32. It consists of 7 layers down-projecting the 128-dimensional input ($32 \times 4$) to a scalar as the predicted $q$ score. For every MLP used in experiments, all the layers except the last use the $\mathrm{ELU}(x)$ activation function. An exception is the final MLP, whose last layer uses the $\mathrm{ELU}(x)+1$ as the activation function to ensure positive $q$ output.

A subtle point to notice is the necessity of using either nonlinear readout or nonlinear interaction for generating the bidomain representation. Otherwise, if both operators are a simple summation, the representation for all the connected bidomains ($\textbf{h}_{\mathcal{D}c}$) is essentially the global summation of all nodes in all the connected bidomains. In other words, the nonlinearity of MLP in the readout operation or the interaction operator allows our model to capture the bidomain partitioning information in $\textbf{h}_{\mathcal{D}c}$.


\label{subsec-hyperparameter}
\section{Notes on Search}
\label{sec-search}

\subsection{Comparison with \mcsp and \mcsprl}
\label{subsec-comp-search}

The key idea of our model is that under a limited search budget, by exploring the most promising node pairs first, search can reach a larger common subgraph solution faster. In other words, for small graph pairs, all baseline models would obtain the exact MCS result as long as the search algorithm runs to complete, i.e. the stack is eventually empty, meaning no more actions to select and no more states to backtrack to (all states have been visited and fully expanded to all possible next states).

However, for large graph pairs, the task is NP-hard, and the complete search becomes nearly impossible. Exceptions exist though: For example, if the pruning condition based on the upper bound estimation is powerful enough to prune many states, the search may finish in relatively few iterations. However, we observe that the state-of-the-art solvers, \mcsp and \mcsprl, cannot finish completely for all the graph pairs used in testing. Instead of trying to improve the upper bound estimation to be more exact, in this paper, our goal is to learn a better node pair selection policy, to replace the heuristics used by baseline solvers. 

Notice that our focus on node pair selection policy instead of upper bound estimation implies that a better selection policy would mean the search can quickly find a larger solution and update its best solution found so far $maxSol$. This not only mean when the search budget is used up, the result returned is larger, but also mean that for subsequent iterations (before the iteration limit is reached), more states would be pruned by checking $\mathit{UB}_t \leq |maxSol|$, thus further helping the search. In summary, in our framework, the upper bound computation strategy remains unchanged, yet the successful node pair selection policy benefits the search in two major ways.

Since we use \mcsp in the imitation learning stage of training our DQN, and compare with \mcsp and \mcsprl in the main text, we describe their node pair selection heuristics. For \mcsp, when entering a new state, it first selects the node with the largest node degree in $\ga$, and then enumerates through all the nodes in $\gb$ in descending order of node degrees. In the original implementation provided by \mcsp, this is achieved by recursive function calls. After all the nodes in $\gb$ are visited, i.e. the depth-first search of all the node pairs $(i,j)$ finishes where $i$ is the largest-degree node in $\ga$ and $j$ is every node in $\gb$, the algorithm selects the second largest-degree node in $\ga$, and repeats the enumeration of nodes in $\gb$. After all node pairs are exhausted, the function returns and the algorithm essentially backtracks to the parent state. If the current state is the root node in the search tree, the search is complete and the exact MCS is returned. However, as noted earlier, for large graph pairs it is almost impractical to search exhaustively and a budget on the amount of search conducted has to be applied. Thus, which node pairs to visit first matters a lot for successfully extracting a large solution for large input graphs. However, as seen in Figures~\ref{fig-road_mcsp} and \ref{fig-road_our}, in many cases the true MCS does not contain large-degree nodes, since large-degree nodes tend to form more complicated subgraphs which are harder to match in the other input graph compared to simpler subgraphs like a chain. Thus, by visiting large-degree nodes first, \mcsp may not always yield a large solution fast.

In contrast, \mcsprl maintains a promising score for each node and iteratively updates the scores as search visits more states. The update formula is based on the reduction of upper bound for search, where upper bound in an overestimation of future subgraph size. As search makes progress, the scores are updated in each iteration, and nodes which cause large reduction in upper bound computation get large reward. This has the limitation that for each new graph pair, the scores associated with each node must be re-initialized to 0 and re-learned, since there is no neural network and the only learnable parameters are the scores for each node. At the beginning of search, all scores are initialized to 0, and the search has to break the tie using another heuristic, while once trained, our \mcsrlmodel can be applied to any new testing pair, and at the beginning of search, the learned parameters in \mcsrlmodel starts to benefit the search.
In other words, the whole design of \mcsprl can be regarded as a search framework with shallow learning (without neural networks or training via back-propagation). \mcsrlmodel is the first model to use deep learning for node pair selection.

Another limitation of \mcsprl is that the scores maintained for nodes reflect the potential ability of a node to reduce the upper bound for future iterations in the current search, which is indirect as the MCS aims to find the largest common subgraph, not the reduction of upper bound. Moreover, the upper bound itself is an overestimation of future subgraph size, which may or may not be close enough to the actual best future subgraph size. In contrast, we aim to predict the $q$ score for actions which directly reflect the best future subgraph size. Overall, the lack of deep learning ability causes \mcsprl not only to re-estimate the scores for the nodes for each new graph pair, but also to resort to the upper bound heuristic for updating the scores. 

To ensure the budget on search iterations is applied consistently for all the models evaluated in the main text, we adapt the original recursive implementation of \mcsp
in C++ to an iterative implementation in Python so that all the models compare with each other in the same programming language and the same search backbone algorithm. To be specific, we check the iteration count at the beginning of every search iteration and early stop the search if the pre-defined budget is reached.

\subsection{Tree vs Sequence}
\label{subsec-tree-vs-seq}

At this point, having illustrated the differences between \mcsrlmodel and \mcsp and \mcsprl, it is worth clarifying whether \mcsrlmodel search yields a tree or sequence in different stages. For training stage 1 and 2, as described in Section~\ref{subsec-train-details}, our model is just randomly initialized and not well trained, so the pre-training and imitation learning stages use the policy of \mcsp instead of using its own predicted $q$ scores. In stage 1, the complete search is performed to provide maximum amount of supervised signals, i.e. $y_t$, but in stage 2, we start using the RL training framework, i.e. experience replay buffer, target network, etc, so we run the agent multiple times until a terminal state is reached, corresponding to a sequence in a tree, which starts from the root node and ends at a leaf node. For stages 2 and 3, since sequences are generated instead of trees, for each sequence, the upper bound check always passes, because the pruning only happens when backtracking is allowed, i.e. a tree is formed. To see this more clearly, recall the pruning only happens if $\mathit{UB}_t$ $\leq |maxSol|$ in Algorithm 1 in the main text. However, during the sequence generation process, $maxSol$ keeps increasing by one each time a new state is reached. Since $\mathit{UB}_t$ $\leftarrow$ $|curSol|$ + overestimate$(s_{t})$, $\mathit{UB}_t > |curSol|$ for non-terminal $s_t$, and $|curSol|=|maxSol|$, and thus $\mathit{UB}_t > |maxSol|$, and thus the pruning never happens.

At the beginning of stage 2, the sequences we collect are usually not long, since the policy is the same as \mcsp in stage 2. However, in stage 3, we start using our predicted $Q(s_t,a_t)$ to get such sequences, and at the end of training, when we apply \mcsrlmodel to testing pairs during inference, as shown in the main text, we perform better than \mcsp and all the other baselines. In inference, we completely rely on our predicted $q$ scores as the policy, and a search tree is yielded, although the tree is not complete since the graph pairs are large and a search budget is reached. 

\subsection{Terminal Conditions}

This section discusses on how a terminal state, i.e. a leaf node in the search tree, is determined. Notice our definition of bidomain (equivalence class) does not include node labels, and in each iteration, we allow the matching between nodes in the same bidomain with the same node label. We consider the node labels as additional pruning on each bidomain, i.e. we further only allow nodes with the same label to match within bidomain when considering actions to be fed into DQN. Suppose  $\ga$ and $\gb$ are connected graphs. There are two cases:
\begin{itemize}
    \item Case 1: Nodes are unlabeled (or equivalently, all the nodes have the same label). The terminal condition is that there is no non-empty connected (adjacent) bidomains. For example, there are still some adjacent bidomains, but for each bidomain $\langle \mathcal{V}'_{k1}, \mathcal{V}'_{k2} \rangle$, at least one of $\mathcal{V}'_{k1}$ and $ \mathcal{V}'_{k2}$ is empty (containing no nodes), so there is no nodes to match in each bidomain. Examples are states 3 and 6 in Figure~\ref{fig-eq-states}.
    \item Case 2: Nodes are labeled. For the terminal condition, there may be still some non-empty connected bidomains, but the node labels do not match causing no more node pairs to select from. For example, one bidomain contains $\mathcal{V}'_{k1}$ with C and N as node labels and $ \mathcal{V}'_{k2}$ with H as node label. Then essentially there is no more node pairs left.
\end{itemize}

\subsection{Promise-based Search: Improving Search with Backtracking}
\label{subsec-backtrack-dqn}

Unlike \mcsp or \mcsprl, \mcsrlmodel is optimized to find the largest common subgraph in one try, not to prune the search space. This is because, in practice, even with advanced pruning techniques, it is not practical to exhaust the entire search space for large graphs. As a consequence, \mcsrlmodel may still fall into local solutions if it follows the branch-and-bound algorithm. Thus, \mcsrlmodel improves upon \mcsp search by backtracking to an earlier state with the most promise of finding a larger subgraph when the current best solution is not improved upon within a fixed number of iterations. Typically, when the action space is larger, there is more potential for equally good or better actions to exist outside of the one selected, thus a state's action space size is equated to its promise.

In implementation, \mcsrlmodel keep track of promise by maintaining a priority queue of states, where a state's priority is given by its action space size, in parallel with the search stack. Thus, whenever \mcsrlmodel suspects the model is in a local solution, the next state popped on line 7 of Algorithm 1 in the main text will be the state with the largest action space from the priority queue, instead of the top state in the search stack. \mcsrlmodel detects when the model is in a local solution by keeping track of the largest subgraph found since the last time the priority queue was popped (or since the start of search). If this local best solution is not improved within a fixed number of iterations, the model knows it is in a local solution. In practice, we set this number to 3.

\subsection{Notes on Equivalent States and Multiple Ground Truths}

\begin{figure*}[h]
\centering
\includegraphics[width=1.0\textwidth]{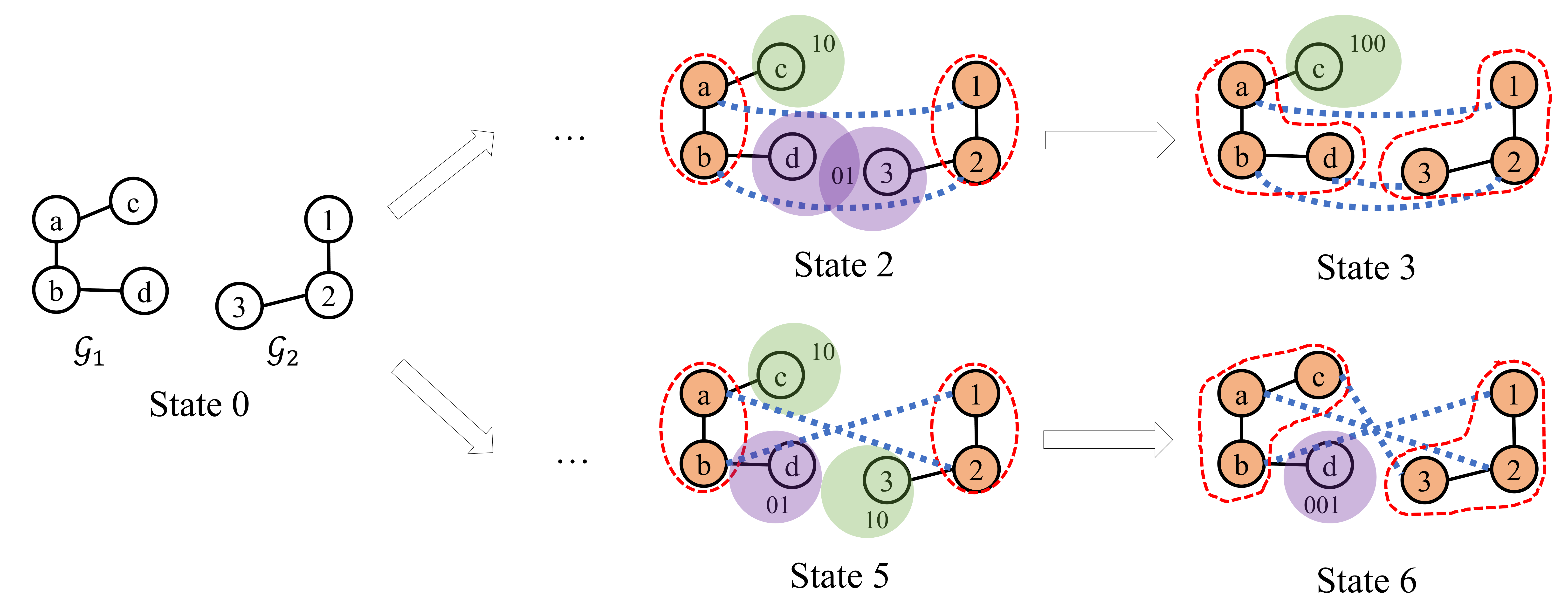}
\caption{An example illustrating the idea of equivalent states. It is important to note that states 2 3, 5, 6 are different since their node-node mappings are different. However, the solutions derived from both states 3 and 6 have the same subgraph size, 3. In other words, there can be multiple ways to arrive at the same solution size, with different underlying sequential processes to reach the final states.}
\label{fig-eq-states}
\vspace*{-4mm}
\end{figure*}

It is well known that for the MCS detection task, there can be multiple ground truth solutions with the same subgraph size. For example, in Figure~\ref{fig-eq-states}, both states 3 and 6 correspond to the same subgraph size but the states 3 and 6 are different due to their different node-node mappings. Our model maintains the node-node mappings for each state, and therefore states 3 and 6 would be reached as different states. It is important to note that for large graph pairs, reaching both states 3 and 6 usually does not happen, since the search would first reach state 3 and need many iterations to backtrack to state 0 and further many iterations to reach state 6. 

There is an even more subtle point in Figure~\ref{fig-eq-states}. Suppose the search first matches node a to node 1, denoted as state 1, then matches node b to node 2, leading to state 2. After several iterations, it backtracks to state 0 and chooses to match node b to node 2, denoted as state x, then matches node a to node 1, denoted as state y. Although states 1 and x are different, but states 2 and y are equivalent, since both states 2 and y have the same node-node mapping, i.e. a to 1 and b to 2. Thus, the search maintains an additional set of visited states and at each time a new state is reached, a checking is performed to avoid revisiting the same state twice. 

The node-node mapping is is an important component of the definition of state, not only because it differentiates the otherwise equivalent states, but also because different node-node mappings can lead to different final states and future reward (and thus must be considered by the design of DQN). Suppose the bidomain ``01'' in state 1 contains more than two nodes, i.e. there are many nodes connected to b in $\ga$ besides node d and many nodes connected to node 2 in $\gb$ besides node 3. Then state 2 is an intuitively more preferred state compared to state 5, since the matching of node b to node 2 allows more node pairs to be matched to each other in future, thus a larger action space in state 2. The value associated with state 2 thus should be larger than state 5.

\subsection{Dealing with Large Action Space}

For large graph pairs, the successful detection of MCS not only depend on the design of our critical DQN component as well as its training, but also rely on techniques which prune the large action space at each state. 

The bidomain partitioning idea has been outlined in the main text which effectively reduces the action space size by only matching nodes in the same bidomain. However, for extremely large and dense graphs, the bidomains may not split the rest of the graphs enough and the action space may still be too large. For example, consider two fully connected graphs $\ga$ and $\gb$, i.e. for every two nodes there is an edge. Then initially, there is no nodes selected, and there is only one bidomain consisting of all the node pairs. What is worse, at any state, there is always only one bidomain consisting of all the node pairs in the remaining subgraphs. Therefore, to reduce the action space further, we only compute the $q$ scores for $N_d$ nodes at most in each state.
Specifically, we first sort the candidate bidomains by their size in ascending order, and select the first $N_b$ small bidomains. Next, we sort the nodes in each bidomain by degree in descending order and select the first $N_d/N_b$ nodes with large node degrees. For our experiments, $N_b=1$ and $N_d=20$ in \mcsrlmodel;  $N_b=1$ and $N_d=3$ in \mcsrlmodelscl. We find, in practice, these settings do not drastically alter performance.


In future, additional techniques for pruning the action space can be explored. For example, instead of bounding the computation to be $N_d$, we may perform a hierarchical graph matching by first running a clustering algorithm and then matching clusters which will be bounded by the number of clusters. Another possible direction is to learn an additional Q function learning which node is more promising instead of which node pair, i.e. $Q(s_t,a_t^{(i)})$ for node $i$ and $Q(s_t,a_t^{(j)})$ for node $j$ in the action $a_t = (i, j)$. We suppose such additional Q function may bring further performance gain. 







\subsection{Analysis of Time Complexity}


Overall the branch-and-bound search has exponential worst-case time complexity due to the NP-hard nature of exact MCS detection, and our goal is to use additional overhead per search iteration to make “smarter” decision each iteration so that we can find a larger common subgraph faster (in less iterations \textit{and} real running time). Per iteration, our model  requires the neural network operations to compute a Q score instead of simply using a degree heuristic which is $\mathcal{O}(1)$. Here we analyze the time complexity of these neural operations:
\begin{itemize}
    \item To compute the node embeddings, the complexity is the same as the GNN model, which in our case is $\mathcal{O}(|\mathcal{V}|+|\mathcal{E}|)$ for GAT (since nodes must aggregate embeddings from neighbors and attention scores must be computed for each edge). Notice the node embeddings are computed by local neighborhood aggregation, and will not be updated in search, and therefore we compute the node embeddings only once at the beginning of search, and can be cached for efficiency.
    \item At each iteration, to compute a $Q$ score for a state-action pair, we run Equation (3) (in the main text) which requires computing the whole-graph, subgraph, and bidomain embeddings. Overall the time complexity for each state-action pair is $\mathcal{O}(|\mathcal{V}|-|\mathcal{V}_s|)$ where $\mathcal{V}_s$ is the number of nodes in the currently matched subgraph.  The whole-graph embeddings do not change across search, so they only need to be computed once at the beginning. The subgraph embeddings can be maintained incrementally, i.e. adding new node embeddings as search grows the subgraph.  The bidomain embeddings are computed via a series of READOUT and INTERACT operations (Equation (2)): For READOUT: We use summation followed by MLP so the runtime is $\mathcal{O}(|\mathcal{V}|-|\mathcal{V}_s|)$; For INTERACT: We use a 1D CNN followed by MLP which depends on the embedding dimension set to a constant, and does not depend on the number of nodes in the input graphs.
\end{itemize}

Overall the time complexity for each iteration is $\mathcal{O}\big(N_d^2 (|\mathcal{V}|-|\mathcal{V}_s|)\big)$.
\section{Baseline Description and Comparison}
\label{sec-baselines}


For all the models used in experiments, we evaluate their performance under the same search framework, i.e. with consistent search iteration counting, upper bound estimation, etc. \mcsp and \mcsprl use heuristics to select node pairs, which is ineffective as shown in the main text and has been described in Section~\ref{subsec-comp-search}. Therefore, this Section focuses on the comparison with the rest baselines, i.e. \gwqap~\citep{xu2019scalable}, \pca~\citep{wang2019learning}, and \mcstreemodel~\citep{bai2020neural}.

\gwqap performs Gromov-Wasserstein discrepancy~\citep{peyre2016gromov} based optimization \textit{for each graph pair} and outputs a matching matrix $\bm{Y}$ for all node pairs indicating the likelihood of matching which is treated the same way as our $q$ scores, i.e. at each search iteration we index into $\bm{Y}$ to select a node pair. \pca and \mcstreemodel also output a matching matrix but require supervised training, and thus are trained using the same training data graph pairs as our \mcsrlmodel but with different loss functions and training signals. 
During testing, we apply the trained model on all testing graph pairs. For medium-size synthetic and real-world testing graph pairs, each method is given a budget of 500 search iterations. For large real-world graph pairs, each method is given a budget of 7500 search iterations. For million-node real-world graph pairs, each method is given a budget of 50 minutes. These budgets were chosen based on when the models' performances stabilized. We then describe each method in more details.

\subsection{\gwqap}
\gwqap is a state-of-the-art graph matching model for general graph matching. The task is not about MCS specifically, but instead about matching two graphs with its own criterion based on the Gromov-Wasserstein discrepancy~\citep{peyre2016gromov}. Therefore, we suppose the matching matrix $\bm{Y}$ generated for each graph pair can be used as a guidance for which node pairs should be visited first. In other words, we pre-compute the matching scores for all the node pairs before the search starts, and in iteration, we look up the matching matrix and treat the score as the $q$ score for action selection. \pca and \mcstreemodel essentially compute a matching matrix too, and it is worth mentioning that all the three methods cannot learn a score based on both the state and the action. They can be regarded as generating the matching scores based on the whole graphs only without being conditioned and dynamically updated on states and actions. 

\subsection{\pca}
\pca is a state-of-the-art image matching model, where each image is turned into a graph with techniques such as Delaunay triangulation~\citep{lee1980two}. It utilizes similarity scores and normalization to perform graph matching. We adapt the model to our task by replacing these layers with 3 GAT layers, consistent with \mcsrlmodel. As the loss is designed for matching image-derived graphs, we alter their loss functions to binary cross entropy loss similar to \mcstreemodel which will be detailed below.

\subsection{\mcstreemodel}
\mcstreemodel is proposed for MCS detection with similar idea from \pca that a matching matrix is generated for each graph pair using GNNs. However, they both require the supervised training signals, i.e. the complete search for training graph pairs must be done to guide the update of \pca and \mcstreemodel. In contrast, \mcsrlmodel is trained under the RL framework which does not require the complete search (in stage 2 and 3, only sequences are generated as detailed in Section~\ref{subsec-tree-vs-seq}). This has the benefit of exploring the large action space in a ``smarter'' way and eventually allows our model to outperform \pca and \mcstreemodel. In implementation, the complete search is not possible for large training graph pairs, so instead we apply a search budget and use the best solution found so far to guide the training of \pca and \mcstreemodel.

Regarding the subgraph extraction strategy, for all the baselines, we use the same branch and bound algorithm, which is the state-of-the-art search designed for MCS~\citep{mccreesh2017partitioning}. However, as mentioned in Section~\ref{subsec-backtrack-dqn}, only our model is equipped with the ability to backtrack in a principled way. The main text shows the performance gain to \mcsrlmodel brought by the backtracking ability.

\section{More Results with Analysis}
\label{sec-more-results}

\subsection{Best Solution Sizes across Time}

\begin{figure*}
     \centering
     \subfloat[\road]{\includegraphics[width=0.4\textwidth]{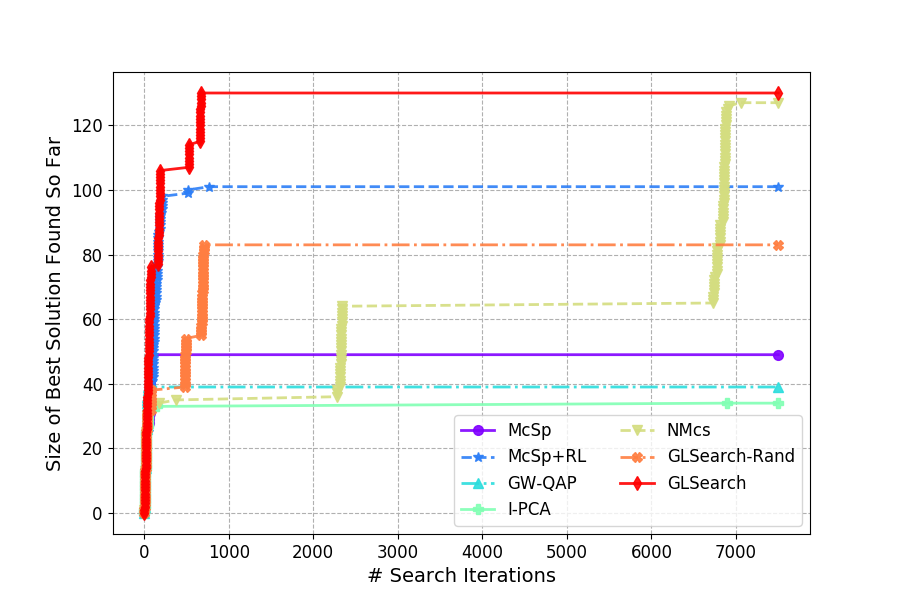}}
     \subfloat[\kgen]{\includegraphics[width=0.4\textwidth]{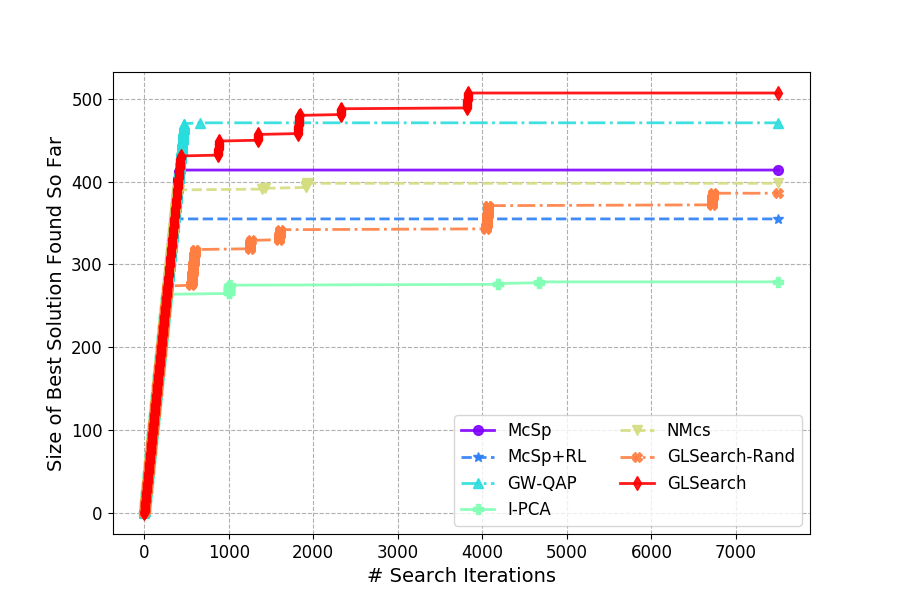}}

     \subfloat[\kgzh]{\includegraphics[width=0.4\textwidth]{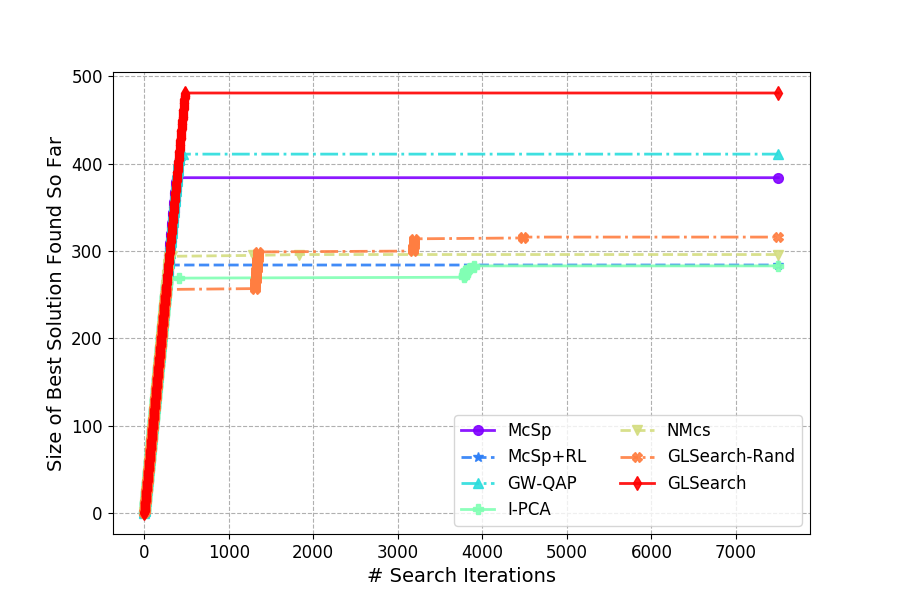}}
     \subfloat[\kgenzh]{\includegraphics[width=0.4\textwidth]{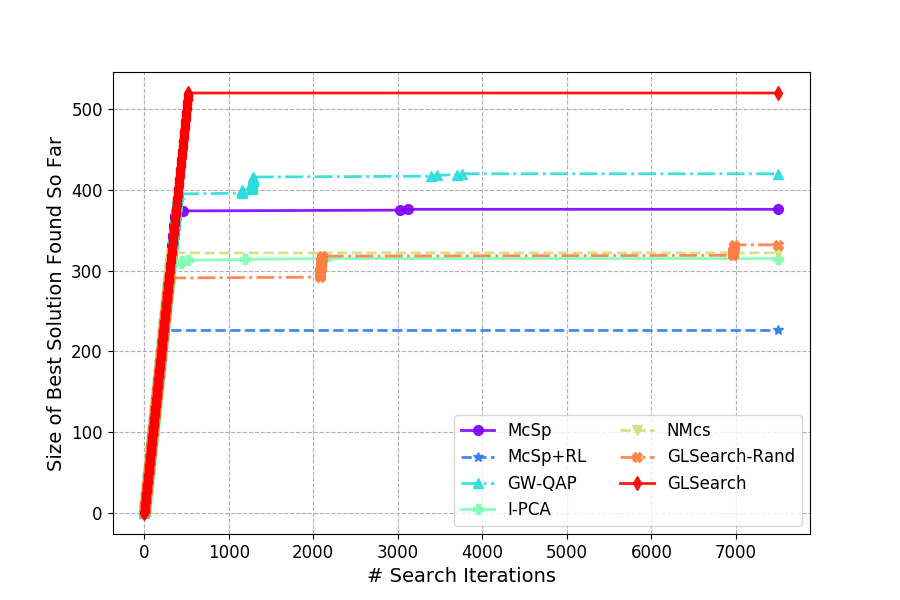}}

     \subfloat[\enron]{\includegraphics[width=0.4\textwidth]{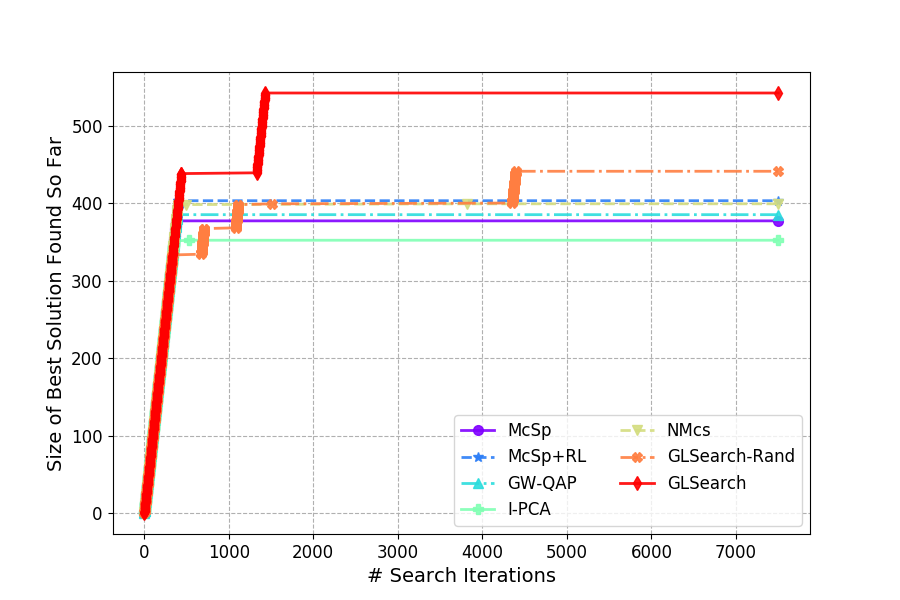}}
     \subfloat[\amazon]{\includegraphics[width=0.4\textwidth]{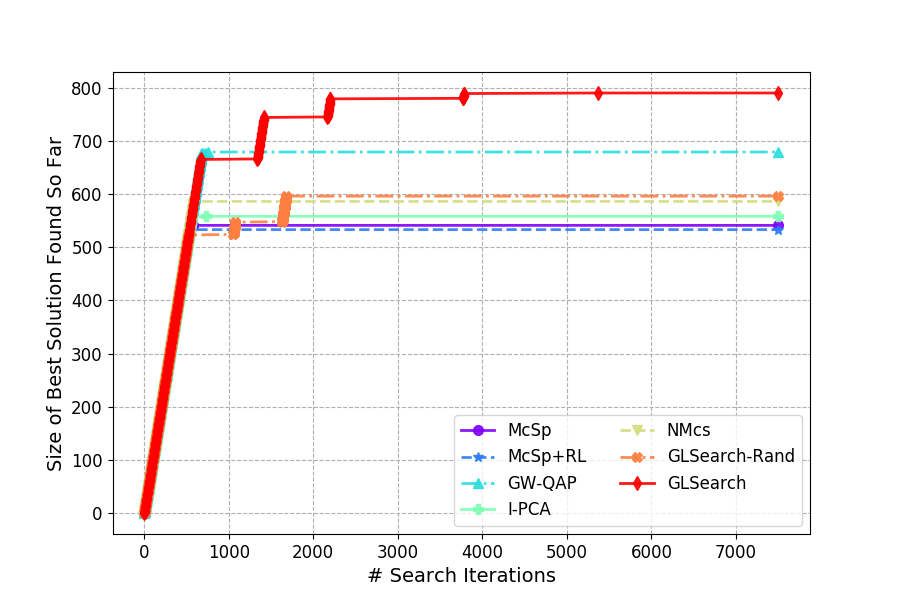}}

     \subfloat[\circuit]{\includegraphics[width=0.4\textwidth]{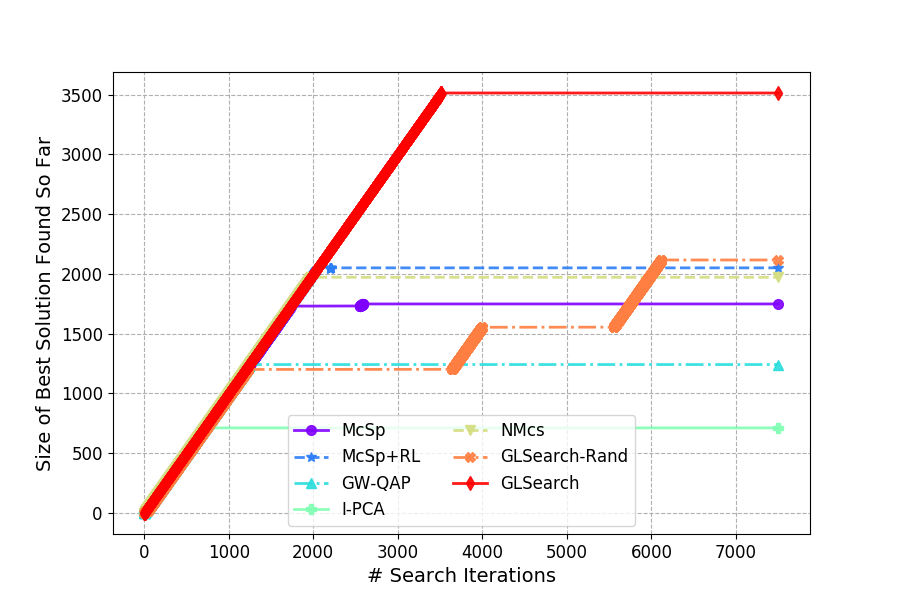}}
     \subfloat[\ppi]{\includegraphics[width=0.4\textwidth]{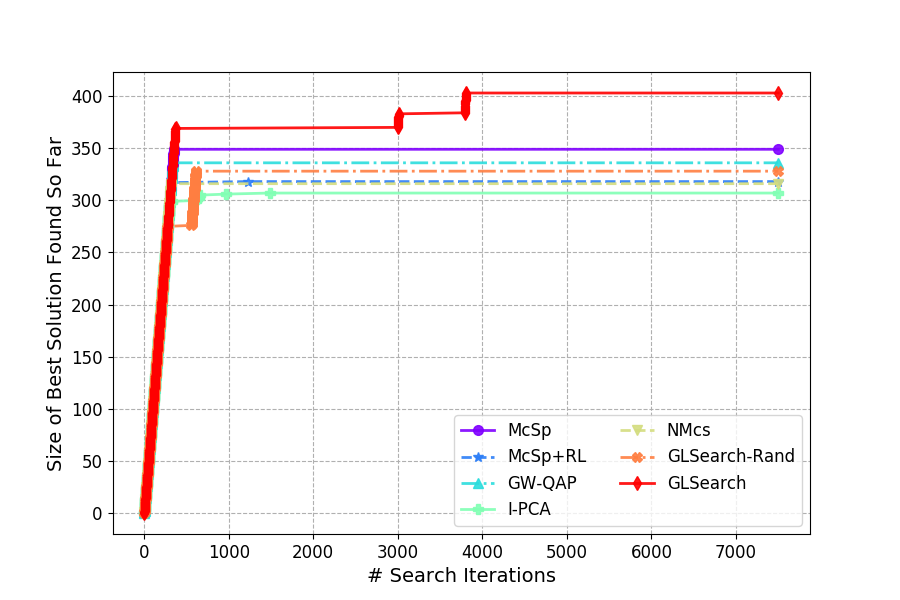}}
     \caption{For each method, we maintain the best solution found so far in each iteration during the search process. We plot the size of the largest extracted common subgraphs found so far vs search iteration count for all the methods across all the datasets. The larger the subgraph size, the better (``smarter'') the model in terms of quickly finding a large MCS solution under limited budget for large graphs.}
     \label{fig-iter-size}
\end{figure*}

Figure~\ref{fig-iter-size} shows that under the budget we set, for the large real-world graph pairs, all the methods reach a ``steady state'' where the best solution found so far no longer grows. This means the search continues but the search cannot find a larger solution, illustrating the fact that search gets ``stuck''. In theory, given infinitely many iterations, all the models will eventually find the true MCS, which is the largest, but since the task is NP-hard, the search space can be exponential in the worst case (subject to pruning but in all the testing graph pairs the search has not finished yet), such budget has to be applied to search. 
At the end of the budget iterations, though, all the models have made ``mistakes'' (visiting unpromising states) and in order to find even larger common subgraphs, the search needs to backtrack potentially many times to fix those ``mistakes'' (backtracking to a very early state). 

Admittedly, there is always the possibility that for more iterations, some baseline method may find a larger solution. Besides, in each iteration \mcsrlmodel does take more running time overhead as shown in Table~\ref{table-additional-runtime}. However, the point of our model is to quickly find a larger solution in as few iterations as possible, not to find a large solution given too many iterations. In other words, the goal of \mcsrlmodel is to be \textbf{smart} enough to \textbf{quickly} find a large solution instead of purely finding a good solution. Figure~\ref{fig-iter-size} shows that \mcsrlmodel not only finds solutions larger than baselines \textit{when} the iteration budget is reached but also finds larger solutions faster than baselines \textit{before} the budget is reached.

\begin{table*}[h]
\small
  \begin{center}
    \caption{Average running time per iteration (msec). 
    }
    \begin{tabular}{l|llllllll}
    \label{table-additional-runtime}
    \textbf{Method}&
    \textbf{\road} &
    \textbf{\kgen} &
    \textbf{\kgzh} &
    \textbf{\kgenzh} &
    \textbf{\enron} &
    \textbf{\amazon} &
    \textbf{\circuit} &
    \textbf{\ppi}
    \\
    \hline
    \mcsp         & 2.040  & 10.724 & 1.415  & 0.974  & 1.722   & 2.891  & 1.776   & 0.498  \\
    \mcsprl      & 0.894  & 6.834  & 2.103  & 1.247  & 2.166   & 3.107  & 2.080   & 0.559  \\
    \hline
    \gwqap        & 0.548  & 0.834  & 0.546  & 4.692  & 1.041   & 3.419  & 1.550   & 0.546 \\
    \pca          & 1.152  & 1.797  & 0.967  & 0.897  & 1.739   & 2.725  & 3.792   & 0.636 \\
    \mcstreemodel & 2.394  & 4.172  & 4.648  & 5.667  & 9.610   & 9.788  & 7.471   & 15.742 \\
    \hline
    \randq        & 17.392 & 66.418 & 67.342 & 67.946 & 163.005 & 71.972 & 655.447 & 83.488 \\
    \mcsrlmodel   & 8.132  & 66.552 & 71.409 & 96.262 & 135.087 & 51.181 & 37.377  & 60.509 \\
    \end{tabular}
  \end{center}
\end{table*}


\begin{figure*}
     \centering
     \subfloat[\road]{\includegraphics[width=0.4\textwidth]{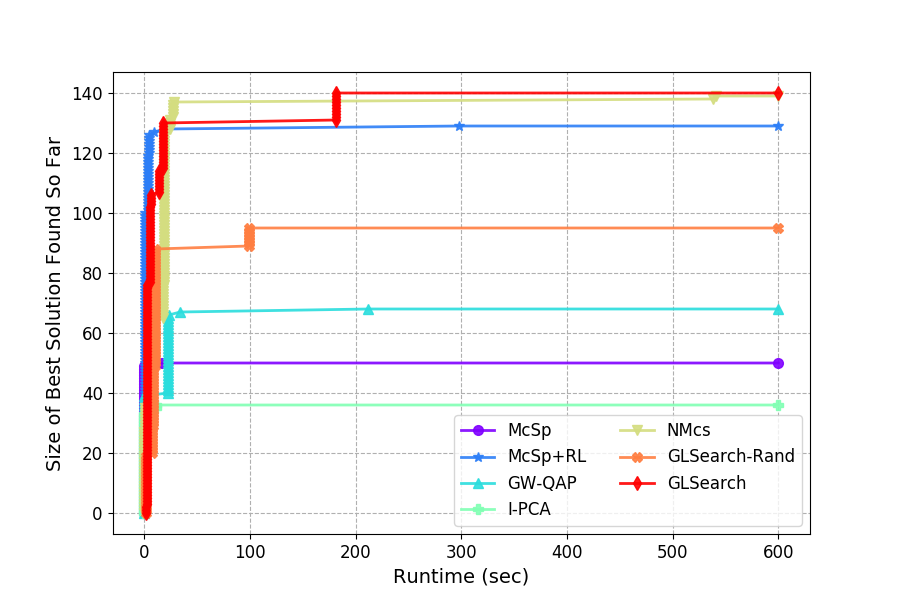}}
     \subfloat[\kgen]{\includegraphics[width=0.4\textwidth]{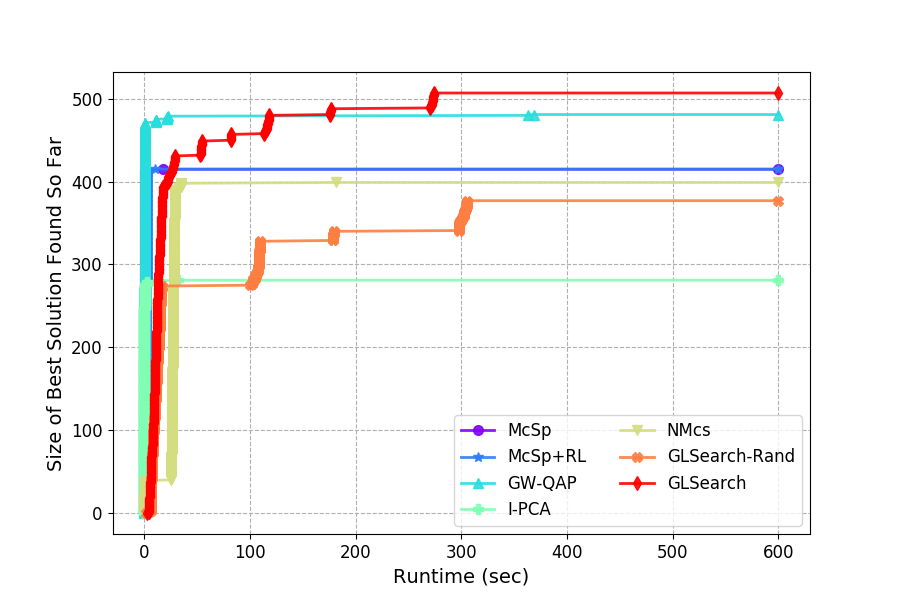}}

     \subfloat[\kgzh]{\includegraphics[width=0.4\textwidth]{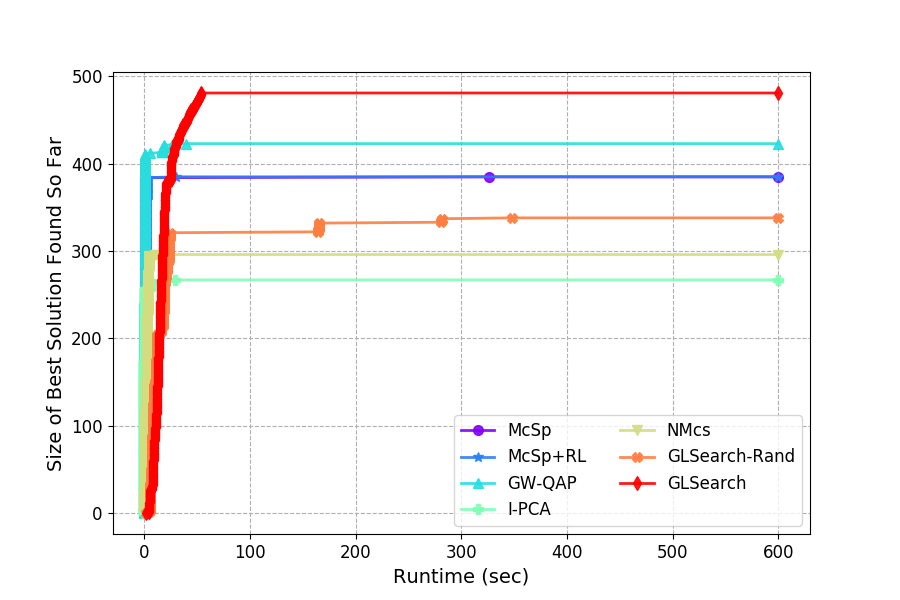}}
     \subfloat[\kgenzh]{\includegraphics[width=0.4\textwidth]{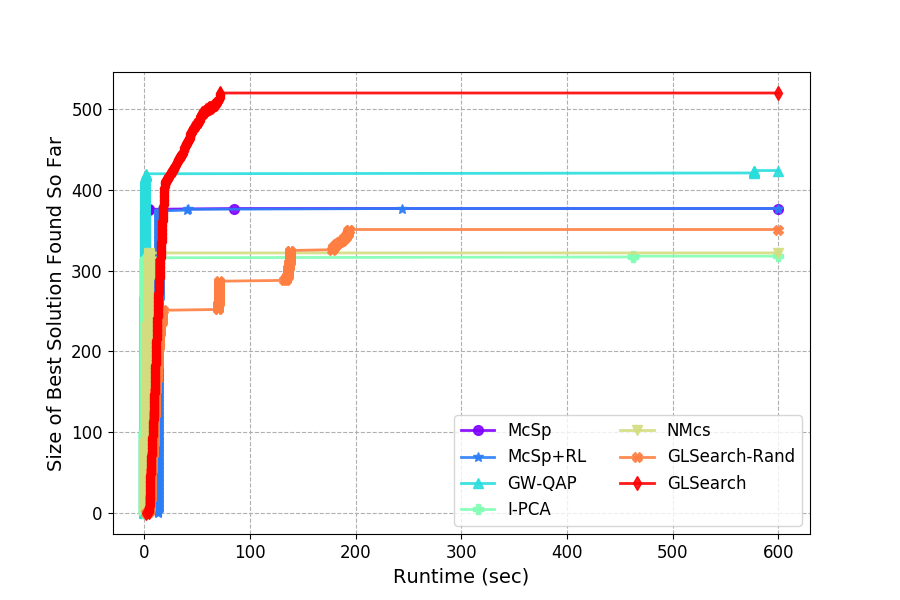}}

     \subfloat[\enron]{\includegraphics[width=0.4\textwidth]{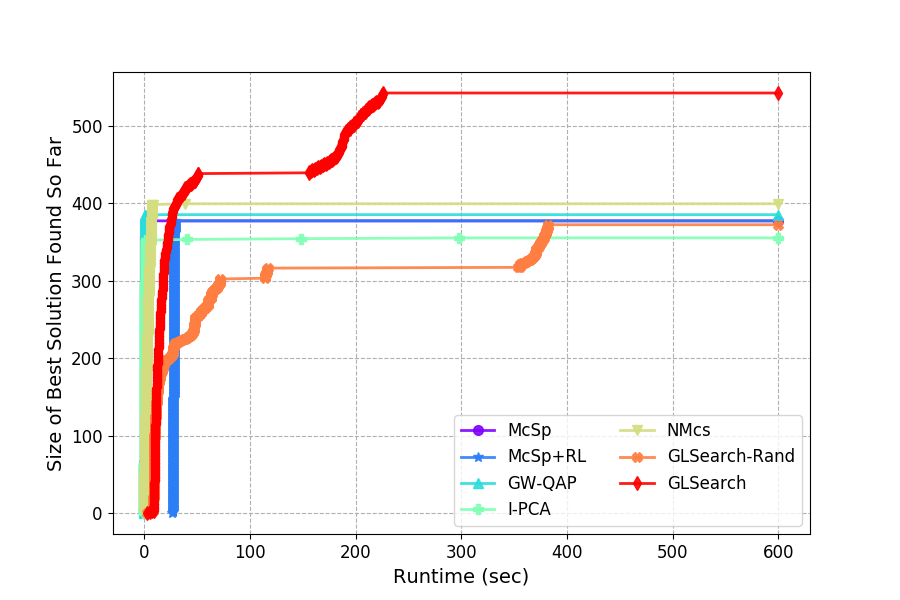}}
     \subfloat[\amazon]{\includegraphics[width=0.4\textwidth]{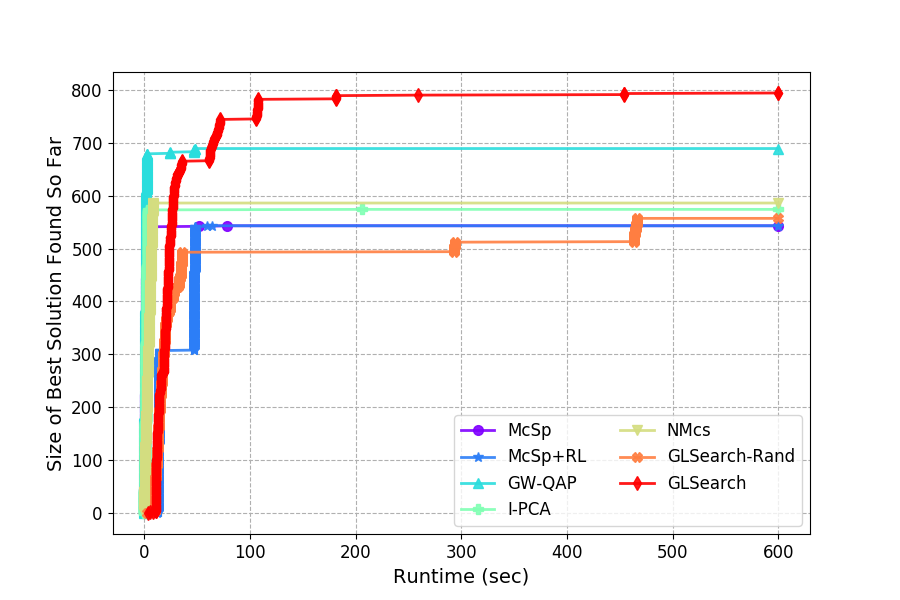}}

     \subfloat[\circuit]{\includegraphics[width=0.4\textwidth]{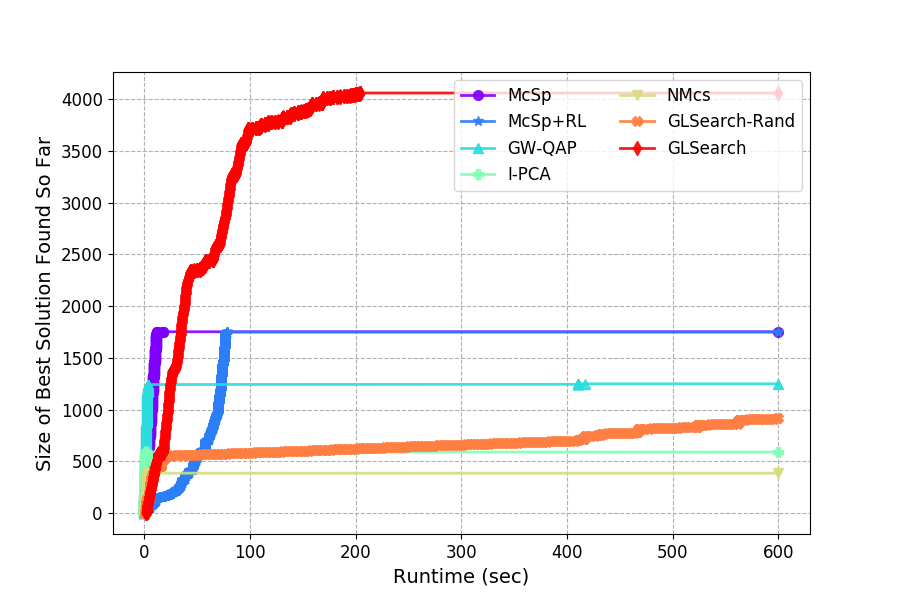}}
     \subfloat[\ppi]{\includegraphics[width=0.4\textwidth]{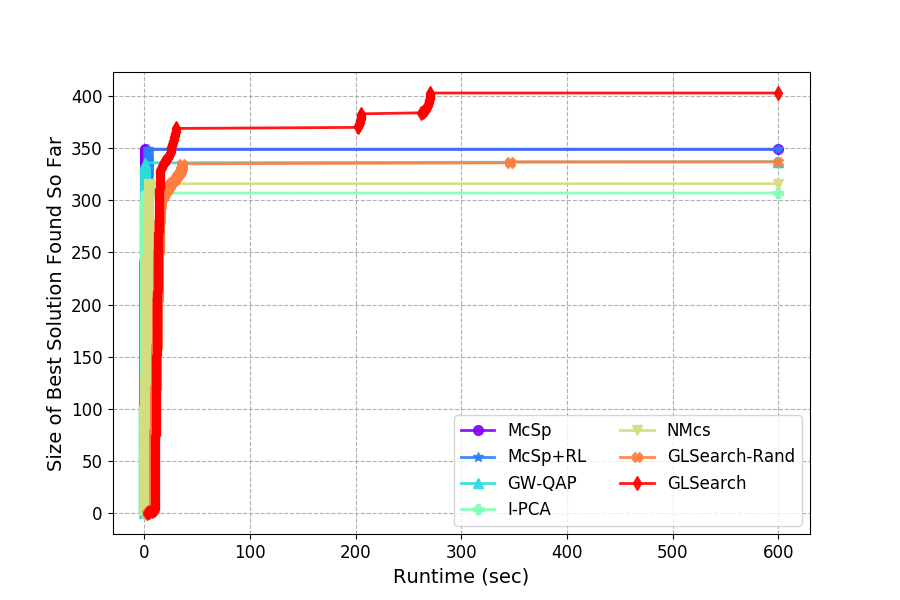}}
     \caption{For each method, we maintain the best solution found so far in each iteration during the search process. We plot the size of the largest extracted common subgraphs found so far vs the real running time for all the methods across all the datasets. The larger the subgraph size, the better (``smarter'') the model in terms of quickly finding a large MCS solution under limited budget for large graphs.}
     \label{fig-time-size}
\end{figure*}

In addition to reaching a larger solution is less iterations, \mcsrlmodel also reaches better solutions with respect to runtime, as shown in Figure~\ref{fig-time-size}.\footnote{In general, we refer to the running time results to show the performance gains of \mcsrlmodel and \mcsrlmodelscl; however, to focus solely on the effects of different actions selected by different policies on the final common subgraph, we also compare by iterations.
} Notice, \mcsrlmodel finds the same large subgraph in 10 minutes as in 7500 iterations. Although per iteration, it is slower than \mcsp per iteration, \mcsrlmodel finds a larger solution in usually less than a minute. Moreover, our implementation can be further optimized. Note, we adapted all baselines to run on Python for fair comparison, and made sure the search iteration counting is consistent across all baselines and the results shown in the main text. At this stage, our main goal is to explore the idea of \textit{\textbf{``learning to search''}}, which has been experimentally verified to be a promising direction of research, and leave the efforts of implementation optimization using various techniques as future focus.

\subsection{Additional Ablation Study}

Table~\ref{table-abl-ptil} shows that pre-training and imitation learning benefit the performance under four out of the eight datasets. On \enron and \ppi, without pre-training, our model performs better, which may be attributed to the fact that they are dense graphs (Table~\ref{table-test-data}) while the training graphs used in stage 1 are relatively small and sparse (Section~\ref{subsec-cur-learning}).

\begin{table*}[h]
\small
  \begin{center}
    \caption{Contribution of pre-training and imitation learning to the performance of \mcsrlmodel. ``no-sup'' denotes the removal of the pre-training stage (The first 3750 iterations: IL; The last 6250 iterations: Normal QDN training); ``no IL'' denotes the removal of the imitation learning stage (The first 3750 iterations: pre-training; The last 6250 iterations: normal DQN training); ``no sup; no IL'' indicates the entire training (10000 iterations) is normal DQN training.}
    
    \begin{tabular}{l|llllllll}
    \label{table-abl-ptil}
    \textbf{Method}&
    \textbf{\road} &
    \textbf{\kgen} &
    \textbf{\kgzh} &
    \textbf{\kgenzh} &
    \textbf{\enron} &
    \textbf{\amazon} &
    \textbf{\circuit} &
    \textbf{\ppi} \\
    \hline
\mcsrlmodel (no sup) & 0.557 & 0.957 & 0.946 & 0.904 & \textbf{1.000} & 0.999 & \textbf{1.000} & \textbf{1.000} \\
\mcsrlmodel (no IL) & \textbf{1.000} & 0.933 & 0.965 & 0.887 & 0.357 & 0.875 & 0.666 & 0.632 \\
\mcsrlmodel (no sup; no IL) & 0.678 & 0.907 & 0.896 & 0.837 & 0.401 & 0.949 & 0.949 & 0.651 \\
    \hline
    \mcsrlmodel & 0.879 & \textbf{1.000} & \textbf{1.000} & \textbf{1.000} & 0.412 &\textbf{1.000} & 0.855 & 0.688\\
    \hline
    \hline
    \nnodes & 149 & 486 & 465 & 471 & 1318 & 790 & 4112 & 587 \\
    \end{tabular}
  \end{center}
\end{table*}

Table~\ref{table-abl-promise} shows that the promised-based search improves the performance under four out of the eight datasets. For the other four datasets, the performance does not change, indicating that the backtracking to an earlier promising state based on the DQN output at least does not hurt the performance. In the cases like \enron and \ppi, whose average node degrees are large, the promise-based search improves the performance by a large amount, showing the usefulness of the proposed strategy.

\begin{table*}[h]
\small
  \begin{center}
    \caption{Contribution of promise-based search (Section~\ref{subsec-backtrack-dqn}) to the performance of \mcsrlmodel. ``no promise'' denotes that the search does not use the proposed promise-based search, i.e. it does not backtrack to an earlier state if the search makes no progress after a certain amount of iterations, and instead, it continues the regular branch and bound search.}
    
    \begin{tabular}{l|llllllll}
    \label{table-abl-promise}
    \textbf{Method}&
    \textbf{\road} &
    \textbf{\kgen} &
    \textbf{\kgzh} &
    \textbf{\kgenzh} &
    \textbf{\enron} &
    \textbf{\amazon} &
    \textbf{\circuit} &
    \textbf{\ppi} \\
    \hline
    \mcsrlmodel (no promise) & 0.879 & \textbf{1.000} & \textbf{1.000} & \textbf{1.000} & 0.412 &\textbf{1.000} & 0.855 & 0.688 \\
    \hline
    \mcsrlmodel & \textbf{1.000} & \textbf{1.000} & \textbf{1.000} & \textbf{1.000} & \textbf{1.000} &\textbf{1.000} & \textbf{1.000} & \textbf{1.000} \\
    \hline
    \hline
    \nnodes & 131 & 508 & 482 & 521 & 543 & 791 & 3515 & 404 \\
    \end{tabular}
  \end{center}
\end{table*}

\subsection{Results on Graph Pairs with Known MCS Size Lower Bound}
\label{subsec-known-mcs}

To better understand the quality of subgraphs found by \mcsrlmodel, we construct new datasets with known lower bound MCS sizes. This is accomplished through generating 2 new graphs that share a common subgraph from the existing large real-world graph datasets. For each real-world graph, $\mathcal{G}_0$, we randomly extract 3 different subgraphs of the same size, $\mathcal{S}_0$, $\mathcal{S}_1$, and $\mathcal{S}_2$, by running breadth first search from 3 different starting nodes and extracting the explored induced subgraph. To construct a new graph pair, ($\mathcal{G}_1$, $\mathcal{G}_2$), we form $\mathcal{G}_1$ by connecting $\mathcal{S}_0$ to $\mathcal{S}_1$ with 20 random edges and form $\mathcal{G}_2$ by connecting $\mathcal{S}_0$ to $\mathcal{S}_2$ with 20 random edges. Thus, connections between $\mathcal{S}_0$ nodes are the same in both $\mathcal{G}_1$ and $\mathcal{G}_2$, but connections between $\mathcal{G}_1 \setminus \mathcal{S}_0$ and $\mathcal{G}_2 \setminus \mathcal{S}_0$ are different. Notice, the lower bound of MCS size in these new datasets would be $|\mathcal{S}_0|$, and we name the new dataset by adding `ss' to the parent dataset's name.

The results on \circuit and these datasets (Table~\ref{table-acc-known-lb}) show no MCS method guarantees to always detect a solution that is as large as the known MCS solution/lower bound. This suggests the difficulty of the task itself. In practice, though, \mcsrlmodel is still preferred compared to baselines due to its better performance.

\begin{table*}[h]
\small
  \begin{center}
    \caption{Results on graph pairs with a common core subgraph (lower bound of MCS), with a fixed runtime of 10 minutes.}

    \begin{tabular}{l|llllll}
    \label{table-acc-known-lb}
    \multirow{2}{*}{\textbf{Method}} &
    \textbf{\road-ss} &
    \textbf{\kgen-ss} &
    \textbf{\kgzh-ss} &
    \textbf{\enron-ss} &
    \textbf{\amazon-ss} &
    \textbf{\ppi-ss} \\
    & 444 & 778 & 762 & 1346 & 1406 & 860 \\
         \hline
    \mcsp         & 0.588   & 0.466   & 0.544   & 0.216   & 1.000   & 0.233  
\\
    \mcsprl       & 0.588   & 0.466   & 0.544   & 0.214   & 1.000   & 0.233 \\
    \hline
    \mcsrlmodel   & \textbf{1.000} & \textbf{1.000} & \textbf{1.000} & \textbf{1.000} & \textbf{1.000} & \textbf{1.000} \\
    \hline
    \hline
    \nnodes & 188     & 389     & 350     & 673     & 703     & 430    \\
    \textsc{Core (Lower Bound) Size} & 222     & 389     & 381     & 673     & 703     & 430    \\
    \end{tabular}
  \end{center}
\end{table*}

\subsection{Result Visualization on \roadca and \roadtx}
\label{subsec-vis-big}

\begin{figure*}[h]
\centering
\includegraphics[width=1.0\textwidth]{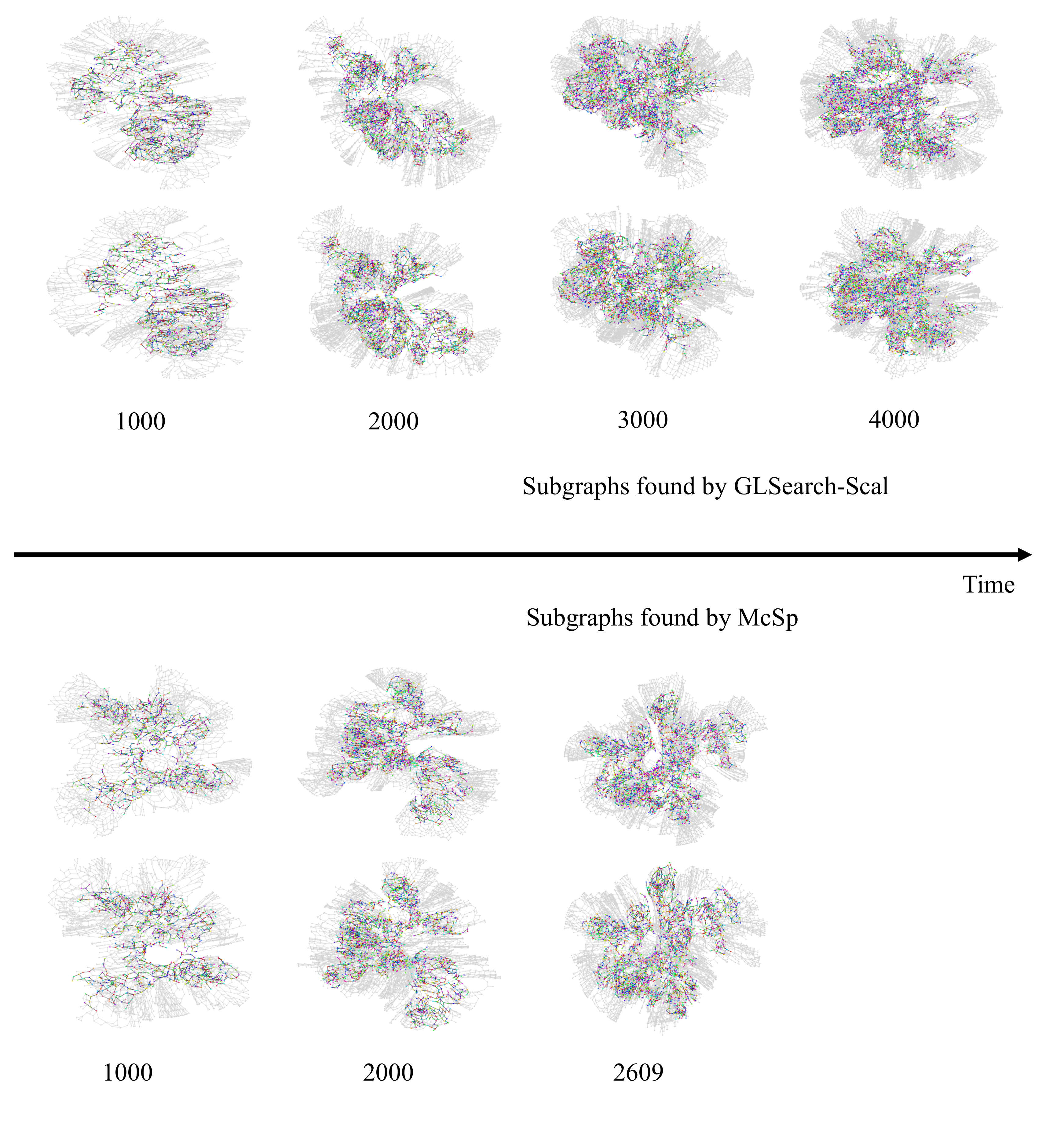}
\caption{Visualization of subgraphs found by \mcsrlmodelscl and \mcsp on \roadca. The subgraphs found by each method grows across time (until the budget of 50 minutes is reached), and the sizes of the subgraphs are denoted at the bottom of each figure. \mcsp only finds 2609 nodes as shown in Figure 4 (a) in the main text, and the solution does not increase further.}
\vspace*{10mm}
\label{fig-road-ca}
\end{figure*}

\begin{figure*}[h]
\centering
 \includegraphics[width=1.0\textwidth]{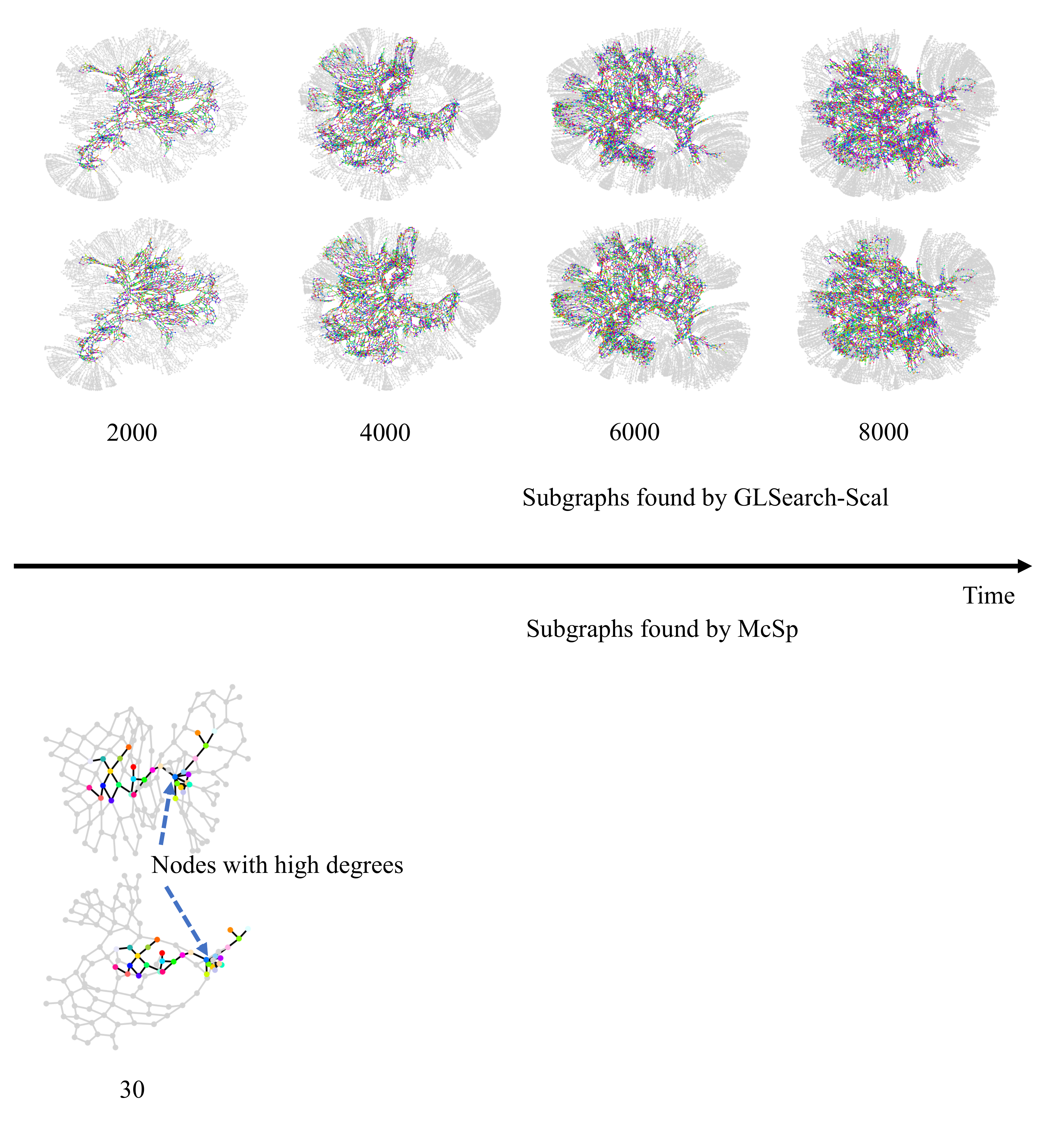}
\caption{Visualization of subgraphs found by \mcsrlmodelscl and \mcsp on \roadtx. The subgraphs found by each method grows across time (until the budget of 50 minutes is reached), and the sizes of the subgraphs are denoted at the bottom of each figure. \mcsp only finds 30 nodes as shown in Figure 4 (b) in the main text, and the solution does not increase further.}
\vspace*{10mm}
\label{fig-road-tx}
\end{figure*}

For the largest two graph pairs, \roadca and \roadtx, in order to clearly see and compare the subgraph growth across time of \mcsrlmodelscl and \mcsp, we perform the following visualization: At every 1000 or 2000 iterations, we plot the graph pair and highlight the matched subgraphs. As shown in Figures~\ref{fig-road-ca} and \ref{fig-road-tx}, from the left to the right, the growing of the extracted common subgraphs can be seen. 

In order to render the best visualization, the following techniques are used: (1) Since the input graphs are two large, we only plot the extracted subgraphs and the remaining graphs around the extracted subgraphs by performing breadth-first search starting from the extracted subgraphs (gray color); (2) For the matched subgraphs, to clearly see the node-node mapping, we ensure the node layout positions are the same across $\ga$ (top) and $\gb$ (bottom), and use colors for the matched subgraph nodes to indicate the node-node mapping\footnote{The nodes of \roadca and \roadtx are unlabeled and the colors are only used to highlight node-node mapping.}.

It is noteworthy that on \roadtx, \mcsp only finds 30 nodes as shown in Figure~\ref{fig-road-tx}. By examining the plot carefully, we can see that this is caused by poor choice of actions that make further selections of node pairs impossible, i.e. there is no more node pairs to choose in action space, reaching a terminal condition\footnote{We aim to find induced common subgraphs, meaning that if a new node pair is selected, all the edges between the new nodes and the existing nodes must also be included, and in this example, any new node pair would lead to the resulting subgraphs \textbf{not} isomorphic to each other.}. As shown in Figure 4 in the main text, \mcsp spends the rest of the time backtracking and exploring the rest of the search space, but given the exponentially growing search space size, \mcsp cannot easily leave the local optimum. Given infinitely long running time, however, all methods under the branch and bound algorithm will eventually find the exact MCS solution, which is an impractical assumption in real world. This illustrates the necessity of making smart node pair selection choices through the search instead of relying on heuristics. As a fact, \mcsprl also finds 30 nodes, since it degenerates to the same policy as \mcsp at the beginning of each new graph pair as mentioned in the main text. We verify that its found subgraphs are indeed exactly the same as \mcsp.

Another insight from Figure~\ref{fig-road-tx} is that, the initial node pair selection made by \mcsp, which is highlighted as ``nodes with high degrees'' in the plot, misleads \mcsp into eventually finding a very small solution. In contrast, \mcsrlmodelscl uses embeddings at subgraph, whole-graph, and bidomain levels to make a decision at each step, which capture the network structure better than just the node degree information used by \mcsp. A real-world analogy can be road networks in downtown areas which tend to be grid structures versus road networks in rural areas which tend to be less rigid. \mcsrlmodelscl takes graph structure into account and matches one downtown area with another downtown area, so the resulting matched subgraphs can be very large, while \mcsp is misled by its heuristic into matching two high-degree nodes\footnote{The node degree of both nodes is 12 and we verify they are the highest-degree nodes in $\ga$ and $\gb$.} at the beginning of search, but unfortunately the two areas do not have similar road network structures and eventually the matched subgraphs are small. It must be noted, however, that this analogy is only a high-level hypothesis for this specific case, and in general, for more complicated graph structures, the actual decisions made by \mcsrlmodelscl usually cannot be easily explained using such an analogy. 

\section{Extensions of \mcsrlmodel}

\mcsrlmodel can be extended for a flurry of other MCS definitions, e.g. approximate MCS, MCS for weighted and directed graphs, etc. via a moderate amount of change to the search and learning components. In this section, we briefly outline what could be done for these tasks.

For approximate MCS detection, the bidomain constraint must be relaxed. One method of relaxing this constraint is to allow sets of nodes belonging to different but similar bidomains to match to each other. For instance, nodes in $\ga$ from the bidomain of bitstring ``00110'' could map with nodes in $\gb$  from the bidomain of bitstring ``00111'', since they are only 1 hamming distance away. Such relaxations as this can be made stricter or looser based on the application. The difference would be the search framework, thus the learning part of \mcsrlmodel can largely stay the same.

Regarding MCS for graphs with non-negative edge weights, assuming our task is to maximize the sum of edge weights in the MCS, instead of defining $r_t=1$, we can alter the reward function to be the difference of the sum of edge weights before and after selecting a node pair $r_t = \Sigma_{e \in S_t^{(u,v)}} w(e) - \Sigma_{e \in S_t} w(e)$ where $S_t$ is the edges of currently selected subgraph, $S_t^{(u,v)}$ is the edges of the subgraph after adding node pair, $(u,v)$, and $w(\cdot)$ is a function that takes and edge and returns its weight. As the cumulative sum of rewards at step T is the sum of edge weights $\Sigma_{t \in [1,...,T]} r_t = \Sigma_{e \in S_T} w(e)$ and reinforcement learning aims to maximize the cumulative sum of rewards, we can adapt \mcsrlmodel to optimize for MCS problems with weighted edges.

Regarding MCS for directed graphs, the bidomain constraint may be altered such that every bit in the bidomain string representations now has 3 states: `0' for disconnected, `1' for connected by in-edge, and `2' for connected by out-edge. By considering the inward/outward direction of a bitstring, we can guarantee the isomorphism of directed graphs. In this case, the search framework would only differ in how bidomains are partitioned. The learning part of \mcsrlmodel would stay the same for this application.

\begin{figure*}[h]
\centering
\includegraphics[width=0.5\textwidth]{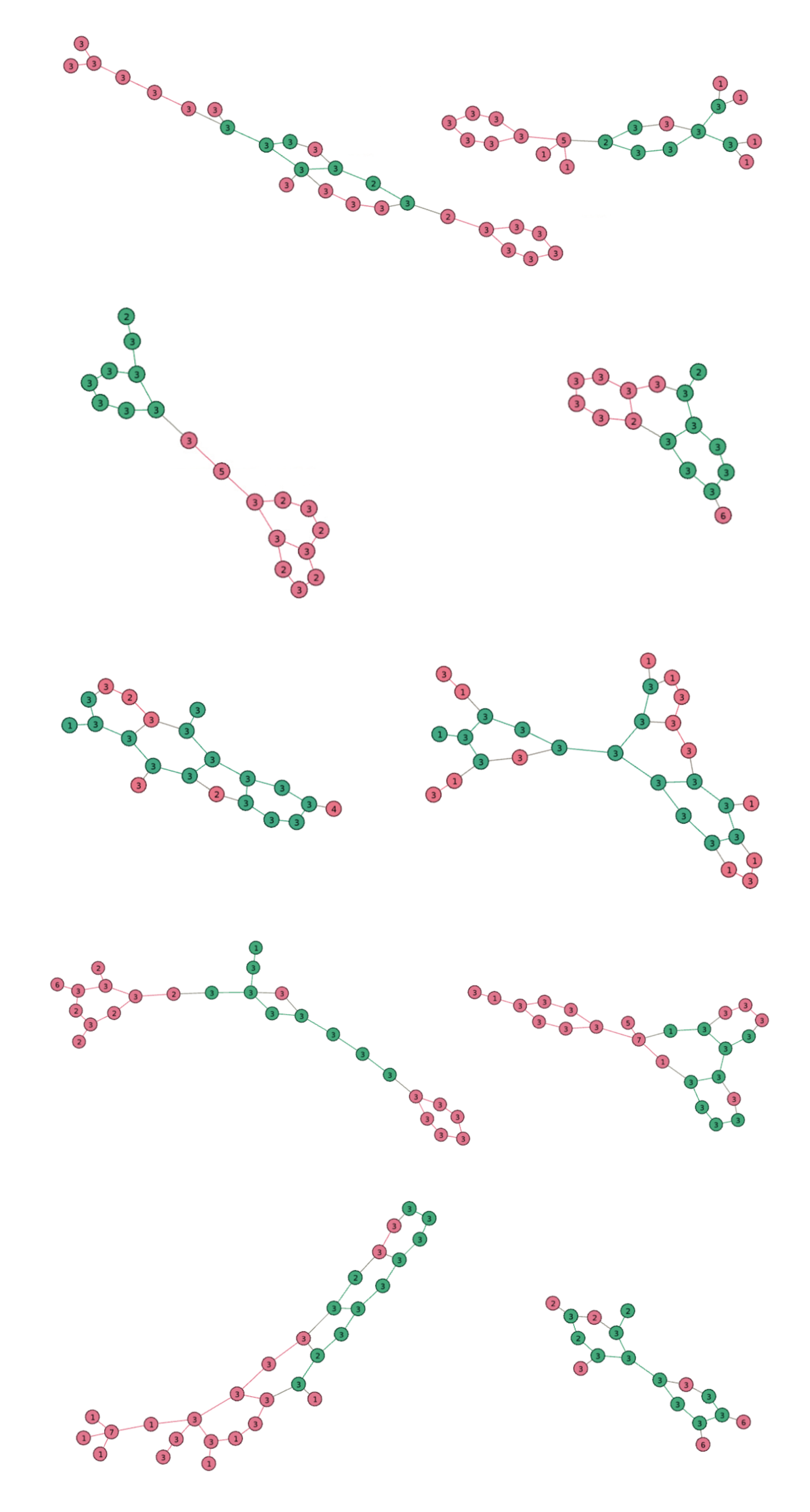}
\caption{Visualization of 5 sampled graph pairs with the MCS results by \mcsrlmodel on \nci. Each chemical compound node has its label indicated in the plot. Extracted subgraphs are highlighted in green.}
\label{fig-nci_our}
\end{figure*}

Regarding MCS for chemical compounds, we are aware of works that aim to tackle MCS in the chemoinformatics domain~\citep{schietgat2013polynomial,duesbury2018comparison}, and we do admit that the current definition of MCS may not satisfy constraints in the chemoinformatics domain, e.g. Figure~\ref{fig-nci_our}. However, since we aim to design a general solver, we believe our current work has strong potential to be extended in the future with domain-specific constraints.




More generally, we believe that there are many more extensions to \mcsrlmodel in addition to the ones listed, such as disconnected MCS, network alignment, or subgraph extraction. Further exploration of these are to be done as future efforts.

\section{Additional Result Visualization}


We plot the testing graph pairs and the results of \mcsp and \mcsrlmodel in this Section using a software called Gephi~\citep{bastian2009gephi}. For all the figures except Figure~\ref{fig-circuit_orig}, we use two colors for nodes, one for the selected subgraphs by the model, the other for the remaining subgraphs that cannot be further matched within the search budget. When plotting, we use larger circle size for nodes with larger degrees.


In general, \mcsrlmodel is a less interpretable but more powerful method compared to heuristic baselines. That said, \mcsrlmodel presents some insights that may be useful for producing new hand-crafted heuristics.
 
\begin{figure*}[h]
\centering
\includegraphics[width=0.6\textwidth]{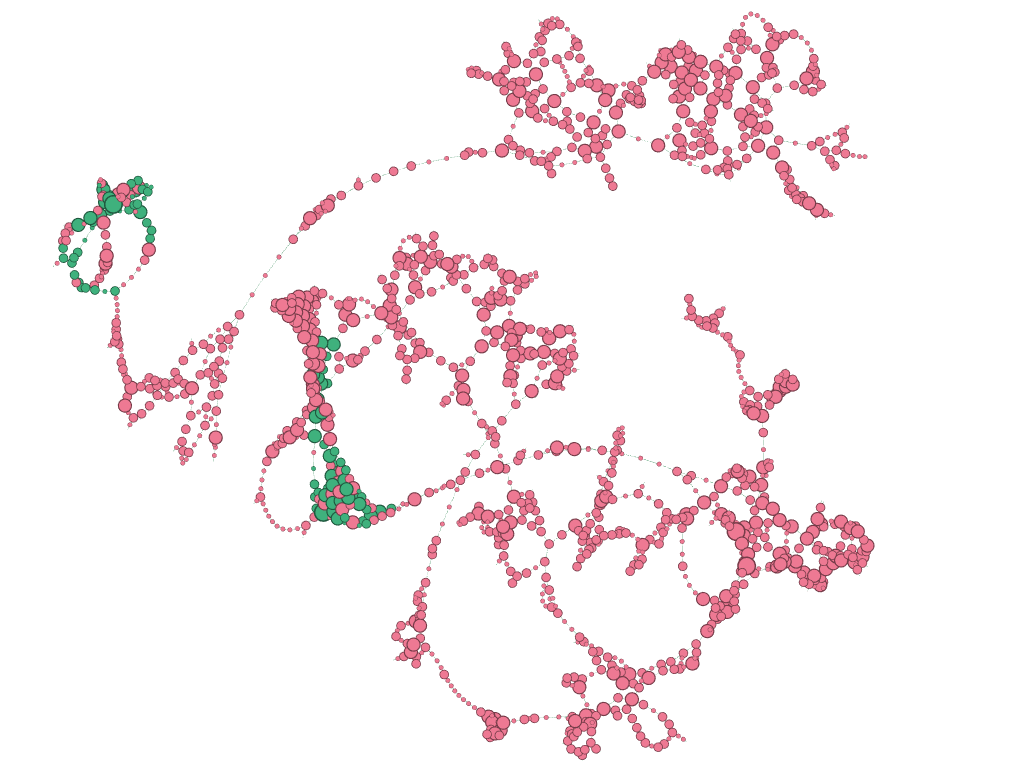}
\caption{Visualization of \mcsp result on \road. Extracted subgraphs are highlighted in green.}
\label{fig-road_mcsp}
\end{figure*}

\begin{figure*}[h]
\centering
\includegraphics[width=0.6\textwidth]{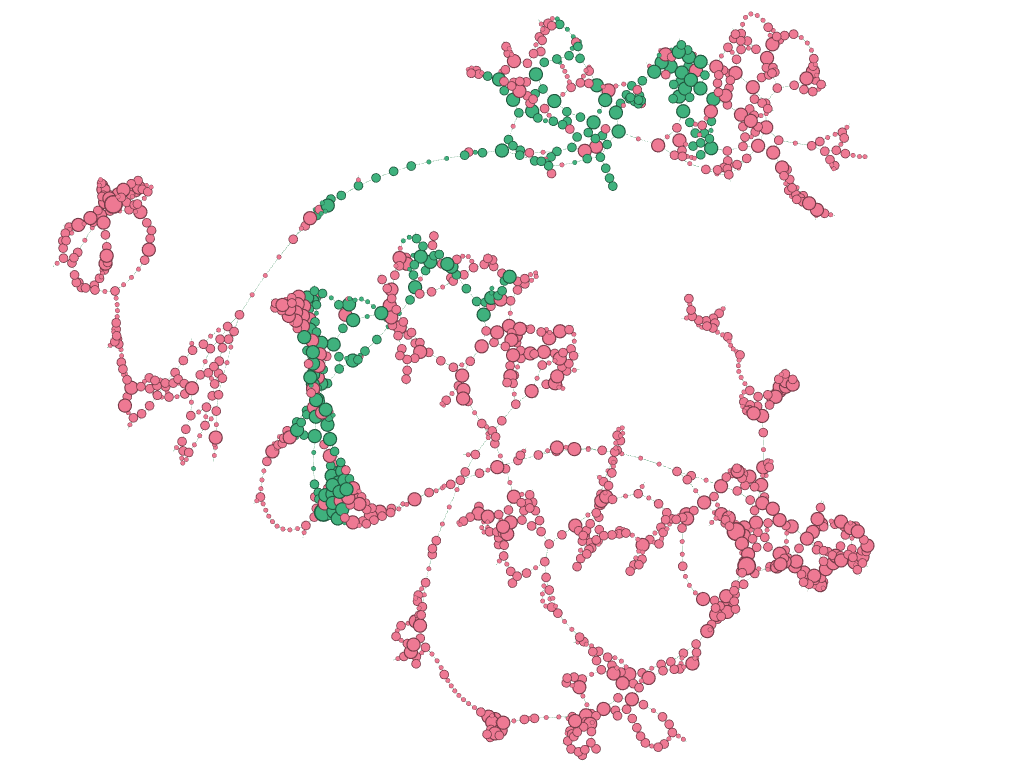}
\caption{Visualization of \mcsrlmodel result on \road. Extracted subgraphs are highlighted in green.}
\label{fig-road_our}
\end{figure*}

\mcsrlmodel identifies ``smart'' nodes which can lead to larger common subgraphs faster. For example, in the road networks (\road), as in Figure~\ref{fig-road_mcsp} and \ref{fig-road_our}, our learned policy selects nodes with smaller degrees which allow for easier matching. The common subgraphs in road networks are most likely long chains, where nodes tend to have low degrees. In contrast \mcsp always chooses high-degree nodes first leading to smaller extracted subgraphs.
 
\mcsrlmodel identifies ``smart'' \textit{matching} of nodes which can lead to larger common subgraphs faster. For example, in the circuit graph (\circuit), we find 3 high-degree nodes that, when correctly matched, greatly reduces the matching difficulty of remaining nodes (see Figure~\ref{fig-circuit_orig} and \ref{fig-circuit_mcsp}). Upon further analysis, \mcsp incorrectly matches the 3 high degree nodes (matching high degree node to low degree node). This happens when matching high-degree node correctly would break the isomorphism constraint (due to the current selected subgraph being incorrectly matched). \mcsrlmodel conscientiously adds node pairs so that it will always be able to match the 3 high degree nodes correctly.
 
We believe two aspects of \mcsrlmodel design lead to this phenomenon. First, \mcsrlmodel encodes neighborhood structures that are k-hop away. \mcsp only looks at a single node and not its relationship with k-hop neighbors. Second, \mcsrlmodel considers scores on the node-node pair granularity, thus it will only match nodes with similar local neighborhoods. \mcsp only considers scores on the node granularity, potentially matching 2 nodes with dissimilar neighborhoods together.
 
From these insights, one can potentially design a heuristic to first detect highly valuable nodes and guide a policy which prioritizes the matching of these critical nodes, or create better heuristics that consider not only uses the features of a single node but also the similarity between the 2 nodes being matched.
 

\clearpage

\begin{figure*}[h]
\centering
\includegraphics[width=0.6\textwidth]{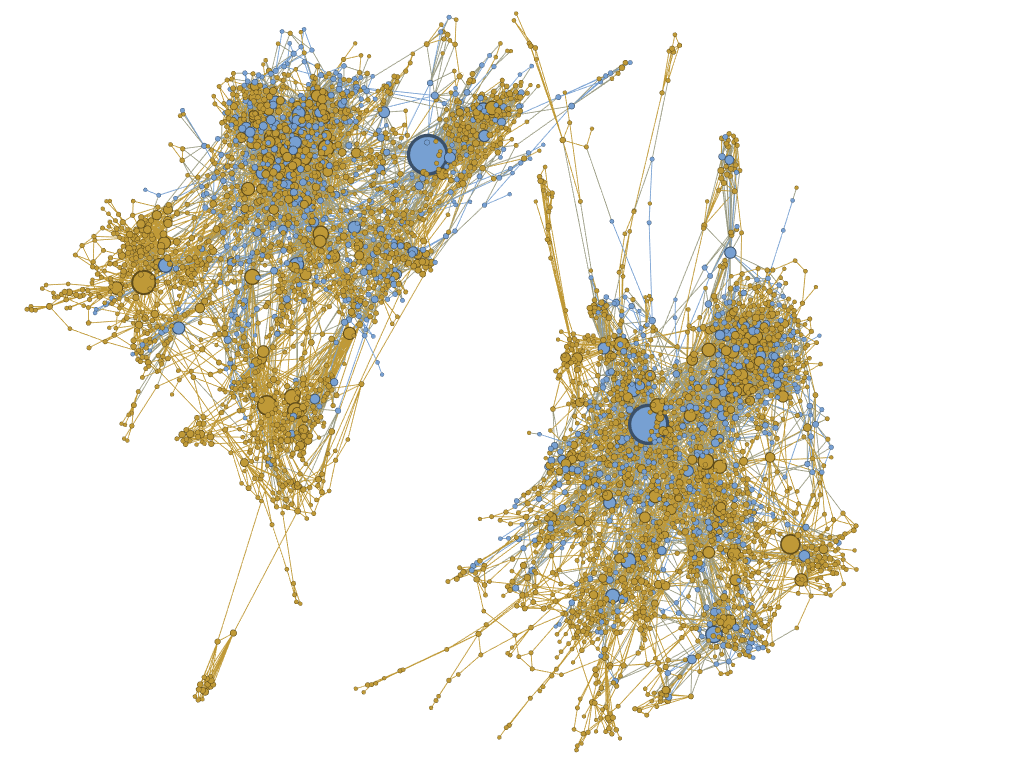}
\vspace*{-4mm}
\caption{Visualization of \mcsp result on \kgen. Extracted subgraphs are highlighted in blue.}
\label{fig-kgen_mcsp}
\end{figure*}

\begin{figure*}[h]
\centering
\includegraphics[width=0.6\textwidth]{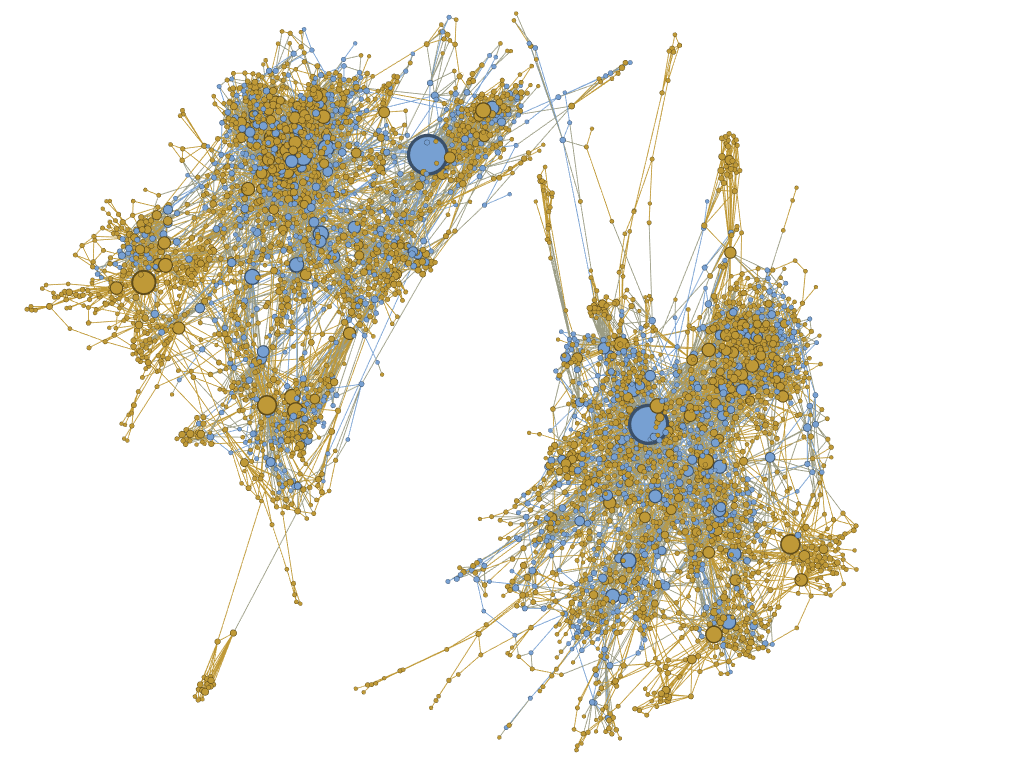}
\caption{Visualization of \mcsrlmodel result on \kgen. Extracted subgraphs are highlighted in blue.}
\label{fig-kgen_our}
\end{figure*}

\begin{figure*}[h]
\centering
\includegraphics[width=0.6\textwidth]{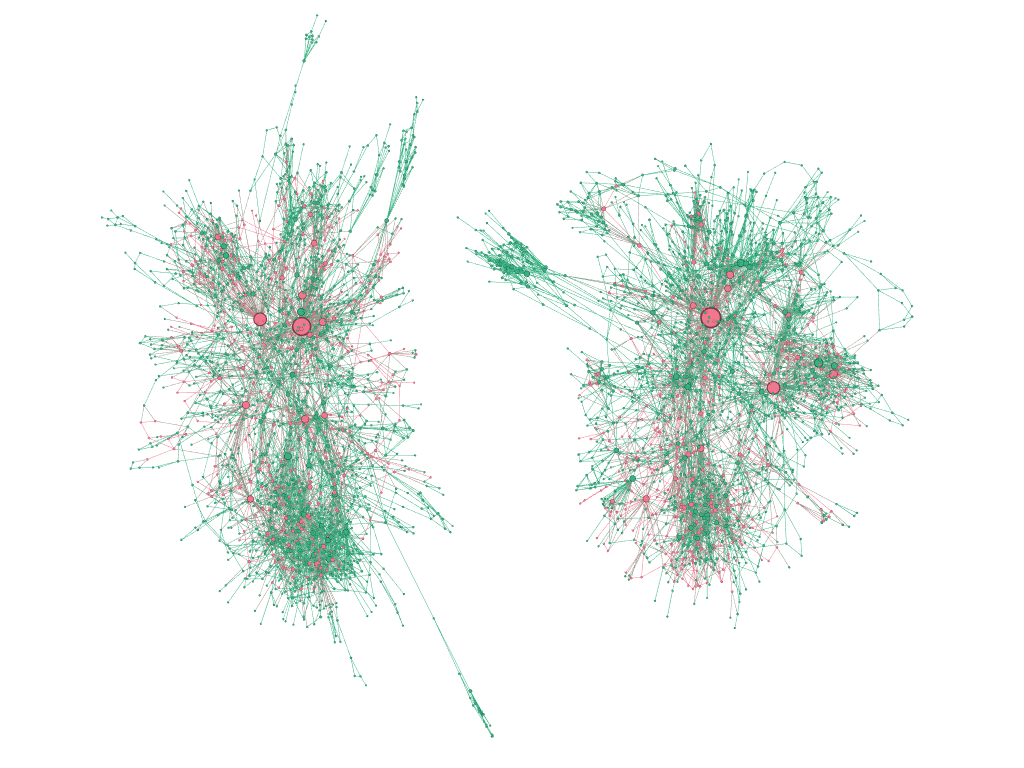}
\caption{Visualization of \mcsp result on \kgzh. Extracted subgraphs are highlighted in pink.}
\label{fig-kgen_mcsp}
\end{figure*}

\begin{figure*}[h]
\centering
\includegraphics[width=0.6\textwidth]{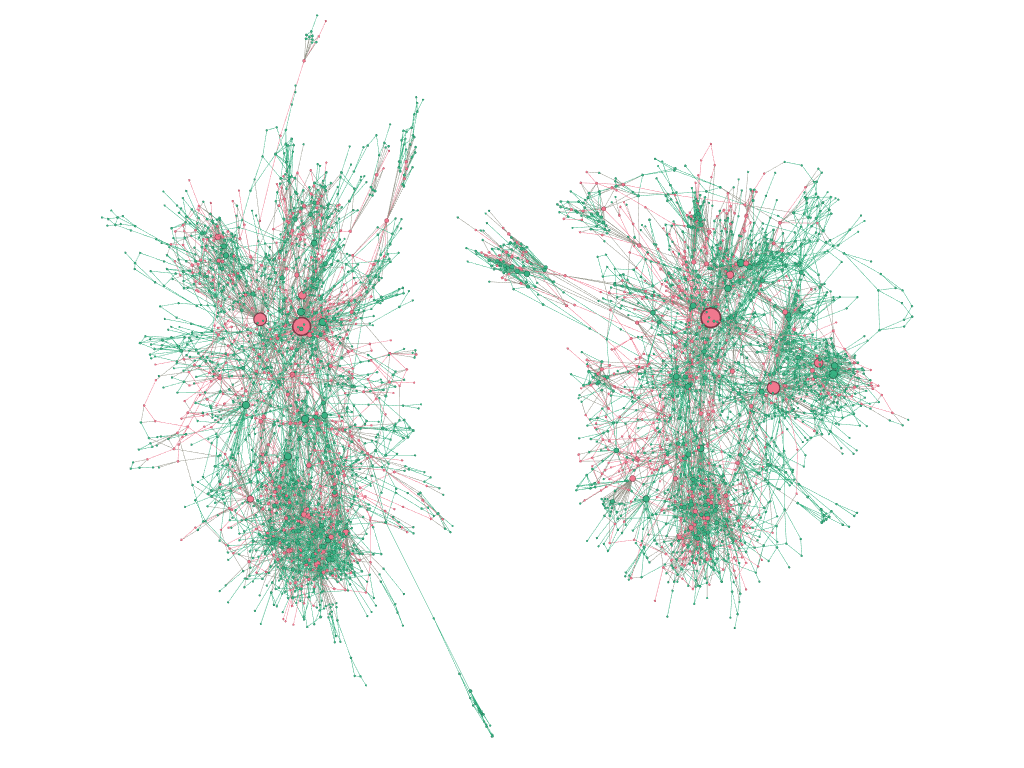}
\caption{Visualization of \mcsrlmodel result on \kgzh. Extracted subgraphs are highlighted in pink.}
\label{fig-kgen_our}
\end{figure*}

\begin{figure*}[h]
\centering
\includegraphics[width=0.6\textwidth]{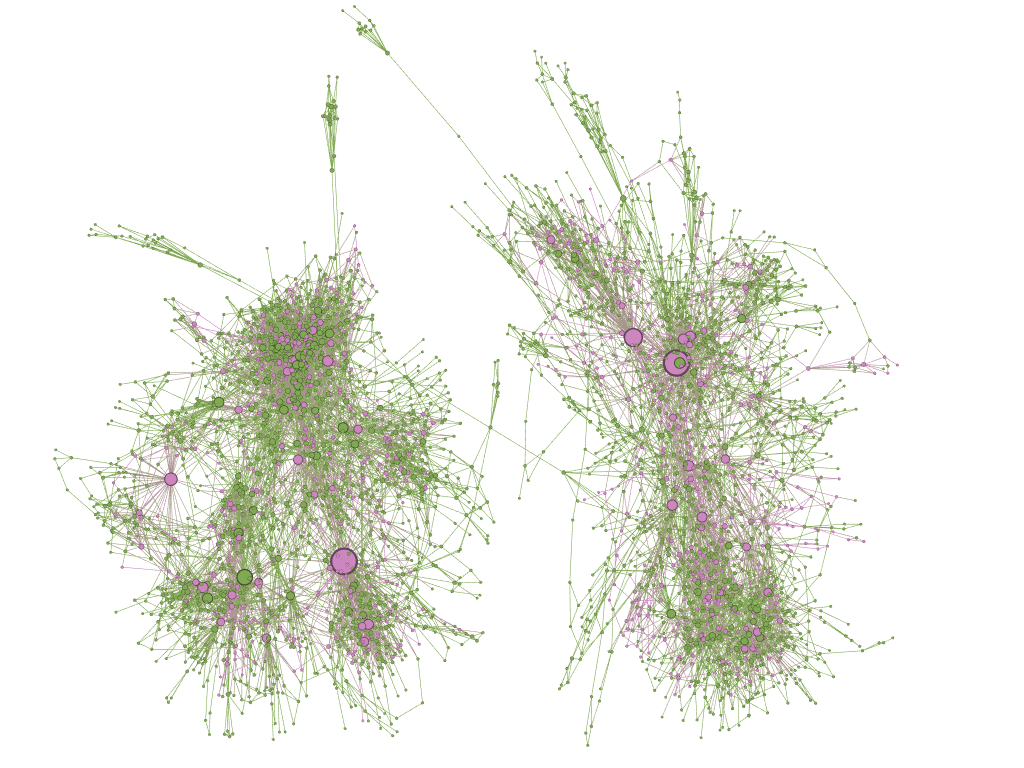}
\caption{Visualization of \mcsp result on \kgenzh. Extracted subgraphs are highlighted in purple.}
\label{fig-kgenzh_mcsp}
\end{figure*}

\begin{figure*}[h]
\centering
\includegraphics[width=0.6\textwidth]{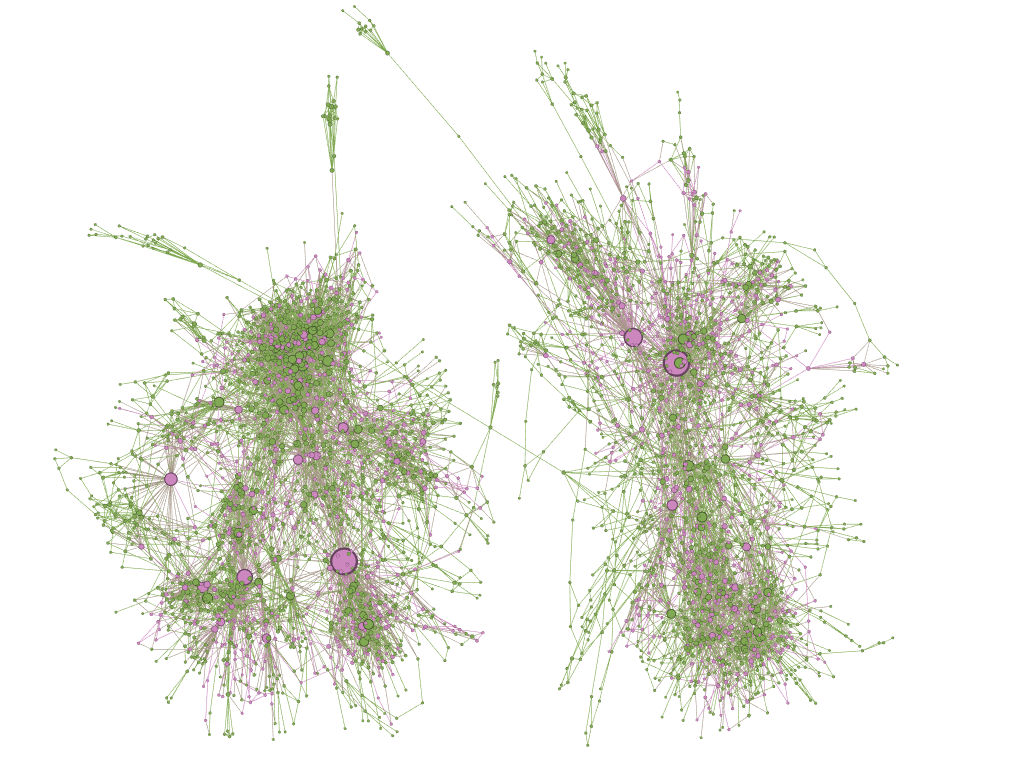}
\caption{Visualization of \mcsrlmodel result on \kgenzh. Extracted subgraphs are highlighted in purple.}
\label{fig-kgenzh_our}
\end{figure*}

\begin{figure*}[h]
\centering
\includegraphics[width=0.6\textwidth]{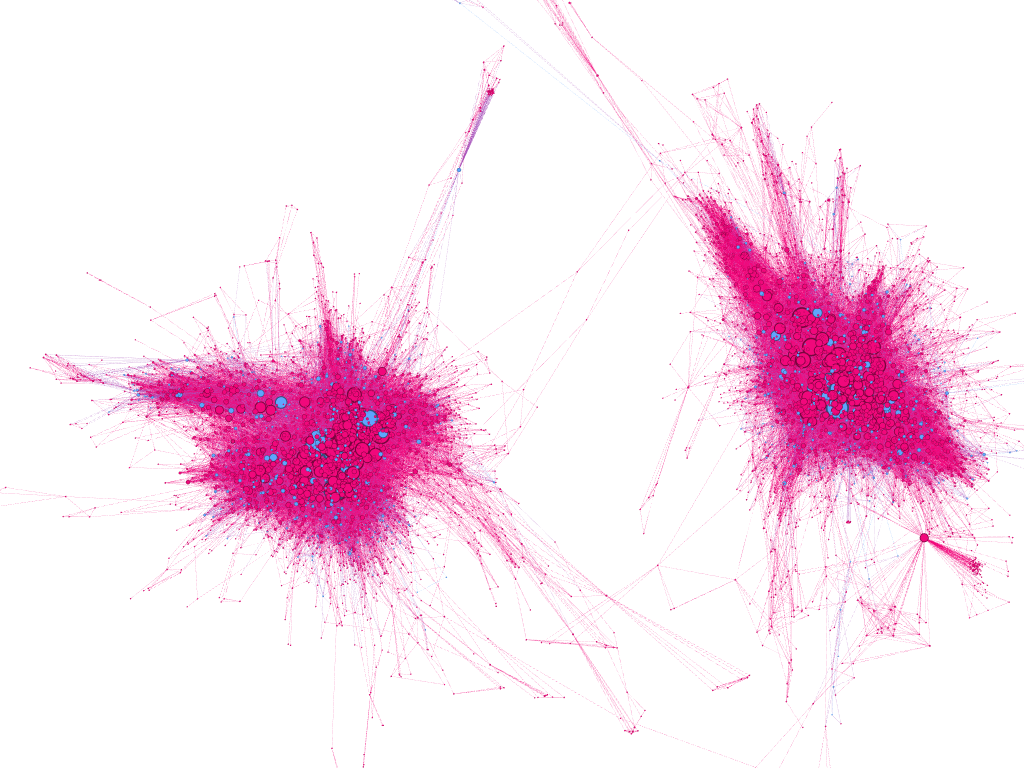}
\caption{Visualization of \mcsp result on \enron. Extracted subgraphs are highlighted in blue.}
\label{fig-enron_mcsp}
\end{figure*}

\begin{figure*}[h]
\centering
\includegraphics[width=0.6\textwidth]{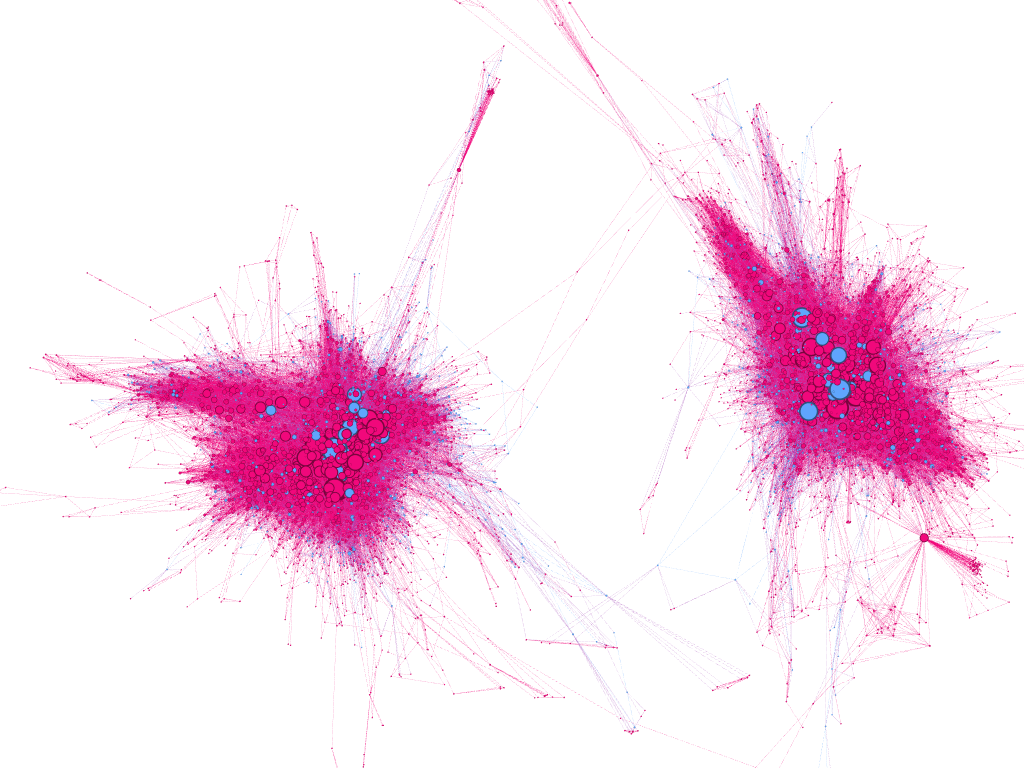}
\caption{Visualization of \mcsrlmodel result on \enron. Extracted subgraphs are highlighted in blue.}
\label{fig-enron_our}
\end{figure*}

\begin{figure*}[h]
\centering
\includegraphics[width=0.6\textwidth]{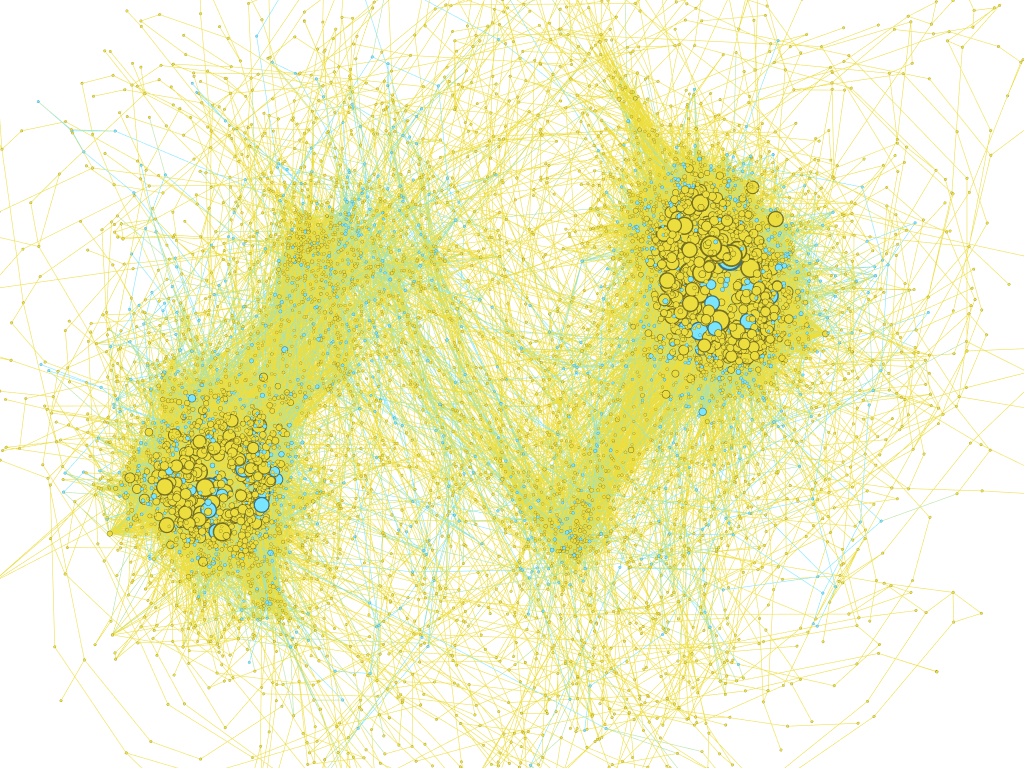}
\caption{Visualization of \mcsp result on \amazon. Extracted subgraphs are highlighted in blue.}
\label{fig-amazon_mcsp}
\end{figure*}

\begin{figure*}[h]
\centering
\includegraphics[width=0.6\textwidth]{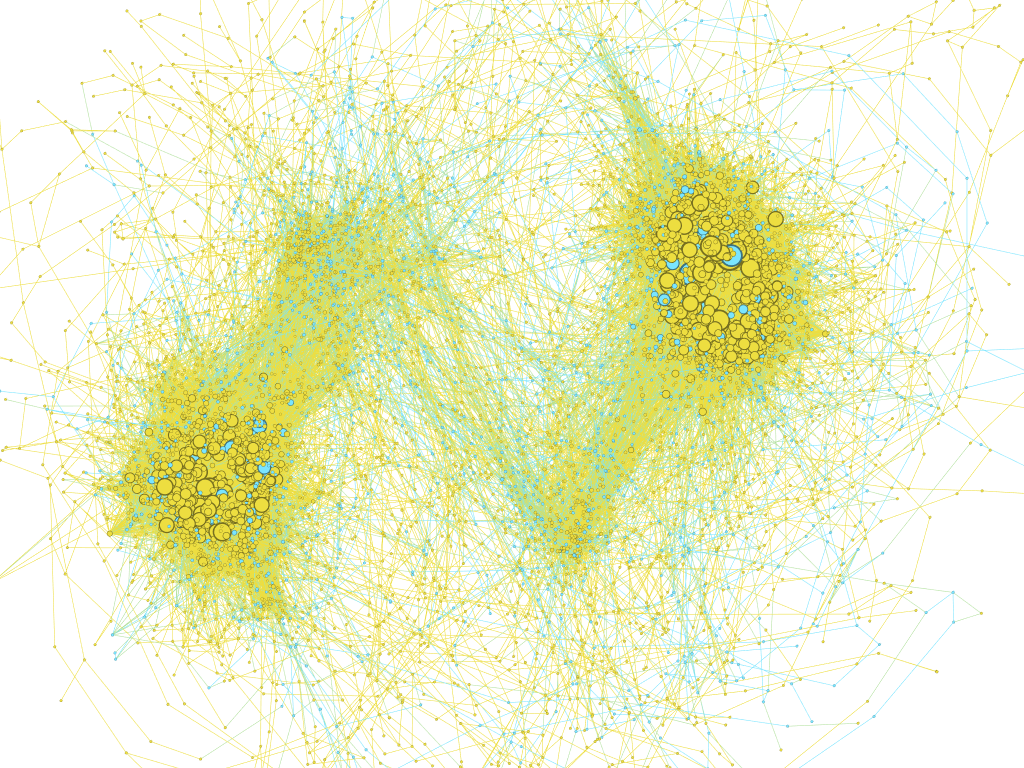}
\caption{Visualization of \mcsrlmodel result on \amazon. Extracted subgraphs are highlighted in blue.}
\label{fig-amazon_our}
\end{figure*}

\begin{figure*}
\centering
\includegraphics[width=1.0\textwidth]{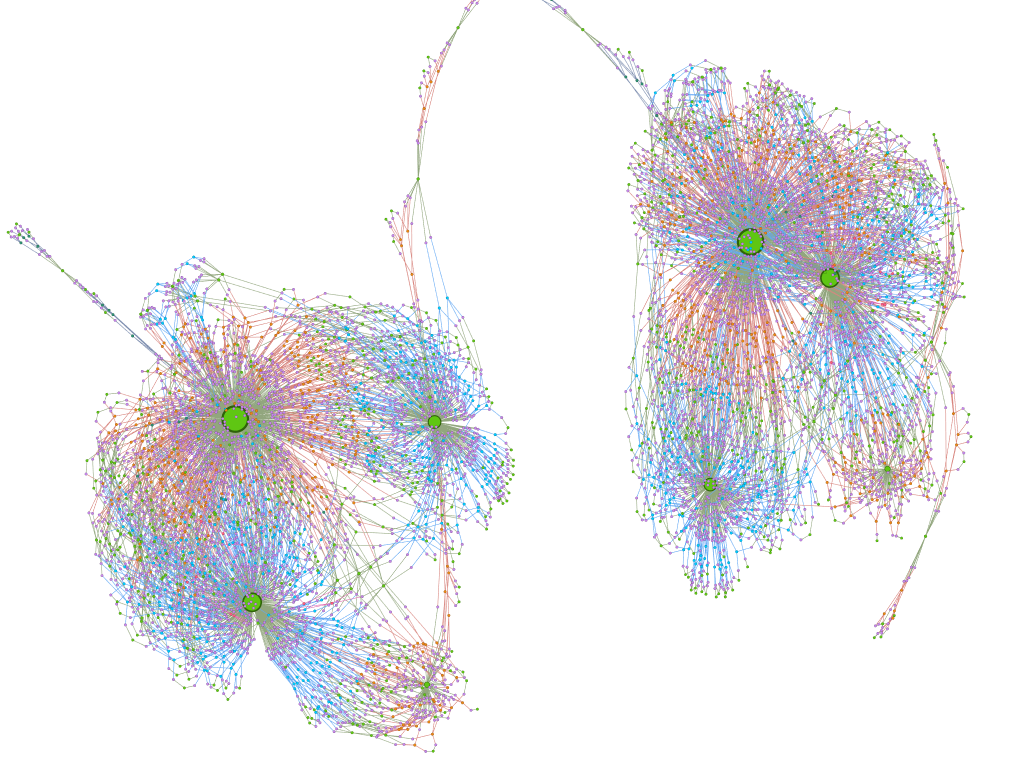}
\caption{Visualization of the original graph pair of \circuit. The two graphs are in fact isomorphic. Different colors denote different node labels. There are 6 node labels in total: M (71.67\%), null (10.41\%), PY (9.1\%), NY (8.23\%), N (0.37\%), and P (0.21\%).}
\vspace*{10mm}
\label{fig-circuit_orig}
\end{figure*}

\begin{figure*}[h]
\centering
\includegraphics[width=0.6\textwidth]{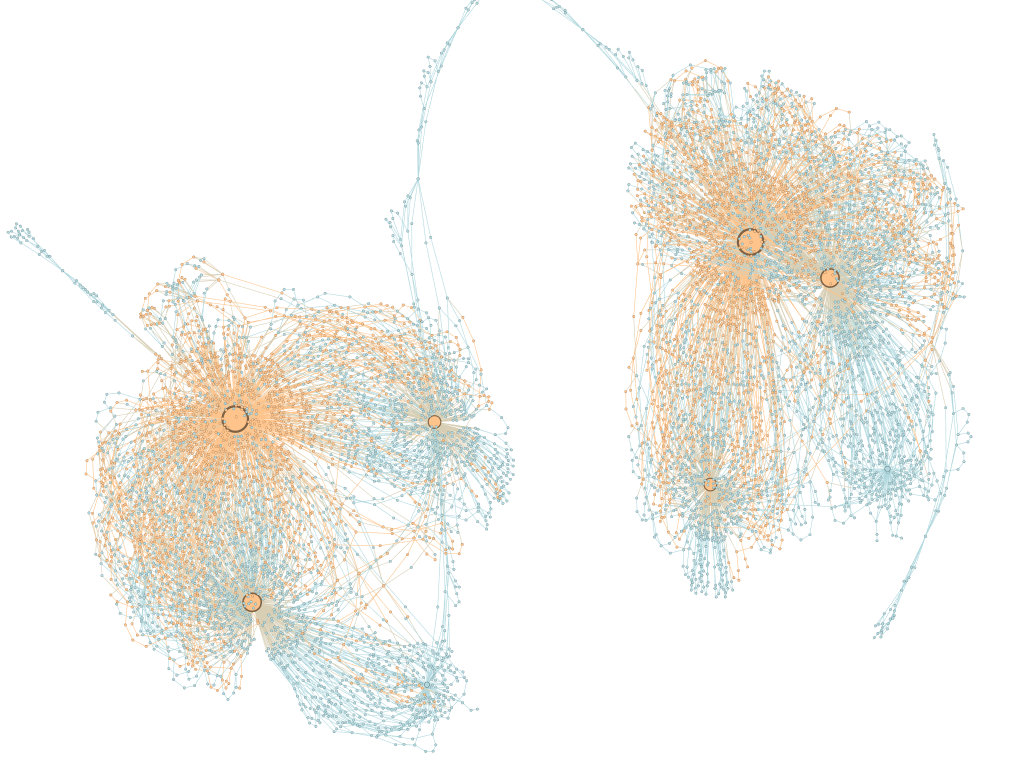}
\caption{Visualization of \mcsp result on \circuit. Extracted subgraphs are highlighted in yellow.}
\label{fig-circuit_mcsp}
\end{figure*}

\begin{figure*}[h]
\centering
\includegraphics[width=0.6\textwidth]{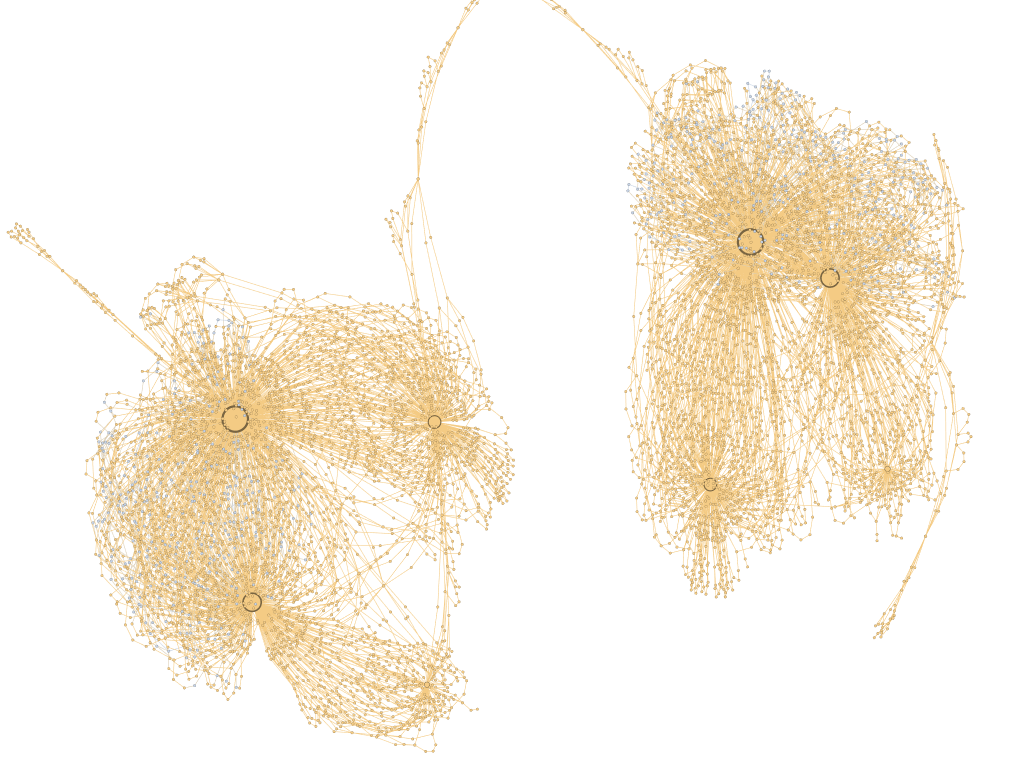}
\caption{Visualization of \mcsrlmodel result on \circuit. Extracted subgraphs are highlighted in yellow.}
\label{fig-circuit_our}
\end{figure*}

\begin{figure*}[h]
\centering
\includegraphics[width=0.6\textwidth]{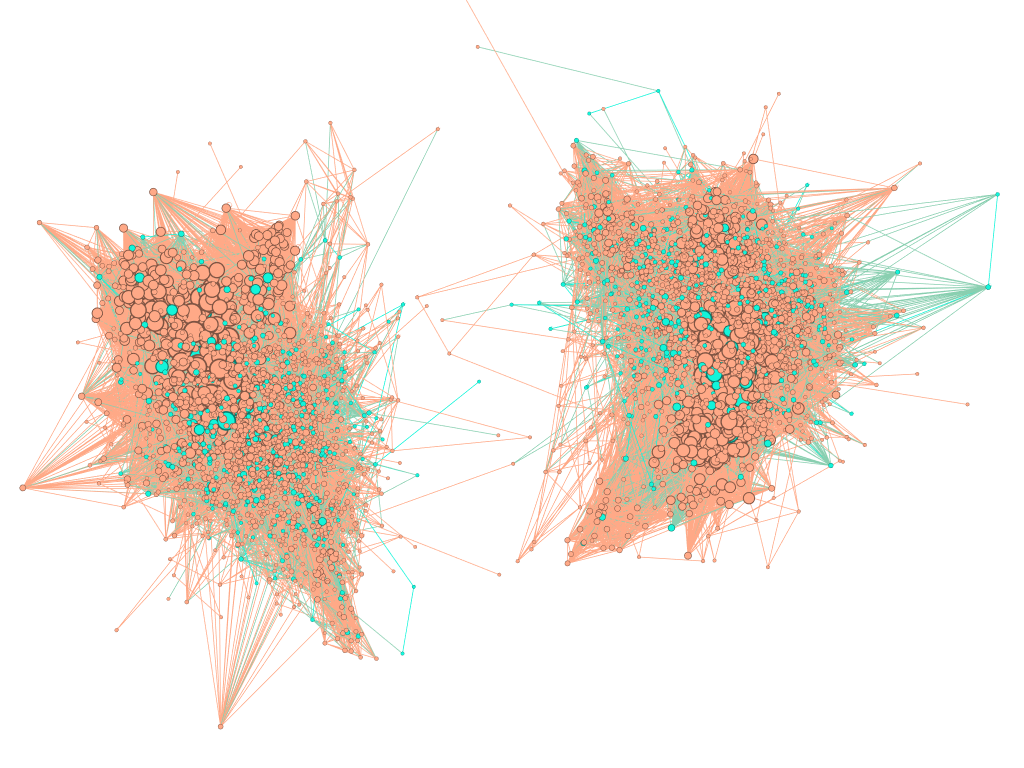}
\caption{Visualization of \mcsp result on \ppi. Extracted subgraphs are highlighted in cyan.}
\label{fig-ppi_mcsp}
\end{figure*}

\begin{figure*}[h]
\centering
\includegraphics[width=0.6\textwidth]{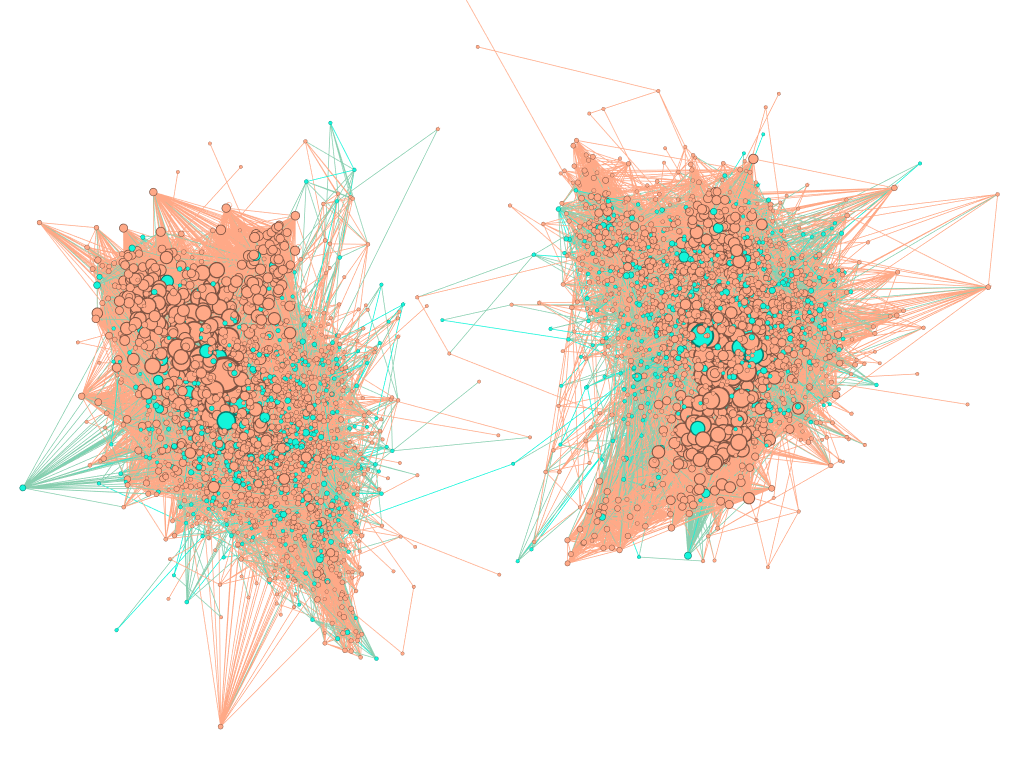}
\caption{Visualization of \mcsrlmodel result on \ppi. Extracted subgraphs are highlighted in cyan.}
\label{fig-ppi_our}
\end{figure*}


\end{document}